\documentclass[10pt,twocolumn,letterpaper]{article}

\usepackage{cvpr} 

\usepackage{booktabs}
\usepackage{multirow}
\usepackage{colortbl}

\definecolor{lightyellow}{RGB}{255, 249, 227}  
\definecolor{lightgreen}{RGB}{235, 252, 235}   
\definecolor{lightblue}{RGB}{230, 245, 255}    
\definecolor{deepyellow}{RGB}{255, 140, 0}  
\definecolor{deepgreen}{RGB}{0, 128, 0}     
\definecolor{cvprblue}{rgb}{0.21,0.49,0.74}
\usepackage[pagebackref,breaklinks,colorlinks,allcolors=cvprblue]{hyperref}

\usepackage[most]{tcolorbox}  
\definecolor{colorcardbox}{RGB}{240, 248, 255} 
\definecolor{colorcardborder}{RGB}{52, 52, 173} 

\usepackage[most]{tcolorbox}
\usepackage{float}

\newtcolorbox{textcolorbox}[1][]{
    enhanced,
    unbreakable,
    floatplacement=none,
    colback=colorcardbox,
    colframe=colorcardborder,
    colbacktitle=colorcardborder,
    coltitle=white,
    arc=2pt,
    title=#1,
    fonttitle=\bfseries,
    left=5pt,
    right=5pt,
    top=4pt,
    bottom=4pt,
    before skip=0.5em,
    after skip=0.75em,
    fontupper=\small
}

\def\paperID{} 
\def\confName{CVPR}
\def\confYear{2026}

\title{PhysicsMind: Sim and Real Mechanics Benchmarking for Physical Reasoning and Prediction in Foundational VLMs and World Models}

\author{%
\parbox{\textwidth}{%
\centering
\textbf{Chak-Wing Mak}\textsuperscript{*,1},
\textbf{Guanyu Zhu}\textsuperscript{*,1},
\textbf{Boyi Zhang}\textsuperscript{*,1},
\textbf{Hongji Li}\textsuperscript{2},
\textbf{Xiaowei Chi}\textsuperscript{1},
\textbf{Kevin Zhang}\textsuperscript{1},
\textbf{Yichen Wu}\textsuperscript{3},
\textbf{Yangfan He}\textsuperscript{4},
\textbf{Chun-Kai Fan}\textsuperscript{1},
\textbf{Wentao Lu}\textsuperscript{5},
\textbf{Kuangzhi Ge}\textsuperscript{1},
\textbf{Xinyu Fang}\textsuperscript{1},
\textbf{Hongyang He}\textsuperscript{6},
\textbf{Kuan Lu}\textsuperscript{7},
\textbf{Tianxiang Xu}\textsuperscript{1},
\textbf{Li Zhang}\textsuperscript{5,8},
\textbf{Yongxin Ni}\textsuperscript{9},
\textbf{Youhua Li}\textsuperscript{10},
\textbf{Shanghang Zhang}\textsuperscript{\textdagger,1}\\[0.8em]
{\small
\textsuperscript{1}Peking University \quad
\textsuperscript{2}Mohamed bin Zayed University of Artificial Intelligence \quad
\textsuperscript{3}National University of Singapore \quad
\textsuperscript{4}University of North Carolina at Chapel Hill \quad
\textsuperscript{5}University of Science and Technology of China \quad
\textsuperscript{6}Manifold.AI \quad
\textsuperscript{7}Cornell University \quad
\textsuperscript{8}Hong Kong Polytechnic University \quad
\textsuperscript{9}National University of Singapore \quad
\textsuperscript{10}City University of Hong Kong
}\\[1em]
{\small
\textsuperscript{*}\textit{Equal contribution.} \qquad
\textsuperscript{\textdagger}\textit{Corresponding author.}
}
}%
}

\begin{document}
\maketitle

\begin{abstract}

Modern foundational Multimodal Large Language Models (MLLMs) and video world models have advanced significantly in mathematical, common-sense, and visual reasoning, but their grasp of the underlying physics remains underexplored. Existing benchmarks attempting to measure this matter rely on synthetic, Visual Question Answer templates or focus on perceptual video quality that is tangential to measuring how well the video abides by physical laws. To address this fragmentation, we introduce \textbf{PhysicsMind}, a unified benchmark with both real and simulation environments that evaluates law-consistent reasoning and generation over three canonical principles: Center of Mass, Lever Equilibrium, and Newton’s First Law. PhysicsMind comprises two main tasks: i) VQA tasks, testing whether models can reason and determine physical quantities and values from images or short videos, and ii) Video Generation(VG) tasks, evaluating if predicted motion trajectories obey the same center-of-mass, torque, and inertial constraints as the ground truth. A broad range of recent models and video generation models is evaluated on PhysicsMind and found to rely on appearance heuristics while often violating basic mechanics. These gaps indicate that current scaling and training are still insufficient for robust physical understanding, underscoring PhysicsMind as a focused testbed for physics-aware multimodal models. Our data will be released upon acceptance.

\end{abstract}
\section{Introduction}
\vspace{-0.05in}
Recent multimodal foundation models and video generation models (often used as world models in recent works) have made rapid progress in mathematical reasoning, commonsense inference, and generic multimodal understanding~\cite{qwenteam2025qwen3,nvidia2025cosmosworldfoundationmodel,openai_sora2_system_card}, but their physical understanding and prediction of real-world dynamics are often limited~\cite{motamed2025physicsiq, phybench2025}. Compared to symbolic or linguistic tasks, such capabilities require models to not only reason about abstract concepts such as force, mass, momentum, and inertia, but also to predict how these quantities manifest in visually observable and temporally coherent motion. These laws of mechanics govern how objects move, interact, or remain stable, and are therefore central to perception, prediction, and embodied decision making in both artificial and biological agents, thus making it an important quality to measure across current Visual-Language Models (VLMs) and Video Generation Models (VGMs).

\begin{figure}[t]
    \centering
    \includegraphics[width=1\linewidth]{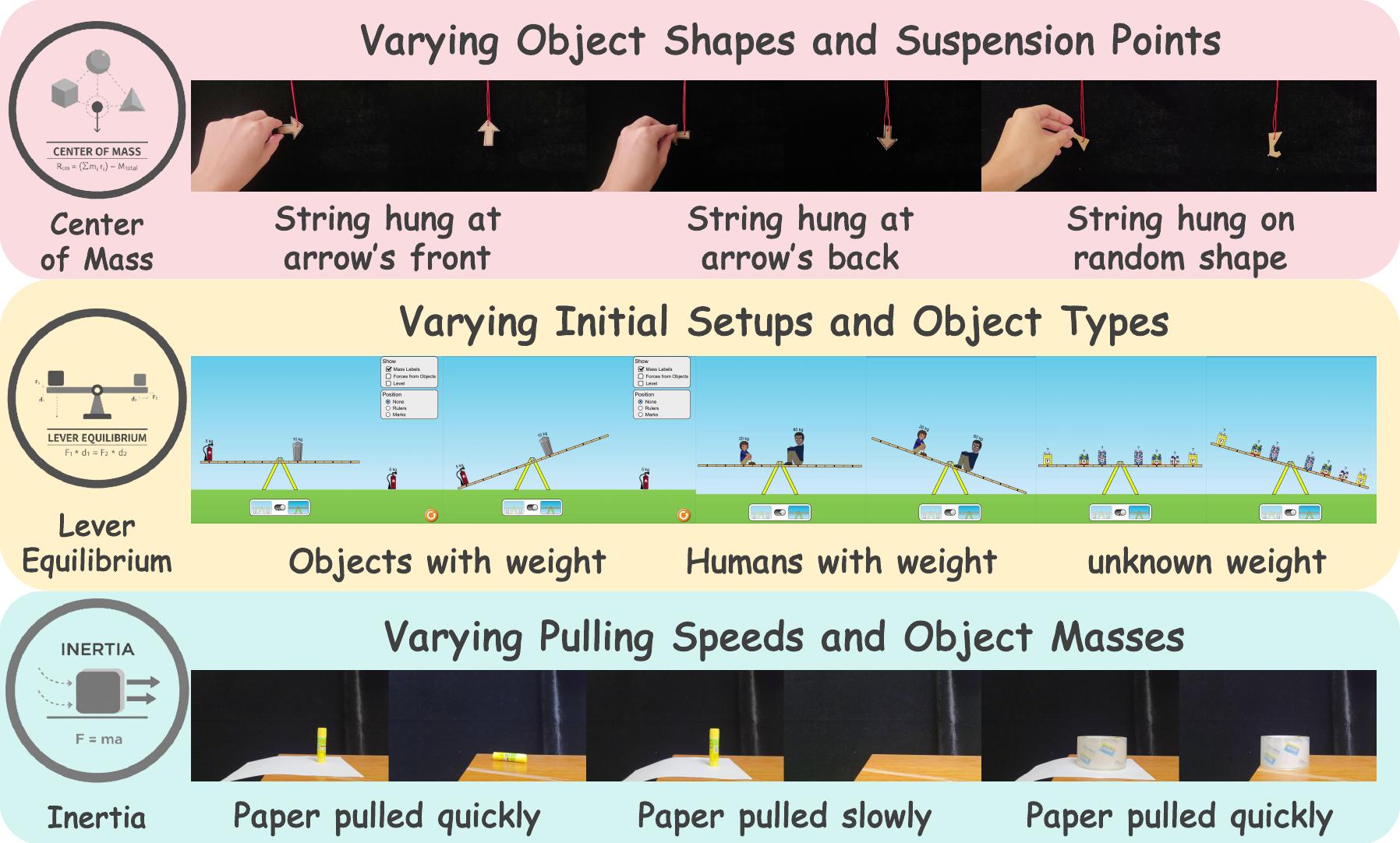}
    \caption{ Three canonical mechanics scenarios in PhysicsMind: Center of Mass, lever equilibrium, and Newton’s first law, each realized with various real tabletop and simulated configurations.}
    \label{fig:dataset}
    \vspace{-6mm}
\end{figure}

Prior works on the physical understanding of generative models fall primarily into two large categories.
The first line studies VLMs on physical question answering and visual reasoning, such as IntPhys~\cite{riochet2018intphys}, CRAFT~\cite{smith2022craft}, CausalVQA~\cite{foss2025causalvqaphysicallygroundedcausal}, MVP~\cite{krojer2025shortcutawarevideoqabenchmarkphysical},  PhysBench~\cite{phybench2025}, and PhysUniBench~\cite{wang2025physunibench}. These benchmarks provide short videos or images paired with template questions about collisions, stability, and causal relations. While they are valuable for probing whether models can answer physics questions, the visual scenarios are often synthetic, with regular question formats highly regular, and plain evaluation methods, allowing methods to exploit linguistic or visual shortcuts without interpreting and reasoning about mechanical interactions.

The second category evaluates VGMs and World Models on their abilities to produce physically plausible dynamics. Recent datasets and benchmarks such as VBench~\cite{li2024vbench}, PhyGenBench~\cite{phygenbench2024}, Physics-IQ~\cite{motamed2025physicsiq}, MORPHEUS~\cite{morpheus2025}, WorldModelBench~\cite{worldmodelbench2025}, and WorldScore~\cite{worldscore2025} are widely used to test large diffusion models such as Sora, Veo, etc. However, these methods suffer from simple evaluation protocols focusing on pixel-level reconstruction quality, motion smoothness, or human-judged visual plausibility. In practice, visually impressive sequences can still violate basic mechanics, exhibit unstable or energy non-conserving motion, or hallucinate physically impossible trajectories. 

To move beyond this fragmented landscape, we focus on three canonical mechanics problems, Center of Mass, Lever Equilibrium, and Newton’s First Law, which capture complementary aspects of physical understanding while remaining simple enough for both controlled tabletop experiments and 2D simulations. Building on this structure, we design PhysicsMind, a unified benchmark that bridges physical reasoning and physical prediction with two complementary tasks. The first is video question answering (VQA), where VLMs receive images or short clips together with multiple-choice questions about the end state, direction of rotation, or stability of the system. This task probes whether models can extract relevant physical quantities from visual input and apply the appropriate law to reach a discrete conclusion. The second is video generation, where state-of-the-art video generators are conditioned on an initial state and produce future trajectories. In this work, we evaluate whether generated motions follow the same center-of-mass, lever, and inertia laws that govern the ground-truth experiments.
This design allows us to explore two main topics. First, how well do real-looking generated videos abide by physical laws, and second, if models that appear competent in simulation environments generalize well to the same principles under real-world noise, dynamic differences, and visual variability.


We perform extensive evaluations of state-of-the-art VLMs and video generators. Despite strong perceptual capabilities and high visual fidelity, current baselines show consistent gaps with basic physics: in VQA tasks, current VLMs rely on appearance heuristics and struggle with counterfactual and logically related questions, while in video generation VGMs often violate center-of-mass constraints, mispredict lever outcomes, or ignore inertia, indicating that current scaling and training strategies have not yielded robust physical understanding or faithful prediction. Our main contributions are as follows.

\begin{enumerate}
    \item We introduce PhysicsMind, a unified physics benchmark that evaluates both reasoning and prediction under the same laws in simulated and real-world settings. We focus on 3 mechanics scenarios, including Center of Mass, Lever Equilibrium, and Newton’s First Law.
    
    \item We design physics-aware evaluation protocols that go beyond overall accuracy or visual quality: for VQA, law-specific subtasks (e.g., position vs. rotation, equilibrium vs. adjustment), and for video generation, metrics for center-of-mass alignment, lever final-state correctness, and trajectory, speed, and acceleration consistency.
    
    \item We conduct a systematic study of modern VLMs and video generators on PhysicsMind, revealing consistent failure modes such as reliance on appearance cues, weak causal reasoning, and frequent violations of basic mechanics, thereby outlining concrete targets for future physics-aware multimodal modeling.
    
\end{enumerate}

\begin{table*}[htbp]
\centering
\caption{
Comparison of physics-oriented VQA benchmarks. Columns summarize dataset realism, task modality, annotation format, and the scope of physical evaluation; PhysicsMind adds law-targeted cross-checks on shared real and simulated setups.
}
\label{tab:VQA_compare}

\resizebox{1\textwidth}{!}{
\begin{tabular}{lccccccccc}
\toprule
Benchmark & Models & Realism & Task Modal & Answer Type & Video Length & Max Objects & Eval Aspects & CrossCheck \\
\midrule
IntPhys \cite{riochet2018intphys} & 2 & Sim & Dynamic & Score & 7s & 3 & 3 & No \\
CRAFT \cite{smith2022craft} & 9 & Sim & Dynamic & Choice & \textcolor{deepgreen}{10s} & \textcolor{deepgreen}{6+} & 3 & No \\
CausalVQA \cite{foss2025causalvqaphysicallygroundedcausal} & 6 & Real & Dynamic & Choice & \textcolor{deepgreen}{24s} & 3 & 5 & No \\
MVP \cite{krojer2025shortcutawarevideoqabenchmarkphysical} & 13 & Real & Dynamic & Choice & 9s & \textcolor{deepgreen}{6+} & 5 & No \\
PhysBench \cite{phybench2025} & \textcolor{deepgreen}{39} & \textcolor{deepgreen}{Real+Sim} & \textcolor{deepgreen}{Dynamic+Static} & Choice & 8 frames & 5 & 4 & \textcolor{deepgreen}{Yes} \\
\midrule
PhysicsMind (Ours) & \textcolor{deepgreen}{22} & \textcolor{deepgreen}{Real+Sim} & \textcolor{deepgreen}{Dynamic+Static} & Choice & \textcolor{deepgreen}{4--11s} & \textcolor{deepgreen}{6+} & \textcolor{deepgreen}{6} & \textcolor{deepgreen}{Yes} \\
\bottomrule
\end{tabular}}
\end{table*}

\begin{table*}[htbp]
\centering
\Large
\caption{
Comparison of video generation and world-model benchmarks. The table contrasts realism, supervision, and evaluation scope; PhysicsMind provides paired real–sim trajectories and law-aware cross-checks for mechanics-focused video prediction.
}
\label{tab:gen_compare}

\resizebox{1\textwidth}{!}{
\begin{tabular}{lccccccccc}
\toprule
Benchmark & Realism & Ground-Truth Video & Video Length (s) & Input Modalities & Max Objects & Video Dim & Eval Aspects & CrossCheck \\
\midrule
WorldModelBench \cite{worldmodelbench2025} & Sim & No & 2--5 & Text/Image & 1 & 3 & \textcolor{deepgreen}{7} & No \\
MORPHEUS \cite{morpheus2025} & Real & No & 2--4 & \textcolor{deepgreen}{Image/Video} & 2 & 3 & 3 & No \\
WorldScore \cite{worldscore2025} & Real & \textcolor{deepgreen}{yes} & 2--3 & Image & \textcolor{deepgreen}{6+} & 3 & \textcolor{deepgreen}{10} & No \\
PhyGenBench \cite{phygenbench2024} & Sim & No & \textcolor{deepgreen}{3--11} & Text & 1 & 3 & 4 & No \\
VBench \cite{li2024vbench} & Sim & No & 2 & Text & 3 & 3 & \textcolor{deepgreen}{16} & No \\
Physics-IQ \cite{motamed2025physicsiq} & \textcolor{deepgreen}{Real+Sim} & \textcolor{deepgreen}{yes} & 8 & \textcolor{deepgreen}{Image/Video} & 3 & 3 & 4 & No \\
\midrule
PhysicsMind (Ours) & \textcolor{deepgreen}{Real+Sim} & \textcolor{deepgreen}{Yes} & \textcolor{deepgreen}{5--10} & \textcolor{deepgreen}{Image/Video} & \textcolor{deepgreen}{6+} & \textcolor{deepgreen}{2/3} & \textcolor{deepgreen}{8} & \textcolor{deepgreen}{Yes} \\
\bottomrule
\end{tabular}
}
\end{table*}

\vspace{-0.1in}
\section{Related Work}
\vspace{-0.05in}
\noindent\textbf{Physical reasoning benchmarks.}
Physical reasoning benchmarks aim to test whether a model can infer and apply basic physical principles from visual input. Early work relied on synthetic, highly controlled environments. CLEVR~\cite{johnson2017clevr} targets compositional reasoning in rendered scenes, while IntPhys~\cite{riochet2018intphys}, CRAFT~\cite{smith2022craft}, and PHYRE~\cite{bakhtin2019phyre} probe intuitive physics and causal interactions in simulators. More recent datasets, such as CausalVQA~\cite{foss2025causalvqaphysicallygroundedcausal}, MVP~\cite{krojer2025shortcutawarevideoqabenchmarkphysical}, PhysBench~\cite{phybench2025}, PhysUniBench~\cite{wang2025physunibench}, and Physics-IQ~\cite{motamed2025physicsiq} move toward richer video question answering with more realistic footage. However, most of these benchmarks use templated question formats, focus on narrow sets of scenarios, and report aggregate accuracy without explicitly tying evaluation to concrete mechanics laws or to predictive behavior. PAC Bench~\cite{gundawar2025pacbench} systematically evaluates VLMs on their understanding of physical properties, affordances, and constraints for robot manipulation. Recent work has also explored why spatial reasoning remains challenging for VLMs from an attention mechanism perspective~\cite{chen2025adaptvis}. 

\noindent\textbf{Video generation and world models.}
 World models~\cite{openai_sora2_system_card,nvidia2025cosmosworldfoundationmodel,zhang2025worldlm,chi2025wow,chi2024eva} and video generators learn predictive dynamics from streams of observations.Classical world models such as PlaNet~\cite{hafner2019planet} and the Dreamer family~\cite{hafner2019dreamer} are typically evaluated through downstream control returns rather than explicit tests of physical consistency.   More recent benchmarks for generative dynamics, including VBench~\cite{li2024vbench}, PhyGenBench~\cite{phygenbench2024}, Physics-IQ~\cite{motamed2025physicsiq}, MORPHEUS~\cite{morpheus2025}, WorldModelBench~\cite{worldmodelbench2025}, WorldScore~\cite{worldscore2025}, WorldPrediction~\cite{chen2025worldprediction}, and AutumnBench~\cite{warrier2024benchmarkingworld}, assess visual fidelity, motion smoothness, and human preference for sequences produced by text- or image-conditioned generators.These metrics provide useful signals for perceptual quality but only indirectly reflect whether trajectories respect center-of-mass balance, lever behavior, or inertial motion, and they are rarely aligned with law-targeted reasoning tasks on the same physical scenarios. He et al.~\cite{he2024llmsmultimodal} reviewed the intersection of LLMs with multimodal generation and editing. Motamed et al.~\cite{motamed2025physicsiq} investigated whether generative video models truly understand physical principles, revealing that visual realism does not imply physical understanding. Lin et al.~\cite{lin2025physicscognition} systematically reviewed the evolution of physical cognition in video generation, highlighting the gap between visual realism and physical consistency.
 
\vspace{-0.01in}  
\noindent\textbf{Position.}
Large multimodal foundation models\cite{qwenteam2025qwen3,OpenAI2025GPT5SystemCard,comanici2025gemini25,anthropic2025claudesonnet45} and modern video generators\cite{wan2025,deepmind_veo_model_card,openai_sora2_system_card} have greatly advanced visual recognition, language-conditioned reasoning, and open-domain synthesis, yet they still often misinterpret basic mechanics and produce physically implausible motion,  as is shown in Table \ref{tab:VQA_compare} and \ref{tab:gen_compare}. PhysicsMind makes this gap explicit by evaluating law-targeted VQA and video prediction on shared center-of-mass, lever, and inertia scenarios with paired real–simulation data and metrics that directly test adherence to physical laws.

\begin{figure*}[htb]
  \centering
  \includegraphics[width=\textwidth]{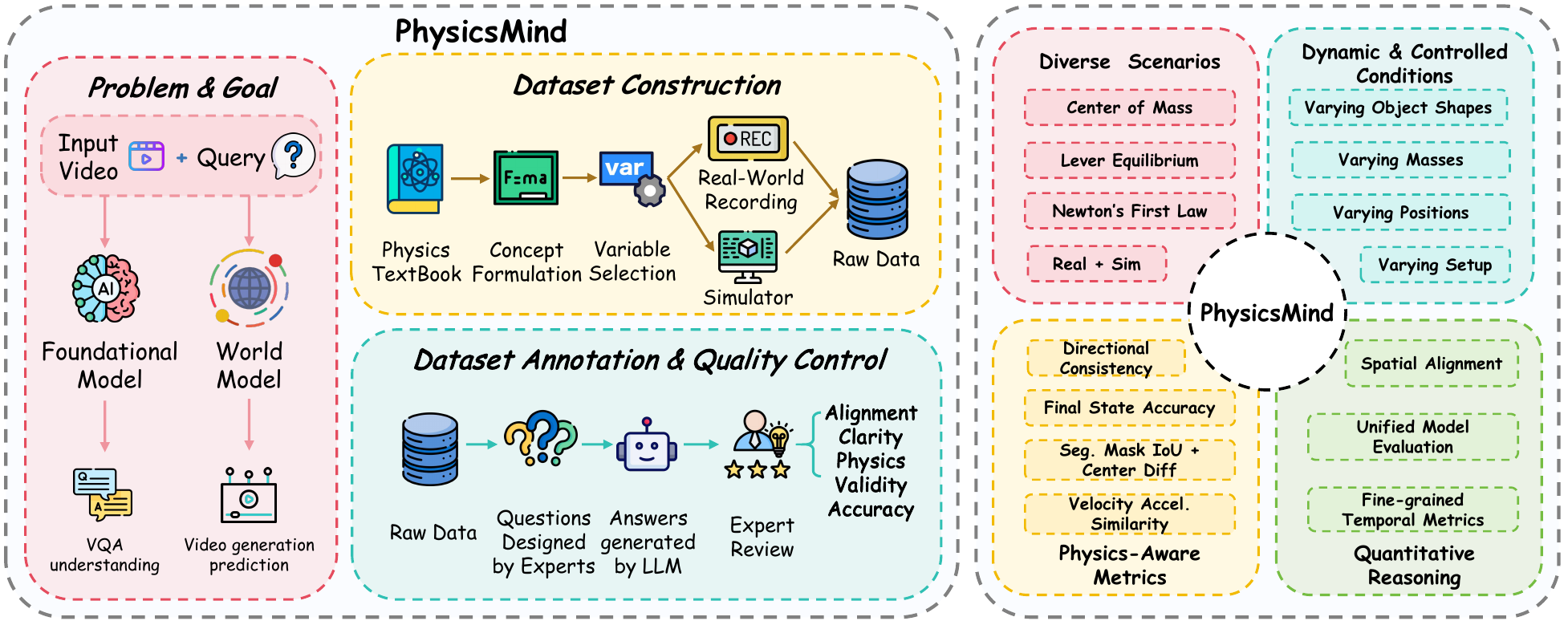}
  \caption{Overview of the PhysicsMind framework. It combines a foundational model with physics-guided dataset construction, expert-verified annotations, and diverse controlled scenarios to enable robust video understanding and physics-aware evaluation.}
  \label{fig:overview}
\end{figure*}

\begin{figure*}[htb]
  \centering
  \includegraphics[width=\textwidth]{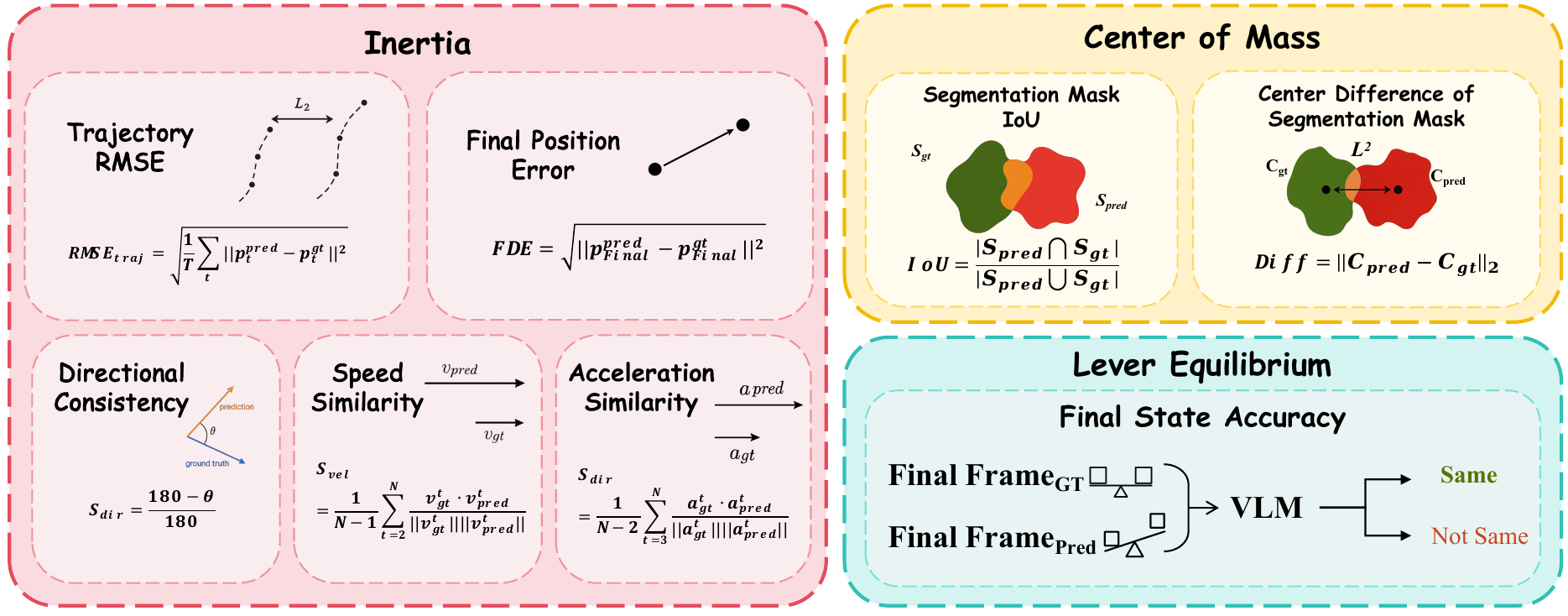}
  \caption{Physics-Aware evaluation metrics for Video Generation Models (VGM). Inertia metrics assess motion and trajectory consistency, Center‑of‑Mass metrics measure segmentation alignment, and Lever‑Equilibrium evaluates final‑state agreement.}
  \label{fig:vg_metrics}
\end{figure*}

\section{The PhysicsMind Benchmark}
\vspace{-0.05in}
\subsection{Problem Formulation}
\vspace{-0.05in}
PhysicsMind formulates the evaluation of physical commonsense reasoning in foundation models and world models as a unified understanding task that spans both VQA and VG. The objective is to assess whether a model can both visualize and reason about fundamental physical phenomena with scientific validity and adherence to physical laws.

The evaluation of physical reasoning in both foundation and world models can be formalized within a unified framework that captures their shared structure and distinct modalities. For foundation models, the problem is defined as follows: given an initial frame \(f\) and a question \(q\), the model produces a response \(\hat{a} = f_{\theta}(f, q)\), which is then evaluated by comparing \(\hat{a}\) with the ground truth \(a^{*}\). For world models, the formulation extends to temporal data: given an input video sequence \(\mathbf{x} = (x_{1}, x_{2}, \dots, x_{T})\) and a corresponding physical query \(q\), the model generates a response \(\hat{a} = f_{\theta}(\mathbf{x}, q)\). Evaluation in this case considers both the final output and the transitional states through an evaluation function \(E(\hat{a}, a^{*})\), which measures the presence and correctness of physical commonsense, where \(a^{*}\) denotes the ground-truth or physically consistent outcome.

\subsection{Overview of PhysicsMind}
\vspace{-0.05in}
PhysicsMind evaluates the physical reasoning capabilities of both foundation models and world models. The benchmark introduces two complementary tasks, Visual Question Answering (VQA) and Video Generation Assessment, shown in the left region of Figure~\ref{fig:overview}. As illustrated in Figures~\ref{fig:dataset}, PhysicsMind covers three physics scenarios: Center of Mass, Lever Equilibrium, and Newton’s First Law. These three physical domains are intentionally selected to represent orthogonal dimensions of physical reasoning: geometry (Center of Mass), causality (Lever Equilibrium), and dynamics (Newton’s First Law). This triad collectively bridges the spectrum from static perception to mechanical interaction to temporal prediction, allowing PhysicsMind to probe both intuitive and formal physical understanding in foundation models. Each scenario is implemented through both real-world recordings and two-dimensional (2D) physics simulations, enabling systematic comparisons between natural and controlled physical conditions.

The dataset comprises short, high-quality video clips with a maximum resolution of 3840 × 2160, recorded at 60 FPS and lasting up to 10 seconds. Real-world samples are captured in controlled indoor environments with fixed camera positions to ensure consistent framing and illumination. Simulated samples are generated using a deterministic 2D physics engine, allowing fine-grained control over object attributes and motion reproducibility. Across all scenarios, the dataset incorporates diverse configurations that vary in initial setups, object shapes, masses, and conditions.

\noindent\textbf{Dataset Construction.} As shown in the top-middle region of Figure~\ref{fig:overview}, the PhysicsMind dataset is developed through a structured pipeline consisting of conceptualization, experimental design, and data generation. The process begins with identifying visually interpretable and experimentally verifiable laws from a high school physics textbook. These principles are translated into controlled physical setups, and a systematic variable selection process ensures diversity and control by varying conditions. To ensure data diversity, both simulated and real-world recordings are used. Simulated data is generated with a deterministic 2D physics engine with precise control, while real-world recordings are produced under fixed camera viewpoints and consistent lighting for stable visual conditions. This dual-source design ensures reproducibility in simulated settings while preserving natural variability in real experiments.

\noindent\textbf{Data Annotation and Quality Control.} PhysicsMind employs a multi-stage annotation and verification workflow (shown in the bottom-middle region of Figure~\ref{fig:overview}). Expert curators first design physics-based questions and answer templates, which are automatically populated using large language models and then manually refined. Each video is paired with its physical descriptors, annotated variables, and validated VQA entries. A series of manual and automated consistency checks follows to confirm alignment, clarity, and physical correctness. Videos that exhibit ambiguity, instability, or deviations from expected dynamics are systematically excluded. This rigorous curation process yields a dataset that integrates controlled precision with realistic diversity, forming a reliable foundation for evaluating physical reasoning in both foundation and world models.

\subsection{Evaluation Metrics}
\vspace{-0.05in}
\textbf{Video Question Answering.}
The core metric is binary exact-match accuracy with respect to the ground-truth option.  
To obtain finer diagnostics, we further report:
\begin{itemize}[leftmargin=*]
    \item \textit{Category-wise accuracy}, split by law-specific subtypes (e.g., equilibrium vs.\ adjustment for levers, position vs.\ rotation for Center of Mass, position vs.\ stability for inertia), to distinguish different reasoning skills.
    \item \textit{Experiment-wise accuracy}, aggregated over physical setups and configurations, to reveal which conditions are easier or harder for models.
    \item \textit{Variant-wise accuracy}, grouped by object type, mass pattern, or motion profile, to test robustness to appearance and parameter changes.
\end{itemize}

\noindent  \textbf{Video Generation.}
To quantify the physical plausibility of generated videos, we introduce a law-aware evaluation framework that decomposes performance by mechanics scenario (see Figure~\ref{fig:vg_metrics}):
\begin{itemize}[leftmargin=*]
    \item \textit{Center of Mass.} We measure geometric fidelity via Intersection-over-Union (IoU) between predicted and reference object masks, together with the distance between their mask centroids.
    \item \textit{Lever Equilibrium.} We compute final-state prediction accuracy, checking whether generated sequence converges to the correct lever outcome implied by torque balance.
    \item \textit{Newton’s first law.} We use a set of kinematic metrics—trajectory deviation, velocity consistency, and acceleration fluctuation—to test whether objects maintain constant motion in the absence of external forces.
\end{itemize}
Detailed formulations are provided in the Appendix.

\begin{table*}[t]
  \caption{VQA Physics Evaluation Results across Models. The Acc columns report accuracy in percentage (\%). Values after $\pm$ represent standard deviation across multiple runs ($n=5$).}
  \label{tab:VQA_results}
  \centering
  \resizebox{\textwidth}{!}{
    \begin{tabular}{l|ccc|ccc|ccc}
      \toprule
      \multirow{3}{*}{Model} &
      \multicolumn{3}{c|}{Center of Mass (VQA)} &
      \multicolumn{3}{c|}{Lever Equilibrium (VQA)} &
      \multicolumn{3}{c}{Newton's First Law (VQA)} \\
      \cmidrule(lr){2-4} \cmidrule(lr){5-7} \cmidrule(lr){8-10}
       & Position & Rotation & Overall &
         Equilibrium & Balance Adj. & Overall &
         Obj. Pos. & Obj. Stability & Overall \\
       & Acc (\%) ↑ & Acc (\%) ↑ & Acc (\%) ↑ &
         Acc (\%) ↑ & Acc (\%) ↑ & Acc (\%) ↑ &
         Acc (\%) ↑ & Acc (\%) ↑ & Acc (\%) ↑ \\
      \midrule

      \rowcolor{lightyellow}
      \multicolumn{10}{c}{\textbf{Closed-source Reasoning Models}} \\

      GPT-5 \cite{OpenAI2025GPT5SystemCard} &
        \underline{\textbf{60.00}}{\tiny$\pm$2.1} & 80.00{\tiny$\pm$1.8} & \underline{\textbf{70.00}}{\tiny$\pm$1.5} &
        66.67{\tiny$\pm$2.4} & \underline{\textbf{85.71}}{\tiny$\pm$1.5} & \underline{\textbf{76.19}}{\tiny$\pm$1.3} &
        60.00{\tiny$\pm$2.8} & \underline{\textbf{95.00}}{\tiny$\pm$1.0} & 77.50{\tiny$\pm$1.8} \\

      o4-min \cite{OpenAI2025O3O4MiniSystemCard} &
        50.00{\tiny$\pm$2.4} & 55.00{\tiny$\pm$2.2} & 52.50{\tiny$\pm$1.9} &
        61.90{\tiny$\pm$2.5} & 52.38{\tiny$\pm$2.4} & 57.14{\tiny$\pm$2.1} &
        \underline{\textbf{75.00}}{\tiny$\pm$2.2} & \underline{\textbf{95.00}}{\tiny$\pm$1.0} & \underline{\textbf{85.00}}{\tiny$\pm$1.5} \\

      GPT-4o \cite{openai2024gpt4ocard} &
        40.00{\tiny$\pm$2.4} & 35.00{\tiny$\pm$2.1} & 37.50{\tiny$\pm$1.9} &
        42.86{\tiny$\pm$2.5} & 66.67{\tiny$\pm$2.1} & 54.76{\tiny$\pm$1.9} &
        45.00{\tiny$\pm$2.8} & 85.00{\tiny$\pm$1.6} & 65.00{\tiny$\pm$2.1} \\

      GPT-4-turbo \cite{openai2024gpt4technicalreport} &
        25.00{\tiny$\pm$2.2} & \underline{\textbf{80.00}}{\tiny$\pm$1.8} & 47.50{\tiny$\pm$2.0} &
        28.57{\tiny$\pm$2.3} & 57.14{\tiny$\pm$2.2} & 42.86{\tiny$\pm$2.0} &
        40.00{\tiny$\pm$2.5} & 80.00{\tiny$\pm$1.8} & 60.00{\tiny$\pm$2.0} \\

      Claude-4.5-sonnet \cite{anthropic2025claudesonnet45} &
        45.00{\tiny$\pm$2.5} & 20.00{\tiny$\pm$1.8} & 32.50{\tiny$\pm$2.1} &
        61.90{\tiny$\pm$2.4} & 61.90{\tiny$\pm$2.0} & 61.90{\tiny$\pm$1.8} &
        10.00{\tiny$\pm$1.4} & 90.00{\tiny$\pm$1.3} & 50.00{\tiny$\pm$2.2} \\

      Claude-3.7-sonnet \cite{anthropic2025Claude37sonnet} &
        30.00{\tiny$\pm$2.3} & 25.00{\tiny$\pm$2.0} & 27.50{\tiny$\pm$1.9} &
        52.38{\tiny$\pm$2.4} & 66.67{\tiny$\pm$2.0} & 59.52{\tiny$\pm$1.9} &
        40.00{\tiny$\pm$2.5} & 65.00{\tiny$\pm$2.1} & 52.50{\tiny$\pm$2.0} \\

      Gemini-2.5-pro \cite{comanici2025gemini25} &
        30.00{\tiny$\pm$2.2} & 70.00{\tiny$\pm$1.9} & 50.00{\tiny$\pm$1.8} &
        57.14{\tiny$\pm$2.3} & 76.19{\tiny$\pm$1.7} & 66.67{\tiny$\pm$1.6} &
        20.00{\tiny$\pm$1.8} & 45.00{\tiny$\pm$2.2} & 32.50{\tiny$\pm$1.9} \\

      \midrule
      \rowcolor{lightyellow}
      \multicolumn{10}{c}{\textbf{Closed-source Chat Models}} \\

      GPT-4o-mini \cite{openai2024gpt4ocard} &
        10.00{\tiny$\pm$1.5} & 20.00{\tiny$\pm$1.8} & 15.00{\tiny$\pm$1.6} &
        23.81{\tiny$\pm$2.1} & 38.10{\tiny$\pm$2.2} & 30.95{\tiny$\pm$1.8} &
        10.00{\tiny$\pm$1.4} & 80.00{\tiny$\pm$1.8} & 45.00{\tiny$\pm$2.1} \\

      GPT-4.1 \cite{openai2024gpt4technicalreport} &
        21.30{\tiny$\pm$2.2} & 46.75{\tiny$\pm$2.3} & 43.00{\tiny$\pm$2.0} &
        50.50{\tiny$\pm$2.5} & 55.25{\tiny$\pm$2.2} & 37.75{\tiny$\pm$2.1} &
        47.25{\tiny$\pm$2.6} & 41.03{\tiny$\pm$2.3} & 47.60{\tiny$\pm$2.0} \\

      Claude-3.5-sonnet \cite{anthropic2024Claude35sonnet} &
        25.00{\tiny$\pm$2.2} & 20.00{\tiny$\pm$1.8} & 22.50{\tiny$\pm$1.9} &
        19.05{\tiny$\pm$2.0} & 52.38{\tiny$\pm$2.2} & 35.71{\tiny$\pm$1.9} &
        40.00{\tiny$\pm$2.5} & 90.00{\tiny$\pm$1.3} & 65.00{\tiny$\pm$2.0} \\

      Gemini-2.5-flash \cite{comanici2025gemini25} &
        15.00{\tiny$\pm$1.8} & 55.00{\tiny$\pm$2.2} & 35.00{\tiny$\pm$1.9} &
        52.38{\tiny$\pm$2.4} & 61.90{\tiny$\pm$2.0} & 57.14{\tiny$\pm$1.8} &
        40.00{\tiny$\pm$2.5} & 85.00{\tiny$\pm$1.6} & 62.50{\tiny$\pm$1.9} \\

      Gemini-2.5-flash-image \cite{comanici2025gemini25} &
        25.00{\tiny$\pm$2.0} & 40.00{\tiny$\pm$2.1} & 32.50{\tiny$\pm$1.8} &
        42.86{\tiny$\pm$2.4} & 61.90{\tiny$\pm$2.0} & 52.38{\tiny$\pm$1.9} &
        45.00{\tiny$\pm$2.6} & 90.00{\tiny$\pm$1.3} & 67.50{\tiny$\pm$1.8} \\

      \midrule
      \rowcolor{lightgreen}
      \multicolumn{10}{c}{\textbf{Open-source Reasoning Models}} \\

      Qwen3-vl-8b-instruct \cite{qwenteam2025qwen3} &
        25.00{\tiny$\pm$2.0} & 40.00{\tiny$\pm$2.1} & 32.50{\tiny$\pm$1.9} &
        38.10{\tiny$\pm$2.3} & 28.57{\tiny$\pm$2.0} & 33.33{\tiny$\pm$1.8} &
        40.00{\tiny$\pm$2.5} & \underline{\textbf{95.00}}{\tiny$\pm$1.0} & 67.50{\tiny$\pm$1.9} \\

      Qwen-vl-max \cite{Qwenteam2024Qwenvlmax} &
        25.00{\tiny$\pm$2.0} & 30.00{\tiny$\pm$1.9} & 27.50{\tiny$\pm$1.8} &
        52.38{\tiny$\pm$2.4} & 57.14{\tiny$\pm$2.1} & 54.76{\tiny$\pm$1.9} &
        40.00{\tiny$\pm$2.5} & 85.00{\tiny$\pm$1.6} & 62.50{\tiny$\pm$1.9} \\

      Qwen2.5-vl-72b-instruct \cite{Qwen2025Qwen25technicalreport} &
        30.00{\tiny$\pm$2.1} & 30.00{\tiny$\pm$1.9} & 30.00{\tiny$\pm$1.7} &
        47.62{\tiny$\pm$2.4} & 57.14{\tiny$\pm$2.0} & 52.38{\tiny$\pm$1.8} &
        40.00{\tiny$\pm$2.5} & 85.00{\tiny$\pm$1.6} & 62.50{\tiny$\pm$1.9} \\

      Deepseek-vl2 \cite{wu2024Deepseek-vl2} &
        40.00{\tiny$\pm$2.4} & 70.00{\tiny$\pm$2.0} & 55.00{\tiny$\pm$1.8} &
        42.86{\tiny$\pm$2.3} & 23.81{\tiny$\pm$1.9} & 33.33{\tiny$\pm$1.7} &
        10.00{\tiny$\pm$1.3} & 95.00{\tiny$\pm$1.0} & 52.50{\tiny$\pm$2.0} \\

      Deepseek-r1 \cite{Deepseekai2025Deepseekr1} &
        27.55{\tiny$\pm$2.2} & 55.30{\tiny$\pm$2.2} & 48.00{\tiny$\pm$1.9} &
        61.75{\tiny$\pm$2.4} & 61.00{\tiny$\pm$2.0} & 46.50{\tiny$\pm$2.1} &
        59.25{\tiny$\pm$2.5} &  \underline{\textbf{95.00}}{\tiny$\pm$1.0} & 67.50{\tiny$\pm$1.8} \\

      \midrule
      \rowcolor{lightgreen}
      \multicolumn{10}{c}{\textbf{Open-source Chat Models}} \\

      Qwen2.5-vl-32b-instruct \cite{qwenteam2025qwen3} &
        40.00{\tiny$\pm$2.4} & 60.00{\tiny$\pm$2.1} & 50.00{\tiny$\pm$1.9} &
        23.81{\tiny$\pm$2.0} & 47.62{\tiny$\pm$2.1} & 35.71{\tiny$\pm$1.8} &
        40.00{\tiny$\pm$2.5} & 75.00{\tiny$\pm$1.9} & 57.50{\tiny$\pm$2.0} \\

      Llama-3.2-90b-vision-instruct \cite{grattafiori2024llama3herdmodels} &
        55.00{\tiny$\pm$2.5} & 55.00{\tiny$\pm$2.2} & 55.00{\tiny$\pm$1.9} &
        42.86{\tiny$\pm$2.3} & 61.90{\tiny$\pm$2.0} & 52.38{\tiny$\pm$1.8} &
        35.00{\tiny$\pm$2.4} & 95.00{\tiny$\pm$1.0} & 65.00{\tiny$\pm$1.9} \\

      Llama-3.2-11b-vision-instruct \cite{grattafiori2024llama3herdmodels} &
        20.00{\tiny$\pm$1.8} & 50.00{\tiny$\pm$2.2} & 35.00{\tiny$\pm$1.9} &
        33.33{\tiny$\pm$2.2} & 28.57{\tiny$\pm$2.0} & 30.95{\tiny$\pm$1.7} &
        15.00{\tiny$\pm$1.6} & 80.00{\tiny$\pm$1.8} & 47.50{\tiny$\pm$2.0} \\

      Grok-4 \cite{xai2025Grok4} &
        25.00{\tiny$\pm$2.0} & 60.00{\tiny$\pm$2.2} & 42.50{\tiny$\pm$1.9} &
        47.62{\tiny$\pm$2.3} & 28.57{\tiny$\pm$2.0} & 38.10{\tiny$\pm$1.8} &
        \underline{\textbf{75.00}}{\tiny$\pm$2.2} & 80.00{\tiny$\pm$1.6} & 77.50{\tiny$\pm$1.7} \\

      Glm-4.5V \cite{Glmvteam2025Glm45v} &
        35.00{\tiny$\pm$2.2} & 35.00{\tiny$\pm$2.0} & 35.00{\tiny$\pm$1.8} &
        61.90{\tiny$\pm$2.3} & 28.57{\tiny$\pm$2.0} & 45.24{\tiny$\pm$2.0} &
        60.00{\tiny$\pm$2.5} & 75.00{\tiny$\pm$1.9} & 67.50{\tiny$\pm$1.8} \\

      \bottomrule
    \end{tabular}
  }
\end{table*}

\begin{table*}[t] \caption{Video generation physics evaluation. Columns report physics-aware metrics (mean $\pm$ std over $n=5$ runs); the best value in each column is bold.} \label{tab:video_generation_results} \small \centering \resizebox{\textwidth}{!}{ \begin{tabular}{lcccccccc} \toprule \multirow{2}{*}{Model} & \multicolumn{2}{c}{Center of Mass (Video)} & \multicolumn{1}{c}{\shortstack{Lever Equilibrium\\ (Video)}} & \multicolumn{5}{c}{Newton's First Law (Video)} \\ \cmidrule(lr){2-3} \cmidrule(lr){4-4} \cmidrule(lr){5-9} & \shortstack{Seg. Mask\\IoU ↑} & \shortstack{Seg. Mask\\Center ↓} & \shortstack{Final State\\Acc. (\%) ↑ } & \shortstack{Trajectory\\RMSE ↓ } & \shortstack{Final.\\Position Error ↓ } & \shortstack{Speed\\Similarity ↑ } & \shortstack{Acceleration\\Similarity ↑ } & \shortstack{Directional\\Consistency ↑ } \\[-3pt] \midrule Veo3.1 \cite{deepmind_veo_model_card} & \textbf{0.019}{$\pm$0.003}& 108.39{$\pm$12.5}& 35{$\pm$6.2} & 0.384{$\pm$0.012} & 0.198{$\pm$0.009} & -0.011{$\pm$0.008} & -0.021{$\pm$0.007} & 0.5419{$\pm$0.014} \\ Sora-2 \cite{openai_sora2_system_card} & 0.167{$\pm$0.008}& 121.42{$\pm$11.8}& 40{$\pm$6.5} & 0.380{$\pm$0.011} & 0.199{$\pm$0.008} & -0.042{$\pm$0.009} & \textbf{0.017}{$\pm$0.006} & 0.5494{$\pm$0.013} \\ LTX-Video \cite{HaCohen2024LTXVideo} & 0.005{$\pm$0.001}& \textbf{76.37}{$\pm$9.2}& 4.76{$\pm$2.1} & 0.406{$\pm$0.013} & 0.213{$\pm$0.010} & -0.011{$\pm$0.008} & -0.056{$\pm$0.009} & 0.5594{$\pm$0.015} \\ CogVideoX1.5-5B-I2V \cite{yang2024cogvideox} & 0.014{$\pm$0.002}& 223.70{$\pm$18.5}& 38.10{$\pm$6.8} & 0.414{$\pm$0.014} & 0.323{$\pm$0.012} & 0.090{$\pm$0.010} & -0.043{$\pm$0.008} & 0.4884{$\pm$0.016} \\ Pyramid Flow \cite{jin2024pyramidal} & 0.012{$\pm$0.002}& 322.97{$\pm$22.3}& \textbf{47.62}{$\pm$7.2} & 0.381{$\pm$0.011} & 0.276{$\pm$0.011} & -0.047{$\pm$0.009} & -0.019{$\pm$0.007} & \textbf{0.6437}{$\pm$0.012}\\ Wan2.2 14B \cite{wan2025} & 0.136{$\pm$0.007}& 181.39{$\pm$15.2}& 33.3{$\pm$6.0} & 0.395{$\pm$0.012} & \textbf{0.134}{$\pm$0.007} & \textbf{0.014}{$\pm$0.006} & -0.098{$\pm$0.010} & 0.4775{$\pm$0.015} \\ Cosmos-predict2 2B \cite{nvidia2025cosmosworldfoundationmodel} & 0.009{$\pm$0.001}& 217.33{$\pm$17.8}& 42.85{$\pm$6.7} & \textbf{0.350}{$\pm$0.010} & 0.243{$\pm$0.010} & -0.013{$\pm$0.008} & -0.019{$\pm$0.007} & 0.4884{$\pm$0.016} \\ \bottomrule \end{tabular} } \vspace{-3mm} \end{table*}

\begin{figure*}[htb]
  \centering
  \includegraphics[width=\textwidth]{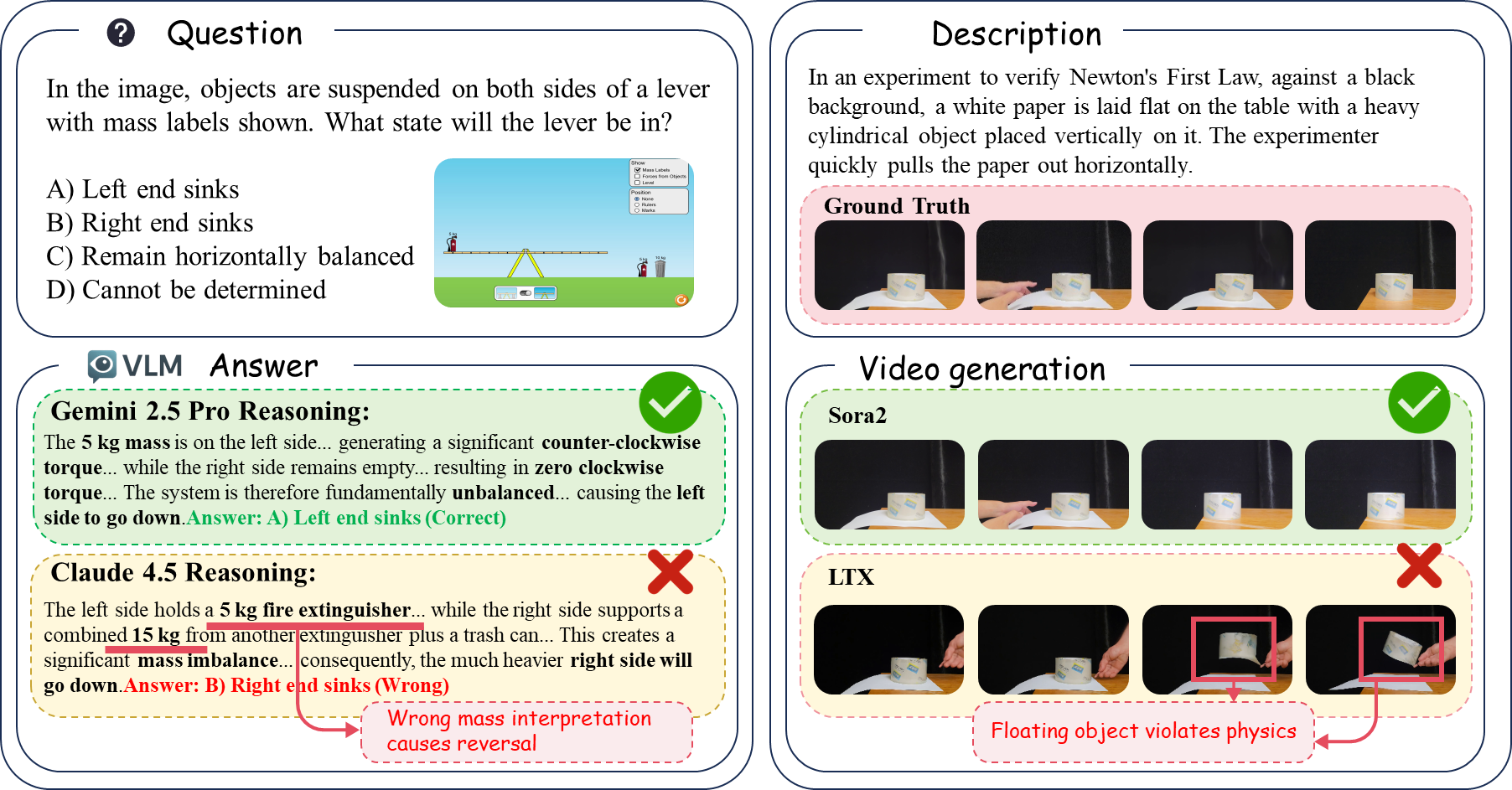}
  \caption{Error analysis of visual reasoning and video generation. Left: Gemini 2.5 Pro correctly predicts lever balance, while Claude 4.5 gives an incorrect prediction. Right: Sora 2 generates physically consistent motion per Newton’s First Law, unlike LTX-Video's generation}
  \label{fig:error}
  \vspace{-3mm}
\end{figure*}

\vspace{-0.1in}
\section{Experiments}

\vspace{-0.05in}
Our study is guided by the following research questions:

\begin{itemize}
    \item \textbf{RQ1. Physical Reasoning in Perception:}  
    Can large multimodal foundation models reason about physical laws from static visual inputs?
    \label{rq1}

    \item \textbf{RQ2. Physical Consistency in Generation:}  
    Do world or video generation models produce motion that remains consistent with core physical principles?
    \label{rq2}

    \item \textbf{RQ3. Sim-to-Real Generalization:}  
    Does this physical common sense learned in simulation generalize to real-world visual data?
    \label{rq3}
\end{itemize}

\subsection{Experimental Setup and Evaluation Protocol}
All models are evaluated on the three PhysicsMind domains, Center of Mass, Lever Equilibrium, and Newton’s First Law, under both perception (VQA) and generation settings. In VQA, a model receives an image and a multiple-choice physics question, and we report accuracy per domain. In video generation, a model is conditioned on an initial frame and a textual description of the setup to produce a short video, and we compute law-aware metrics that assess segmentation overlap and centroid alignment (Center of Mass), final lever state (equilibrium), and trajectory/velocity/acceleration consistency (Newton’s first law). Further implementation details and metric definitions are provided in the Appendix.

\subsection{Evaluation of Physical Common Sense in VQA (\textbf{RQ1})}
\label{sec:vqa_rq1}

Table~\ref{tab:VQA_results} summarizes the performance of multimodal
foundation models on the PhysicsMind VQA benchmark, addressing
\textbf{RQ1}: to what extent contemporary vision-language models
exhibit physical common sense when reasoning about controlled visual
scenarios governed by basic mechanical principles.

\noindent\textbf{Overall Performance Trends.}
Across all models, accuracy spans from 25\% to 85\%, revealing wide disparities in physical reasoning competence. 
Across all models, overall accuracy ranges from roughly random guessing (25\%) up to 85\%, revealing large disparities in physical reasoning competence. No single model dominates across all three domains. GPT-5 achieves the best overall scores on Center of Mass and Lever Equilibrium (70\% overall in both), whereas o4-min attains the highest Newton's First Law accuracy (85\%). This fragmentation suggests specialized rather than generalized physical understanding and is consistent with models relying partly on pattern familiarity instead of robust, law-grounded inference.

\noindent\textbf{Domain-Specific Insights.}
Performance differs markedly by physical domain. Newton's First Law questions obtain the highest mean accuracy (68.5\% on average), likely because they rely on visually salient motion cues and relatively simple binary decisions (move vs.\ remain still), which align well with everyday intuitions and common physical understanding. Center of Mass questions average 47.3\%; these items require implicit localization of equilibrium axes and reasoning about hidden mass distributions, which remain particularly challenging for most models. Lever Equilibrium questions lie in between (mean 52.9\%), indicating partial mastery of torque balance and lever dynamics when textual mass annotations are explicitly provided.

\noindent\textbf{Error Characterization.}
Qualitative inspection (Figure~\ref{fig:error}) reveals two dominant error modes that appear across models and domains. First, models frequently misread fine-grained visual details such as small mass labels or lever markers, leading to incorrect torque comparisons (\emph{visual parsing errors}). Second, they often fail to complete the full reasoning chain from perception to physics computation and textual inference, producing answers that are locally plausible but globally inconsistent with mechanics (\emph{incomplete reasoning}). For example, some models compare left and right object masses but ignore whether the objects are actually placed on the lever, yielding logically invalid torque calculations. These patterns expose a gap between visual perception and multi-step, physically grounded reasoning.

\noindent\textbf{Interpretation.}
Overall, our results indicate that large multimodal models can capture coarse spatial regularities (e.g., heavier objects tend to lower the lever) but still struggle with precise quantitative relationships and counterfactual reasoning. Although top-performing systems substantially outperform random baselines, the persistent domain gaps and systematic error modes suggest that genuine, law-level physical common sense in VQA has yet to emerge. 
PhysicsMind thus serves as a diagnostic probe of current limitations in static and short-horizon visual physics reasoning.

\subsection{Evaluation of Physical Consistency in Video Generation (\textbf{RQ2})}
\label{sec:video_rq2}

Table~\ref{tab:video_generation_results} reports quantitative results for seven recent video generation models across the three PhysicsMind
domains. This section addresses \textbf{RQ2}, asking whether current
world models produce motion that remains consistent with core
mechanical principles.

\begin{table*}[htbp]
\caption{Sim2Real Center-of-Mass results (VQA). Comparison of model accuracies on real and simulated videos (all values in \%). Gap = Sim Overall $-$ Real Overall. Values after $\pm$ represent standard deviation across multiple runs ($n=5$).}
\label{tab:sim2real_vqa}
\centering
\renewcommand{\arraystretch}{1.15}
\resizebox{0.85\linewidth}{!}{%
\begin{tabular}{l|ccc|ccc|c}
\toprule
\multirow{2}{*}{\textbf{Model}} &
\multicolumn{3}{c|}{\textbf{Real Videos}} &
\multicolumn{3}{c|}{\textbf{Simulated Videos}} &
\multirow{2}{*}{\textbf{Gap}} \\
\cmidrule(lr){2-4}\cmidrule(lr){5-7}
 & \textbf{Position Acc. ↑} & \textbf{Rotation Acc. ↑} & \textbf{Overall ↑} 
 & \textbf{Position Acc. ↑} & \textbf{Rotation Acc. ↑} & \textbf{Overall ↑} 
 & \\
\midrule
GPT-5 & 80.00{$\pm$1.2} & 60.00{$\pm$1.5} & 70.00{$\pm$1.3} & 25.00{$\pm$1.5} & 45.00{$\pm$1.8} & 35.00{$\pm$1.5} & \textcolor{deepyellow}{\textbf{$-$35.00}} \\
o4-min & 55.00{$\pm$1.5} & 50.00{$\pm$1.6} & 52.50{$\pm$1.4} & 5.00{$\pm$0.8} & 50.00{$\pm$1.6} & 27.50{$\pm$1.3} & $-$25.00 \\
Deepseek-r1 & 15.00{$\pm$1.1} & 35.00{$\pm$1.5} & 25.00{$\pm$1.2} & 35.00{$\pm$1.8} & 40.00{$\pm$1.8} & 37.50{$\pm$1.5} & \textcolor{deepgreen}{\textbf{12.50}} \\
Claude-sonnet-4-5 & 20.00{$\pm$1.2} & 45.00{$\pm$1.6} & 32.50{$\pm$1.3} & 5.00{$\pm$0.8} & 30.00{$\pm$1.4} & 17.50{$\pm$1.1} & $-$15.00 \\
Gemini-2.5-pro & 70.00{$\pm$1.4} & 30.00{$\pm$1.4} & 50.00{$\pm$1.3} & 25.00{$\pm$1.5} & 40.00{$\pm$1.7} & 32.50{$\pm$1.5} & $-$17.50 \\
Grok-4 & 60.00{$\pm$1.4} & 25.00{$\pm$1.2} & 42.50{$\pm$1.3} & 25.00{$\pm$1.5} & 30.00{$\pm$1.4} & 27.50{$\pm$1.4} & $-$15.00 \\
Qwen3-vl-8b-instruct & 40.00{$\pm$1.5} & 25.00{$\pm$1.2} & 32.50{$\pm$1.3} & 40.00{$\pm$1.8} & 40.00{$\pm$1.8} & 40.00{$\pm$1.5} & 7.50 \\
Deepseek-vl2 & 70.00{$\pm$1.4} & 40.00{$\pm$1.6} & 55.00{$\pm$1.4} & 65.00{$\pm$1.8} & 50.00{$\pm$1.8} & 57.50{$\pm$1.6} & 2.50 \\
Glm-4.5v & 35.00{$\pm$1.4} & 35.00{$\pm$1.4} & 35.00{$\pm$1.3} & 30.00{$\pm$1.3} & 45.00{$\pm$1.7} & 37.50{$\pm$1.5} & 2.50 \\
\bottomrule
\end{tabular}%
}
\vspace{-2mm}
\end{table*}

\textbf{Overall Performance Trends.}
Across all experiments, physical consistency is limited and highly fragmented. No model achieves uniformly strong performance across spatial (Center of Mass), mechanical (torque balance), and dynamic (motion prediction) metrics. For the Center of Mass, models such as Sora-2 and Wan2.2~14B obtain the highest segmentation IoU, whereas Pyramid Flow yields the best lever final-state accuracy (47.62\%). Under Newton’s First Law, Cosmos-predict2~2B achieves the lowest trajectory error (normalized RMSE~=~0.350), yet its directional consistency remains far from perfect. These mixed patterns indicate that existing generators can produce visually plausible videos, but struggle to encode a unified, law-consistent notion of physical dynamics.

\noindent\noindent\textbf{Domain-Specific Insights.}
\emph{Center of Mass.}
All models find it difficult to reproduce accurate object geometry and centroids. Even the strongest systems reach only modest IoU scores (on the order of $10^{-2}$–$10^{-1}$), and centroid errors span from roughly $75$ to over $320$ pixels, suggesting that generators capture coarse placement but lack robust representations of gravitational alignment or equilibrium.

\begin{table*}[htbp]
\caption{Sim2Real center-of-mass leaderboard for video prediction. 
The table compares real and simulated video performance using segmentation-mask IoU and center-distance metrics; Gap (IoU) = Sim IoU $-$ Real IoU and Gap (Center) = Sim Center $-$ Real Center. Values after $\pm$ represent standard deviation across multiple runs ($n=5$).}
\label{tab:sim2real_video}
\centering
\renewcommand{\arraystretch}{1.15}
\setlength{\tabcolsep}{5pt}
\resizebox{0.85\linewidth}{!}{%
\begin{tabular}{l|cc|cc|cc}
\toprule
\multirow{2}{*}{\textbf{Model}} &
\multicolumn{2}{c|}{\textbf{Real Videos}} &
\multicolumn{2}{c|}{\textbf{Simulated Videos}} &
\multicolumn{2}{c}{\textbf{Gap}} \\
\cmidrule(lr){2-3}\cmidrule(lr){4-5}\cmidrule(lr){6-7}
 & \textbf{Segm. Mask IoU ↑} & \textbf{Segm. Mask Center ↓} & 
   \textbf{Segm. Mask IoU ↑} & \textbf{Segm. Mask Center ↓} &
   \textbf{IoU } & \textbf{Center} \\
\midrule
Veo3.1 & 0.0185{$\pm$0.0012} & 108.39{$\pm$5.5} & 0.1517{$\pm$0.0068} & 104.28{$\pm$5.2} & 0.1332 & $-$4.11 \\
Sora-2 & 0.1636{$\pm$0.0065} & 121.42{$\pm$6.2} & 0.0770{$\pm$0.0042} & 99.35{$\pm$5.0} & \textcolor{deepyellow}{\textbf{$-$0.0866}} & $-$22.07 \\
LTX-Video & 0.0050{$\pm$0.0008} & 76.37{$\pm$4.8} & 0.0276{$\pm$0.0022} & 101.98{$\pm$5.1} & 0.0226 & \textcolor{deepgreen}{\textbf{25.61}} \\
CogVideoX1.5-5B-I2V & 0.0140{$\pm$0.0011} & 222.70{$\pm$9.5} & 0.0714{$\pm$0.0045} & 206.17{$\pm$8.8} & 0.0574 & $-$16.53 \\
Pyramid Flow & 0.0120{$\pm$0.0010} & 322.97{$\pm$12.0} & 0.0609{$\pm$0.0040} & 131.45{$\pm$6.5} & 0.0489 & \textcolor{deepyellow}{\textbf{$-$191.52}} \\
Wan2.2 14B & 0.1360{$\pm$0.0060} & 181.39{$\pm$7.8} & 0.1839{$\pm$0.0072} & 90.48{$\pm$4.8} & 0.0479 & $-$90.91 \\
Cosmos-predict2 2B & 0.0090{$\pm$0.0009} & 217.33{$\pm$9.2} & 0.1678{$\pm$0.0070} & 206.17{$\pm$8.5} & \textcolor{deepgreen}{\textbf{0.1588}} & $-$11.16 \\
\bottomrule
\end{tabular}%
}
\vspace{-4mm}
\end{table*}

\emph{Lever Equilibrium.}
Mechanical reasoning around torque balance remains weak. Most models operate near the $50\%$ random baseline (e.g., Sora-2: $40\%$, Pyramid Flow: $47.62\%$), and some fall well below chance (LTX-Video: $4.76\%$). This indicates that lever outcomes are often driven by learned appearance patterns rather than an internalized notion of torque.

\emph{Newton’s First Law.}
Kinematic stability is also limited. While Cosmos-predict2~2B attains the smallest trajectory error, directional consistency values cluster around $0.5$–$0.65$ (where $0.5$ corresponds to random direction alignment), indicating only weak preservation of velocity direction. Combined with sizable speed and acceleration similarity, these results suggest that generated motions frequently deviate from inertia-consistent trajectories.



\noindent\textbf{Error Characterization.}
Qualitative inspection (Figure~\ref{fig:error}, right) highlights systematic temporal reasoning failures. In some cases (e.g., Sora-2), objects remain approximately at rest when the supporting paper is pulled quickly, aligning with Newton’s First Law; in others (e.g., LTX-Video), objects lift or accelerate in implausible ways. Similar to the VQA setting, these errors arise from inadequate object-state tracking and missing causal links between external forces and motion response, with models often treating frames as nearly independent images instead of propagating coherent velocity over time.

\noindent\textbf{Interpretation.} 
Current video generation systems exhibit perceptual realism without reliable physical realism. They rely heavily on visual priors rather than embedded physical simulation, yielding sequences that look convincing but frequently violate balance, torque, and inertia. Bridging this gap will likely require world models that explicitly encode and enforce conservation principles across both space and time, rather than relying solely on large-scale text–image–video pretraining.




\subsection{Sim-to-Real in Physical Reasoning (\textbf{RQ3})}
\label{sec:rq3_sim2real}

To test whether learned physics transfers beyond clean simulators, we use paired center-of-mass scenes with both rendered and visually matched real videos and evaluate models on VQA (Table~\ref{tab:sim2real_vqa}) and video prediction (Table~\ref{tab:sim2real_video}).

\textbf{VQA.}
Closed-source VLMs perform better on real videos than on simulated ones. GPT-5 and o4-min, for example, drop by $25$--$35$ points when moving from real to simulated inputs, and Claude~4.5, Gemini~2.5~Pro, and Grok-4 show similar negative Sim--Real gaps. This is consistent with pretraining on web imagery that is closer to real tabletop scenes than to stylized renders. Open-source models show much smaller or even positive gaps: Qwen3-VL-8B, Deepseek-VL2, and Glm-4.5V have Sim--Real gaps between $+2.50$ and $+7.50$, and Deepseek-R1 performs noticeably better on simulation ($+12.50$), suggesting a stronger reliance on the clean edges and simple backgrounds of synthetic data.

\noindent\textbf{Video generation.}
For video prediction, synthetic worlds are systematically easier. Most generators obtain higher segmentation IoU and lower centroid error on simulated videos than on real ones (positive IoU gaps for Veo3.1, LTX-Video, CogVideoX, Pyramid Flow, Wan2.2, Cosmos-predict2; large negative centroid gaps for Pyramid Flow and Wan2.2). Sora-2 is the only model whose IoU slightly favors real videos, but its centroid accuracy still degrades when moving from renders to real footage. These trends indicate that current world models overfit to simplified synthetic geometry and struggle with clutter, lighting variation, and ambiguous boundaries in real scenes.

\textbf{\noindent}
Across perception and generation, PhysicsMind points to a sim-to-real gap in physical understanding. Closed-source VLMs tend to be stronger on real videos than on simulations, while video generators often show the opposite trend. This asymmetry suggests that models may rely more on visual priors than on domain-invariant notions of mass, balance, and motion, highlighting physics-aware training on both simulated and real videos as a promising direction.

\section{Conclusion and Limitations}
\label{Conclusion}

PhysicsMind offers a controlled testbed for examining how modern Foundation and world models connect perception, reasoning, and prediction to concrete mechanics laws, rather than to visual plausibility alone. Focusing on three textbook scenarios with paired real–simulated videos, we find that both vision–language and video generation models still rely heavily on superficial cues and frequently violate basic constraints on balance, torque, and inertia, indicating that robust physical understanding has yet to emerge.

The benchmark is necessarily limited: PhysicsMind currently focuses on rigid-body setups, short time horizons, and off-the-shelf models. Extending it to richer phenomena (e.g., friction, collisions, fluids), longer and interactive scenarios, and to guide training or design of physics-aware architectures is left for future work.

\clearpage
\newpage

{
    \small
    \bibliographystyle{ieeenat_fullname}
    \bibliography{main}

@String(CVPR= {IEEE Conf. Comput. Vis. Pattern Recog.})

@String(ICLR = {Int. Conf. Learn. Represent.})

@String(CVPR  = {CVPR})

@String(ICLR  = {ICLR})

@inproceedings{johnson2017clevr,
  title={CLEVR: A diagnostic dataset for compositional language and elementary visual reasoning},
  author={Johnson, Justin and Hariharan, Bharath and van der Maaten, Laurens and Hoffman, Judy and Fei-Fei, Li and Zitnick, C Lawrence and Girshick, Ross},
  booktitle={CVPR},
  pages={2901--2910},
  year={2017}
}

@inproceedings{riochet2018intphys,
  title={IntPhys: A benchmark for visual intuitive physics reasoning},
  author={Riochet, Ronan and Castro, Mario Ynocente and Bernard, Mathieu and Lerer, Adam and Fergus, Rob and Izard, Véronique and Dupoux, Emmanuel},
  booktitle={NeurIPS},
  year={2018}
}

@inproceedings{bakhtin2019phyre,
  title={PHYRE: A new benchmark for physical reasoning},
  author={Bakhtin, Anton and van der Maaten, Laurens and Johnson, Justin and Gustafson, Laura and Girshick, Ross},
  booktitle={NeurIPS},
  year={2019}
}

@inproceedings{smith2022craft,
  title={CRAFT: A benchmark for causal reasoning about forces and interactions},
  author={Ates, Tayfun and Atesoglu, M. Samil and Yigit, Cagatay and Kesen, Ilker and Kobas, Mert and others},
  booktitle={Findings of ACL},
  year={2022}
}

@inproceedings{li2024vbench,
  title={VBench: Comprehensive benchmark suite for video generative models},
  author={Huang, Ziqi and He, Yinan and Yu, Jiashuo and Zhang, Fan and Si, Chenyang and others},
  booktitle={arXiv preprint arXiv:2401.12915},
  year={2024}
}

@inproceedings{hafner2019planet,
  title={Learning latent dynamics for planning from pixels},
  author={Hafner, Danijar and Lillicrap, Timothy and Norouzi, Mohammad and Ba, Jimmy},
  booktitle={ICML},
  year={2019}
}

@inproceedings{hafner2019dreamer,
  title={Dream to Control: Learning Behaviors by Latent Imagination},
  author={Hafner, Danijar and Lillicrap, Timothy and Norouzi, Mohammad and Ba, Jimmy},
  booktitle={ICLR},
  year={2020}
}

@inproceedings{morpheus2025,
  title={MORPHEUS: Benchmarking Physical Reasoning of Video Generative Models with Real Experiments},
  author={Zhang, Chenyu and Cherniavskii, Daniil and Tragoudaras, Antonios and Vozikis, Antonios and Nijdam, Thijmen and others},
  booktitle={arXiv preprint arXiv:2504.02918},
  year={2025}
}

@inproceedings{worldmodelbench2025,
  title={WorldModelBench: Judging Video Generation Models as World Models},
  author={Li, Dacheng and Fang, Yunhao and Chen, Yukang and Yang, Shuo and Cao, Shiyi and others},
  booktitle={arXiv preprint arXiv:2502.20694},
  year={2025}
}

@inproceedings{worldscore2025,
  title={WorldScore: A Unified Evaluation Benchmark for World Generation},
  author={Duan, Haoyi and Yu, Hong-Xing and Chen, Sirui and Fei-Fei, Li and Wu, Jiajun},
  booktitle={arXiv preprint arXiv:2504.00983},
  year={2025}
}

@misc{openai2024gpt4technicalreport,
      title={GPT-4 Technical Report}, 
      author={OpenAI and others},
      year={2024},
      eprint={2303.08774},
      archivePrefix={arXiv},
      primaryClass={cs.CL},
      url={https://arxiv.org/abs/2303.08774}, 
}

@misc{openai2024gpt4ocard,
      title={GPT-4o System Card}, 
      author={OpenAI and others},
      year={2024},
      eprint={2410.21276},
      archivePrefix={arXiv},
      primaryClass={cs.CL},
      url={https://arxiv.org/abs/2410.21276}, 
}

@misc{anthropic2024claude35sonnet,
  title={Claude Sonnet 3.5},
  author={Anthropic},
  year={2024},
  month={June},
  url={https://www.anthropic.com/news/claude-3-5-sonnet}    
}

@misc{anthropic2025claude37sonnet,
  title={Claude Sonnet 3.7},
  author={Anthropic},
  year={2025},
  month={February},
  url={https://www.anthropic.com/claude}
}

@misc{qwenteam2024qwenvlmax,
  title={Qwen-VL-Max},
  author={Qwen Team},
  year={2024},
  url={https://qwenlm.github.io/}
}

@misc{qwen2025qwen25technicalreport,
      title={Qwen2.5 Technical Report}, 
      author={Qwen and others},
      year={2025},
      eprint={2412.15115},
      archivePrefix={arXiv},
      primaryClass={cs.CL},
      url={https://arxiv.org/abs/2412.15115}, 
}

@misc{grattafiori2024llama3herdmodels,
      title={The Llama 3 Herd of Models}, 
      author={Aaron Grattafiori and others},
      year={2024},
      eprint={2407.21783},
      archivePrefix={arXiv},
      primaryClass={cs.AI},
      url={https://arxiv.org/abs/2407.21783}, 
}

@misc{OpenAI2025GPT5SystemCard,
  author      = {{OpenAI}},
  title       = {{GPT-5 System Card}},
  institution = {OpenAI},
  year        = {2025},
  month       = aug,
  day         = {13},
  url         = {https://cdn.openai.com/gpt-5-system-card.pdf},
  note        = {Accessed: 2026-01-09}
}

@misc{OpenAI2025O3O4MiniSystemCard,
  author       = {{OpenAI}},
  title        = {{OpenAI o3 and o4-mini System Card}},
  year         = {2025},
  month        = apr,
  day          = {16},
  url          = {https://openai.com/index/o3-o4-mini-system-card/},
  note         = {Accessed: 2026-01-09},
  organization = {OpenAI}
}

@misc{anthropic2025claudesonnet45,
  title={Claude Sonnet 4.5},
  author={Anthropic},
  year={2025},
  month={September},
  url={https://www.anthropic.com/claude}
}

@article{comanici2025gemini25,
  title={Gemini 2.5: Pushing the Frontier with Advanced Reasoning, Multimodality, Long Context, and Next Generation Agentic Capabilities},
  author={Comanici, Gheorghe and others},
  journal={arXiv preprint arXiv:2507.06261},
  year={2025},
  eprint={2507.06261},
  archivePrefix={arXiv},
  primaryClass={cs.CL},
  url={https://arxiv.org/abs/2507.06261}
}

@misc{xai2025grok4,
  title={Grok-4: The Most Intelligent Model in the World},
  author={xAI},
  year={2025},
  month={July},
  url={https://x.ai/grok-4}
}

@misc{qwenteam2025qwen3,
  title={Qwen3 Technical Report},
  author={Qwen Team},
  year={2025},
  eprint={2505.09388},
  archivePrefix={arXiv},
  primaryClass={cs.CL},
  url={https://arxiv.org/abs/2505.09388}
}

@article{wu2024deepseek-vl2,
  title={DeepSeek-VL2: Mixture-of-Experts Vision-Language Models for Advanced Multimodal Understanding},
  author={Wu, Zhiyu and Chen, Xiaokang and Pan, Zizheng and Liu, Xingchao and Liu, Wen and Dai, Damai and others},
  journal={arXiv preprint arXiv:2412.10302},
  year={2024},
  eprint={2412.10302},
  archivePrefix={arXiv},
  primaryClass={cs.CV},
  url={https://arxiv.org/abs/2412.10302}
}

@misc{glmvteam2025glm45v,
  title={GLM-4.5V and GLM-4.1V-Thinking: Towards Versatile Multimodal Reasoning with Scalable Reinforcement Learning},
  author={GLM-V Team and others},
  year={2025},
  eprint={2507.01006},
  archivePrefix={arXiv},
  primaryClass={cs.CV},
  url={https://arxiv.org/abs/2507.01006}
}

@article{deepseekai2025deepseekr1,
  title={DeepSeek-R1: Incentivizing Reasoning Capability in LLMs via Reinforcement Learning},
  author={DeepSeek-AI and others},
  journal={arXiv preprint arXiv:2501.12948},
  year={2025},
  eprint={2501.12948},
  archivePrefix={arXiv},
  primaryClass={cs.CL},
  url={https://arxiv.org/abs/2501.12948}
}

@misc{deepmind_veo_model_card,
  title        = {Veo: Google DeepMind Video Generation Model Card},
  author       = {{Google DeepMind}},
  howpublished = {Model card},
  year         = {2025},
  url          = {https://deepmind.google/models/veo/},
  note         = {Accessed: 2025-11-10}
}

@misc{openai_sora2_system_card,
  title        = {Sora 2 System Card},
  author       = {{OpenAI}},
  howpublished = {System card},
  year         = {2025},
  url          = {https://openai.com/index/sora-2-system-card/},
  note         = {Accessed: 2025-11-10}
}

@article{HaCohen2024LTXVideo,
  title={LTX-Video: Realtime Video Latent Diffusion},
  author={HaCohen, Yoav and Chiprut, Nisan and Brazowski, Benny and Shalem, Daniel and Moshe, Dudu and Richardson, Eitan and others},
  journal={arXiv preprint arXiv:2501.00103},
  year={2024}
}

@article{yang2024cogvideox,
  title={CogVideoX: Text-to-Video Diffusion Models with An Expert Transformer},
  author={Yang, Zhuoyi and Teng, Jiayan and Zheng, Wendi and Ding, Ming and Huang, Shiyu and Xu, Jiazheng and others},
  journal={arXiv preprint arXiv:2408.06072},
  year={2024}
}

@article{jin2024pyramidal,
  title={Pyramidal Flow Matching for Efficient Video Generative Modeling},
  author={Jin, Yang and Sun, Zhicheng and Li, Ningyuan and Xu, Kun and Jiang, Hao and Zhuang, Nan and Huang, Quzhe and others},
  journal={arXiv preprint arXiv:2410.05954},
  year={2024}
}

@article{wan2025,
      title={Wan: Open and Advanced Large-Scale Video Generative Models}, 
      author={Team Wan},
      journal = {arXiv preprint arXiv:2503.20314},
      year={2025}
}

@misc{nvidia2025cosmosworldfoundationmodel,
      title={Cosmos World Foundation Model Platform for Physical AI}, 
      author={Niket Agarwal and others},
      year={2025},
      eprint={2501.03575},
      archivePrefix={arXiv},
      primaryClass={cs.CV},
      url={https://arxiv.org/abs/2501.03575}, 
}

@article{phygenbench2024,
  title={Towards World Simulator: Crafting Physical Commonsense-Based Benchmark for Video Generation},
  author={Meng, Fanqing and Liao, Jiaqi and Tan, Xinyu and Shao, Wenqi and Lu, Quanfeng and others},
  journal={arXiv preprint arXiv:2410.05363},
  year={2024},
  eprint={2410.05363},
  archivePrefix={arXiv},
  primaryClass={cs.CV},
  url={https://arxiv.org/abs/2410.05363}
}

@article{phybench2025,
  title={PhysBench: Benchmarking and Enhancing Vision-Language Models for Physical World Understanding},
  author={Chow, Wei and Mao, Jiageng and Li, Boyi and Seita, Daniel and Guizilini, Vitor and Wang, Yue},
  journal={arXiv preprint arXiv:2501.16411},
  year={2025},
  eprint={2501.16411},
  archivePrefix={arXiv},
  primaryClass={cs.CV},
  url={https://arxiv.org/abs/2501.16411}
}

@misc{foss2025causalvqaphysicallygroundedcausal,
      title={CausalVQA: A Physically Grounded Causal Reasoning Benchmark for Video Models}, 
      author={Aaron Foss and Chloe Evans and Sasha Mitts and Koustuv Sinha and Ammar Rizvi and Justine T. Kao},
      year={2025},
      eprint={2506.09943},
      archivePrefix={arXiv},
      primaryClass={cs.CV},
      url={https://arxiv.org/abs/2506.09943}, 
}

@misc{krojer2025shortcutawarevideoqabenchmarkphysical,
      title={A Shortcut-aware Video-QA Benchmark for Physical Understanding via Minimal Video Pairs}, 
      author={Benno Krojer and Mojtaba Komeili and Candace Ross and Quentin Garrido and Koustuv Sinha and others},
      year={2025},
      eprint={2506.09987},
      archivePrefix={arXiv},
      primaryClass={cs.CV},
      url={https://arxiv.org/abs/2506.09987}, 
}

@article{wang2025physunibench,
  title={PhysUniBench: An Undergraduate-Level Physics Reasoning Benchmark for Multimodal Models},
  author={Wang, Lintao and Su, Encheng and Liu, Jiaqi and Li, Pengze and Xia, Peng and others},
  journal={arXiv preprint arXiv:2506.17667},
  year={2025},
  eprint={2506.17667},
  archivePrefix={arXiv},
  primaryClass={cs.CV},
  url={https://arxiv.org/abs/2506.17667}
}

@article{lin2025physicscognition,
  title={Exploring the Evolution of Physics Cognition in Video Generation: A Survey},
  author={Lin, Minghui and Wang, Xiang and Wang, Yishan and Wang, Shu and Dai, Fengqi and others},
  journal={arXiv preprint arXiv:2503.21765},
  year={2025},
  eprint={2503.21765},
  archivePrefix={arXiv},
  primaryClass={cs.CV},
  url={https://arxiv.org/abs/2503.21765}
}

@article{chen2025worldprediction,
  title={WorldPrediction: A Benchmark for High-level World Modeling and Long-horizon Procedural Planning},
  author={Chen, Delong and Chung, Willy and Bang, Yejin and Ji, Ziwei and Fung, Pascale},
  journal={arXiv preprint arXiv:2506.04363},
  year={2025},
  eprint={2506.04363},
  archivePrefix={arXiv},
  primaryClass={cs.CV},
  url={https://arxiv.org/abs/2506.04363}
}

@article{warrier2024benchmarkingworld,
  title={Benchmarking World-Model Learning},
  author={Warrier, Archana and Nguyen, Dat and Naim, Michelangelo and Jain, Moksh and Liang, Yichao and others},
  journal={arXiv preprint arXiv:2510.19788},
  year={2024},
  eprint={2510.19788},
  archivePrefix={arXiv},
  primaryClass={cs.AI},
  url={https://arxiv.org/abs/2510.19788}
}

@article{zhang2025worldlm,
  title={Can World Models Benefit VLMs for World Dynamics?},
  author={Zhang, Kevin and Ge, Kuangzhi and Chi, Xiaowei and Zhang, Renrui and Shi, Shaojun and Dong, Zhen and Han, Sirui and Zhang, Shanghang},
  journal={arXiv preprint arXiv:2510.00855},
  year={2025},
  eprint={2510.00855},
  archivePrefix={arXiv},
  primaryClass={cs.CV},
  url={https://arxiv.org/abs/2510.00855}
}

@article{chi2025wow,
  title={WoW: Towards a World omniscient World model Through Embodied Interaction},
  author={Chi, Xiaowei and Jia, Peidong and Fan, Chun-Kai and Ju, Xiaozhu and Mi, Weishi and Zhang, Kevin and others},
  journal={arXiv preprint arXiv:2509.22642},
  year={2025},
  eprint={2509.22642},
  archivePrefix={arXiv},
  primaryClass={cs.RO},
  url={https://arxiv.org/abs/2509.22642}
}

@article{chi2024eva,
  title={EVA: An Embodied World Model for Future Video Anticipation},
  author={Chi, Xiaowei and Fan, Chun-Kai and Zhang, Hengyuan and Qi, Xingqun and Zhang, Rongyu and others},
  journal={arXiv preprint arXiv:2410.15461},
  year={2024},
  eprint={2410.15461},
  archivePrefix={arXiv},
  primaryClass={cs.CV},
  url={https://arxiv.org/abs/2410.15461}
}

@article{he2024llmsmultimodal,
  title={LLMs Meet Multimodal Generation and Editing: A Survey},
  author={He, Yingqing and Liu, Zhaoyang and Chen, Jingye and Tian, Zeyue and Liu, Hongyu and others},
  journal={arXiv preprint arXiv:2405.19334},
  year={2024},
  eprint={2405.19334},
  archivePrefix={arXiv},
  primaryClass={cs.AI},
  url={https://arxiv.org/abs/2405.19334}
}

@article{motamed2025physicsiq,
  title={Do generative video models understand physical principles?},
  author={Motamed, Saman and Culp, Laura and Swersky, Kevin and Jaini, Priyank and Geirhos, Robert},
  journal={arXiv preprint arXiv:2501.09038},
  year={2025},
  eprint={2501.09038},
  archivePrefix={arXiv},
  primaryClass={cs.CV},
  url={https://arxiv.org/abs/2501.09038}
}

@article{gundawar2025pacbench,
  title={PAC Bench: Do Foundation Models Understand Prerequisites for Executing Manipulation Policies?},
  author={Gundawar, Atharva and Sagar, Som and Senanayake, Ransalu},
  journal={arXiv preprint arXiv:2506.23725},
  year={2025},
  eprint={2506.23725},
  archivePrefix={arXiv},
  primaryClass={cs.RO},
  url={https://arxiv.org/abs/2506.23725}
}

@article{chen2025adaptvis,
  title={Why Is Spatial Reasoning Hard for VLMs? An Attention Mechanism Perspective on Focus Areas},
  author={Chen, Shiqi and Zhu, Tongyao and Zhou, Ruochen and Zhang, Jinghan and Gao, Siyang and others},
  journal={arXiv preprint arXiv:2503.01773},
  year={2025},
  eprint={2503.01773},
  archivePrefix={arXiv},
  primaryClass={cs.CV},
  url={https://arxiv.org/abs/2503.01773}
}
}

\clearpage
\newpage
\onecolumn
\appendix
\setlength{\parskip}{0.6em}
\setlength{\parindent}{0pt}
\begin{center}
  \textbf{\Large Supplemental Material of PhysicsMind: Sim and Real Mechanics Benchmarking for Physical Reasoning and Prediction in Foundational VLMs and World Models}
\end{center}

\begingroup
  \renewcommand{\contentsname}{}   
  \setcounter{tocdepth}{2}         
  \makeatletter
  \newcommand*\ToCMinPage{12} 
  \let\orig@contentsline\contentsline
  \@ifpackageloaded{hyperref}{%
    \def\contentsline#1#2#3#4{%
      \begingroup
        \edef\pgtmp{#3}%
        \expandafter\endgroup
        \ifnum\pgtmp<\ToCMinPage\relax
        \else
          \orig@contentsline{#1}{#2}{#3}{#4}%
        \fi
    }%
  }{%
    \def\contentsline#1#2#3{%
      \begingroup
        \edef\pgtmp{#3}%
        \expandafter\endgroup
        \ifnum\pgtmp<\ToCMinPage\relax
        \else
          \orig@contentsline{#1}{#2}{#3}%
        \fi
    }%
  }%
  \makeatother
  \tableofcontents
\endgroup

\vspace{2em}

\newpage

\section{Data Curation}
\label{sec:data_curation}


This section outlines the construction of the PhysicsMind benchmark and summarizes its data composition, categories, and statistics. We highlight the dataset's breadth and controlled design, which support reliable evaluation of physical reasoning in multimodal models.

\subsection{Dataset Construction}

\vspace{-0.05in}

PhysicsMind's dataset is built through a structured pipeline that ensures
clear physical interpretability, controlled variation, and consistency across
real and simulated settings.

\begin{itemize}[leftmargin=*]
    \item \textbf{Physics Textbook Identification.}
    Core mechanics concepts were selected from high‑school physics, focusing on laws that are visually interpretable in short clips and experimentally reproducible (Center of Mass, Lever Equilibrium, Newton's First Law).
    \item \textbf{Concept Formulation.}
    Each concept was translated into a standardized scene design with stable camera viewpoints, clear object layouts, and well-defined physical outcomes. This ensured that all scenarios remained unambiguous and visually grounded.
    \item \textbf{Variable Selection.}
    Primary variables were chosen based on direct influence on the physical law. Their ranges were restricted to maintain visual clarity. Secondary variables (colors, minor background changes) were varied within controlled bounds.
    \item \textbf{Real‑World Recording and Simulation.}
    Real videos were captured using a fixed 4K camera on a rigid overhead tripod with controlled lighting. Simulated videos were produced using a deterministic 2D physics engine with fixed parameters. All real setups were first prototyped in simulation to ensure matching dimensions and feasible dynamics.
    \item \textbf{Raw Data Generation.}
    Both simulated and real-world experiments produced short (\( \leq 10 \) s),
    high-resolution clips. We intentionally over-generated raw data to allow
    filtering during the quality-control stage.
\end{itemize}

\subsection{Dataset Annotation and Quality Control}

\vspace{-0.05in}

Annotation and QC follow a multi‑stage process combining expert‑written templates, LLM assistance, and strict manual verification.

\begin{itemize}[leftmargin=*]
    \item \textbf{Raw Data Intake.}
    Only videos with stable framing, clear object visibility, and physically plausible motion were admitted. Clips with camera shake, artifacts, or unexpected behavior were removed.
    \item \textbf{Expert Question Design.}
    3 domain experts created structured templates with slots for physical variables and visual references. This ensured consistent phrasing and parallel coverage across similar configurations.
    \item \textbf{LLM‑Generated Draft Answers.}
    LLMs produced first‑draft answers for each template. These drafts served as placeholders and were not directly used without revision.
    \item \textbf{Expert Review and Validation.}
    Experts reviewed all entries for alignment with the video, clarity, and physics correctness. Automated checks verified object positions and motion consistency. Any example exhibiting ambiguity or annotation mismatch was excluded.
\end{itemize}

\subsection{Overall Statistics}

PhysicsMind comprises a diverse collection of videos and question-answer pairs across three canonical mechanics domains: Center of Mass (CoM), Lever Equilibrium (LE), and Newton's First Law (NI). We leverage both simulated and real-world data to provide varied testing environments. Table \ref{tab:overall_stats} summarizes the overall data statistics for each domain, distinguishing between simulated and real-world videos, as well as the number of VQA pairs and Video Generation (VG) samples.

\begin{table}[h]
\caption{
\textbf{PhysicsMind Statistics.}
Breakdown of real and simulated videos, VQA pairs, 
and video generation (VG) samples across all domains.
}
\label{tab:overall_stats}
\centering
\small
\setlength{\tabcolsep}{6pt}
\renewcommand{\arraystretch}{1.1}
\begin{tabular}{lccccc}
\toprule
\textbf{Domain} &
\textbf{\# Real Videos} &
\textbf{\# Sim Videos} &
\textbf{\# Total Videos} &
\textbf{\# VQA Pairs} &
\textbf{\# VG Samples} \\
\midrule
Center of Mass (CoM)    & 20 & 20 & 40 & 40 & 40 \\
Lever Equilibrium (LE)  & 0  & 21 & 21 & 42 & 21 \\
Newton's First Law (NI) & 20 & 0  & 20 & 40 & 20 \\
\midrule
\textbf{Total} & \textbf{40} & \textbf{41} & \textbf{81} & \textbf{122} & \textbf{81} \\
\bottomrule
\end{tabular}
\end{table}

The dataset features a single core variant for each physical configuration type across all domains. This design ensures consistent testing scenarios while focusing on the model's ability to generalize across different data modalities (simulated vs. real) and tasks (VQA vs. VG).

\subsection{Categories and Subtypes}
\label{sec:categories_subtypes}

PhysicsMind categorizes VQA questions and Video Generation (VG) prompts into 
specific subtypes within each physical domain. This fine-grained categorization 
allows for a detailed analysis of model performance across different aspects of 
physical reasoning. We define distinct subtypes that probe diverse reasoning skills, 
as summarized in Table~\ref{tab:question_subtypes}.

\begin{table}[h]
\caption{\textbf{Question subtypes within PhysicsMind domains.}
Each domain contains distinct reasoning subtypes that capture different 
dimensions of physical understanding for Video Question Answering (VQA) 
and Video Generation (VG) tasks.}
\label{tab:question_subtypes}
\centering
\small
\renewcommand{\arraystretch}{1.15}
\setlength{\tabcolsep}{3.8pt} 
\begin{tabular}{p{0.18\linewidth}|p{0.22\linewidth}|p{0.50\linewidth}}
\toprule
\textbf{Domain} & \textbf{Subtype} & \textbf{Example Reasoning Task} \\
\midrule
\multirow{2}{*}{Center of Mass (CoM)} 
& Position 
& Determine the location of an object's Center of Mass relative to a suspension point. \\
\cline{2-3}
& Rotation 
& Predict the rotational direction of an object around its suspension point when released. \\
\midrule
\multirow{2}{*}{Lever Equilibrium (LE)} 
& Equilibrium 
& Predict the final balanced state of a lever based on masses and distances from the fulcrum. \\
\cline{2-3}
& Balance Adjustment 
& Determine the lever's state after adjusting the position of an object along the lever arm. \\
\midrule
\multirow{2}{*}{Newton's First Law (NI)} 
& Object Position 
& Predict the object's final position after its supporting surface is removed. \\
\cline{2-3}
& Object Stability 
& Determine whether an object tips or remains stable when the support is removed. \\
\bottomrule
\end{tabular}
\end{table}

\subsection{Dataset Breadth and Control}

\begin{figure*}[htbp]
    \centering
    \includegraphics[width=0.95\textwidth]{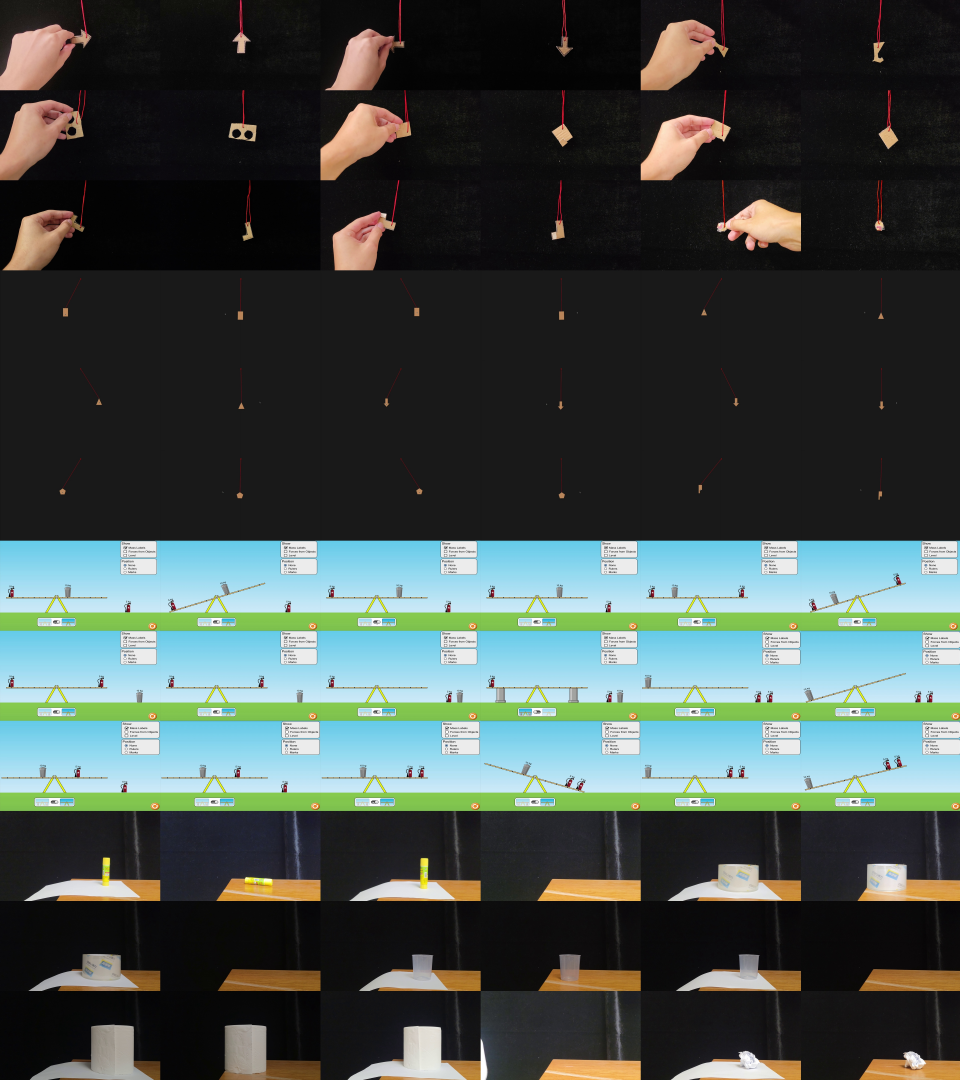}
    \caption{Visual overview of dataset breadth and variation across real and simulated settings.}
    \label{fig:dataset_variants}
\end{figure*}

To demonstrate the range and systematic control present in our dataset, Figure \ref{fig:dataset_variants} provides a visual summary. The illustration showcases how PhysicsMind incorporates variations in object attributes and environmental dynamics across both real-world experiments and physics-based simulations. This structured diversity enables a comprehensive and controllable benchmark for evaluating models of physical reasoning.

\clearpage
\subsection{VQA Question Examples.}

To further clarify the nature of reasoning tasks, we provide specific examples of VQA questions for each subtype. These illustrate the direct queries posed to models.

\begin{textcolorbox}[VQA Prompt Example of Center of Mass (CoM)]
\begin{center}
    \includegraphics[height=1.1in,width=0.95\linewidth,keepaspectratio]{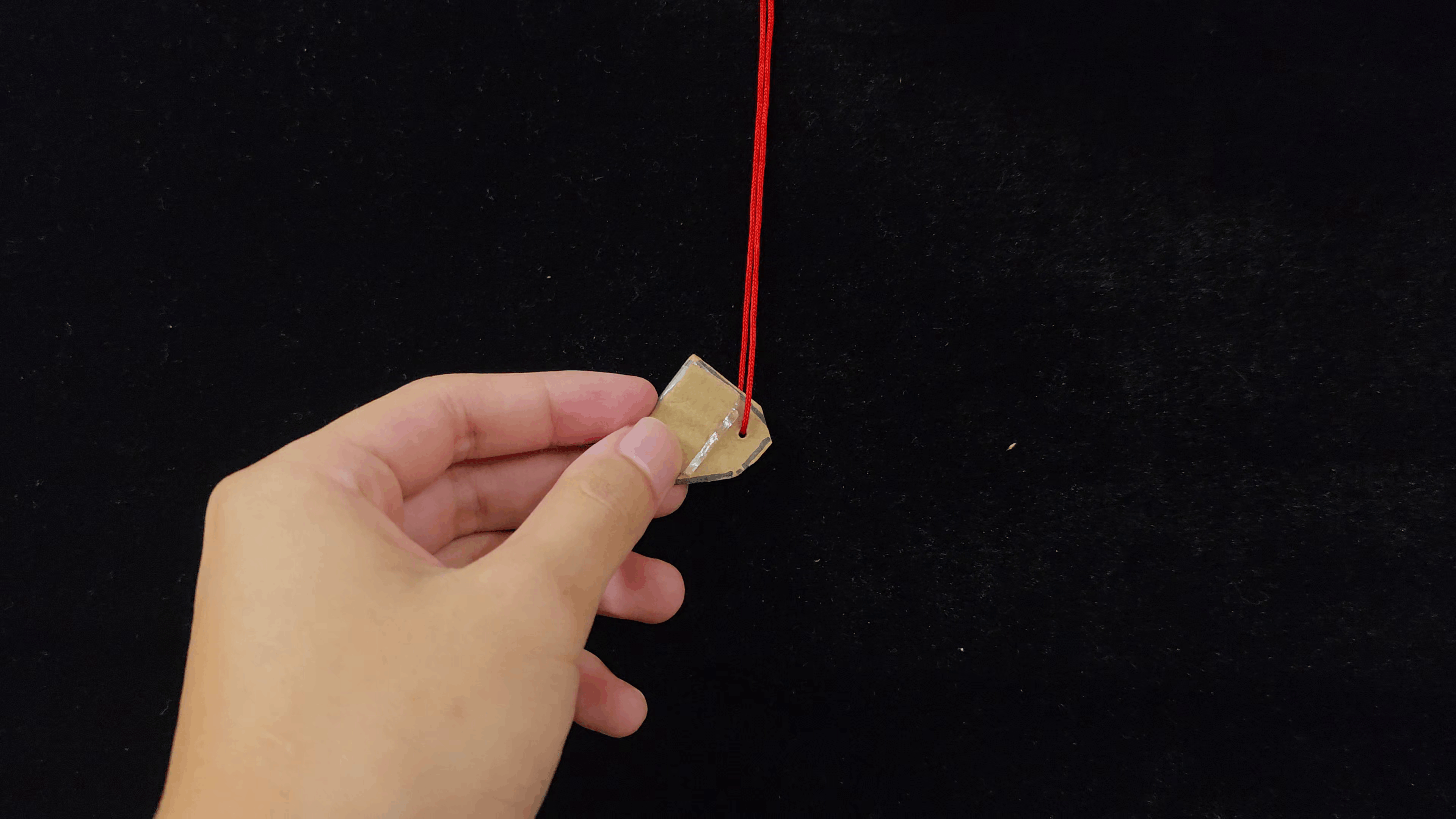}

    \vspace{4pt}
    \small Example image illustrating the CoM scenario.
\end{center}

\begin{itemize}
    \item \textit{Position:} “In the image, where is the hand‑held (suspension) 
    point located relative to the object's center?”

    \vspace{4pt}
    \begin{tabular}{@{}ll}
        \textbf{A)} & To the left of the center \\
        \textbf{\textcolor{cvprblue}{B)}} & \textcolor{cvprblue}{To the right of the center (Correct)} \\
        \textbf{C)} & Above the center \\
        \textbf{D)} & At the center / No clear offset \\
    \end{tabular}

    \vspace{8pt}
    \item \textit{Rotation:} “The small irregular block is suspended by a vertical 
    red string passing through a hole near its upper edge (the suspension point). 
    The hand pinches the block away from that point. When released, around which 
    direction will the block rotate?”

    \vspace{4pt}
    \begin{tabular}{@{}ll}
        \textbf{A)} & Clockwise rotation \\
        \textbf{\textcolor{cvprblue}{B)}} & \textcolor{cvprblue}{Anticlockwise (counter‑clockwise) rotation (Correct)} \\
        \textbf{C)} & Remain stationary (no rotation) \\
        \textbf{D)} & Oscillate without net rotation \\
    \end{tabular}
\end{itemize}
\end{textcolorbox}
\begin{textcolorbox}[VQA Prompt Example of Lever Equilibrium (LE)]
\begin{center}
    \includegraphics[height=1.1in,width=0.95\linewidth,keepaspectratio]{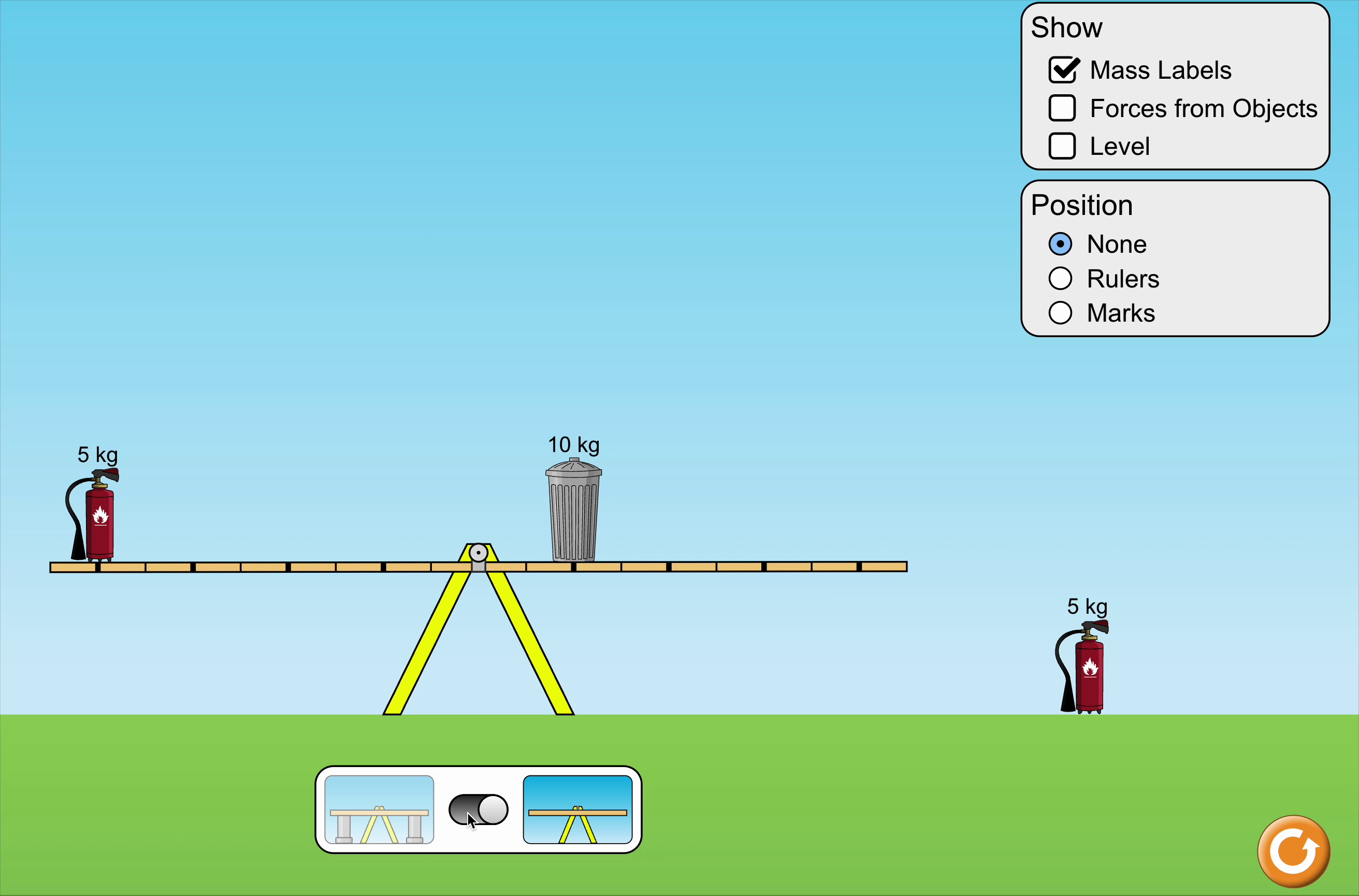}

    \vspace{4pt}
    \small Example image illustrating the Lever Equilibrium scenario.
\end{center}

\begin{itemize}
    \item \textit{Equilibrium:} “In the image, a lever rests on a central fulcrum 
    with objects hung on both sides. Using the displayed mass labels and distances 
    from the fulcrum, determine the lever's final state after release.”

    \vspace{4pt}
    \begin{tabular}{@{}ll}
        \textbf{A)} & Remain horizontally balanced \\
        \textbf{\textcolor{cvprblue}{B)}} & \textcolor{cvprblue}{ Left end sinks, right end rises (Correct)} \\
        \textbf{C)} & Right end sinks, left end rises \\
        \textbf{D)} & Cannot be determined \\
    \end{tabular}

    \vspace{8pt}
    \item \textit{Balance Adjustment:} “If the objects on the left side are moved 
    one position closer to the fulcrum, what new state will the lever reach?”

    \vspace{4pt}
    \begin{tabular}{@{}ll}
        \textbf{A)} & Left end sinks \\
        \textbf{\textcolor{cvprblue}{B)}} & \textcolor{cvprblue}{ Right end sinks (Correct)} \\
        \textbf{C)} & Remain horizontally balanced \\
        \textbf{D)} & Cannot be determined \\
    \end{tabular}
\end{itemize}
\end{textcolorbox}
\begin{textcolorbox}[VQA Prompt Example of Newton’s First Law (NI)]
\begin{center}
    \includegraphics[height=1.1in,width=0.95\linewidth,keepaspectratio]{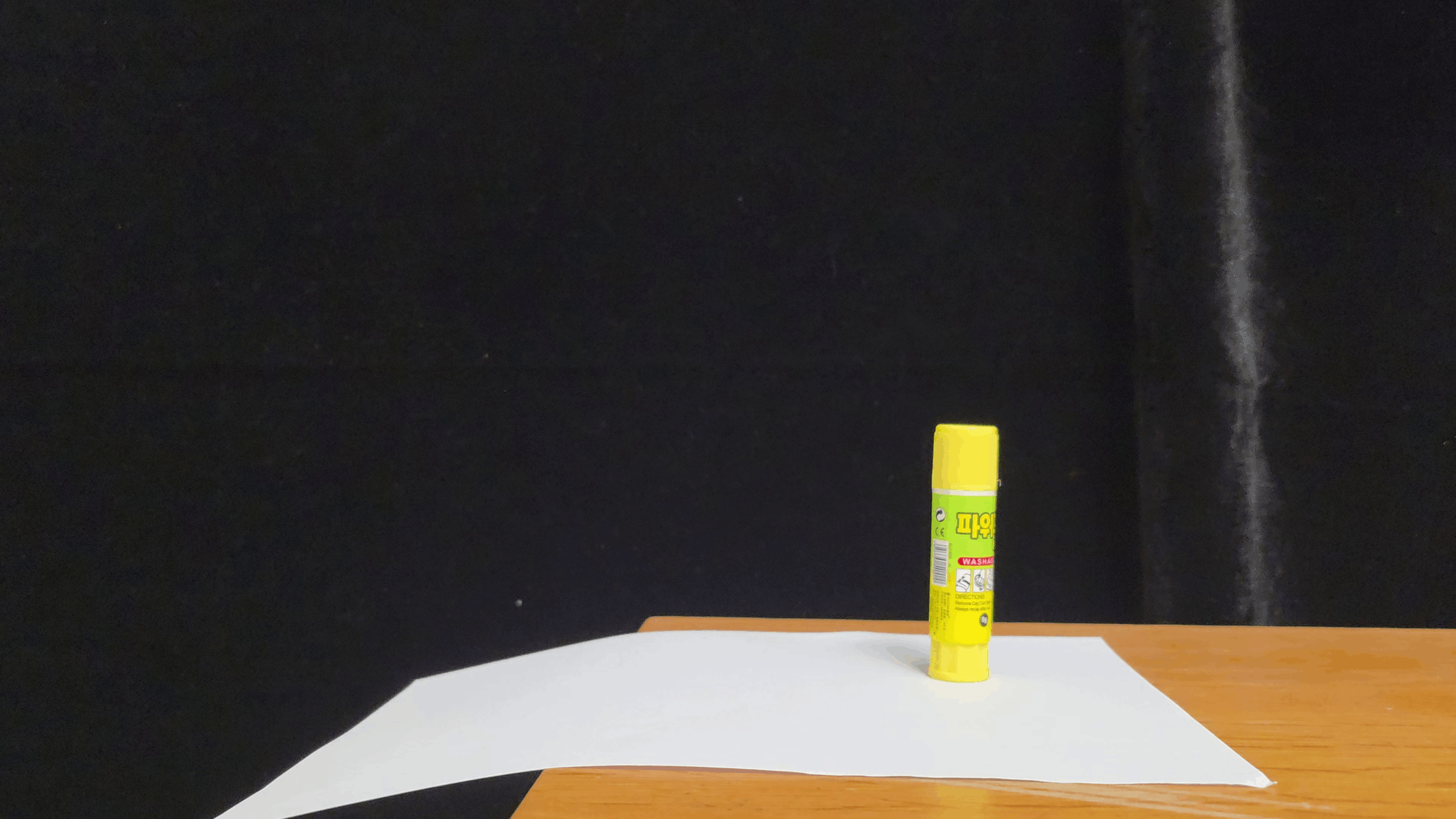}

    \vspace{4pt}
    {\small Example image illustrating the Newton's First Law experiment.}
\end{center}

\begin{itemize}
    \item \textit{Object Position:} “In the image, a sheet of white paper lies flat 
    on a table with a vertical cylindrical object placed on it. The experimenter 
    quickly pulls the paper out horizontally. After the paper is pulled, what is 
    the most likely state of the cylinder?”

    \vspace{4pt}
    \begin{tabular}{@{}ll}
        \textbf{\textcolor{cvprblue}{A)}} & \textcolor{cvprblue}{Remains at original position (Correct)} \\
        \textbf{B)} & Moves with the paper and falls off the table \\
        \textbf{C)} & Moves slightly in the paper’s direction \\
        \textbf{D)} & Moves in the opposite direction of the paper pull \\
    \end{tabular}

    \vspace{8pt}
    \item \textit{Object Stability:} “A lightweight cylindrical object is placed 
    vertically on a sheet of paper. When the paper is rapidly pulled out, what 
    happens to the cylinder?”

    \vspace{4pt}
    \begin{tabular}{@{}ll}
        \textbf{\textcolor{cvprblue}{A)}} & \textcolor{cvprblue}{Will tip over (Correct)} \\
        \textbf{B)} & Will not tip over (remains vertical) \\
        \textbf{C)} & Tilts slightly but stays upright \\
        \textbf{D)} & Cannot be determined \\
    \end{tabular}
\end{itemize}
\end{textcolorbox}

\subsection{Video Generation Prompt Examples}

For the video generation tasks, models receive an \textbf{initial frame} 
(observed from the scene setup) along with a descriptive \textbf{text prompt}. 
Each example illustrates an \textit{Image-to-Video (I2V)} generation instruction, 
where the model must produce physically consistent motion continuing from the 
given frame and description. 

Below are representative example prompts for each physical reasoning domain:

\begin{textcolorbox}[Video Generation Prompt Example of Center of Mass (CoM)]
\begin{center}
    \includegraphics[width=0.35\linewidth,keepaspectratio]{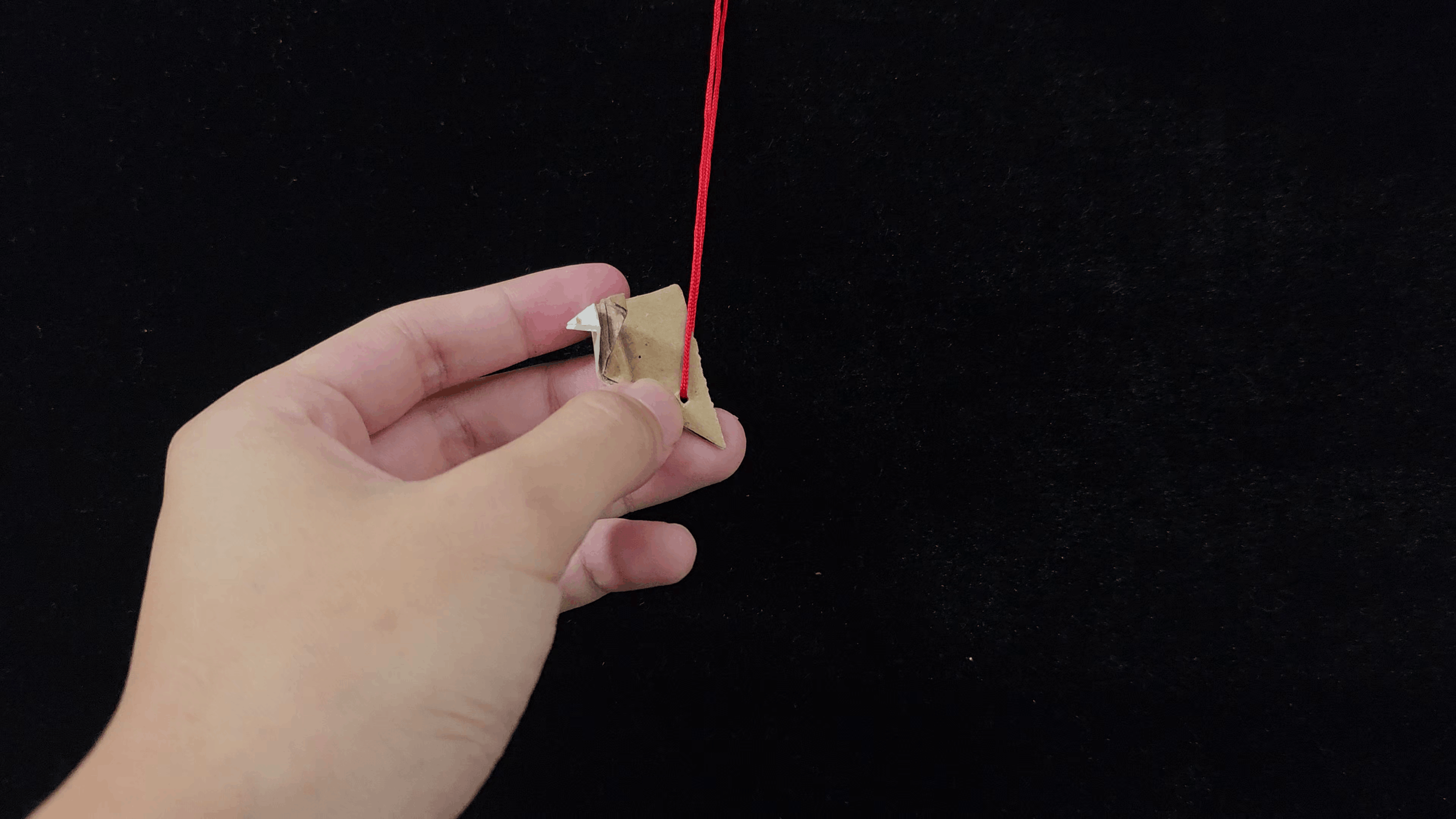}\\
    \small \textit{Initial frame provided to the I2V model for CoM video generation.}
\end{center}

\textit{
“In the suspension method experiment for determining the center of gravity, against a black background, the string is attached to an irregular cardboard whose bottom has been folded, causing the center of gravity to shift downward. The object is held steady and then released. It naturally hangs down and eventually comes to rest in its equilibrium position.”
}
\end{textcolorbox}

\clearpage
\begin{textcolorbox}[Video Generation Prompt Example of Newton’s First Law (NI)]
\begin{center}
    \includegraphics[width=0.2\linewidth,keepaspectratio]{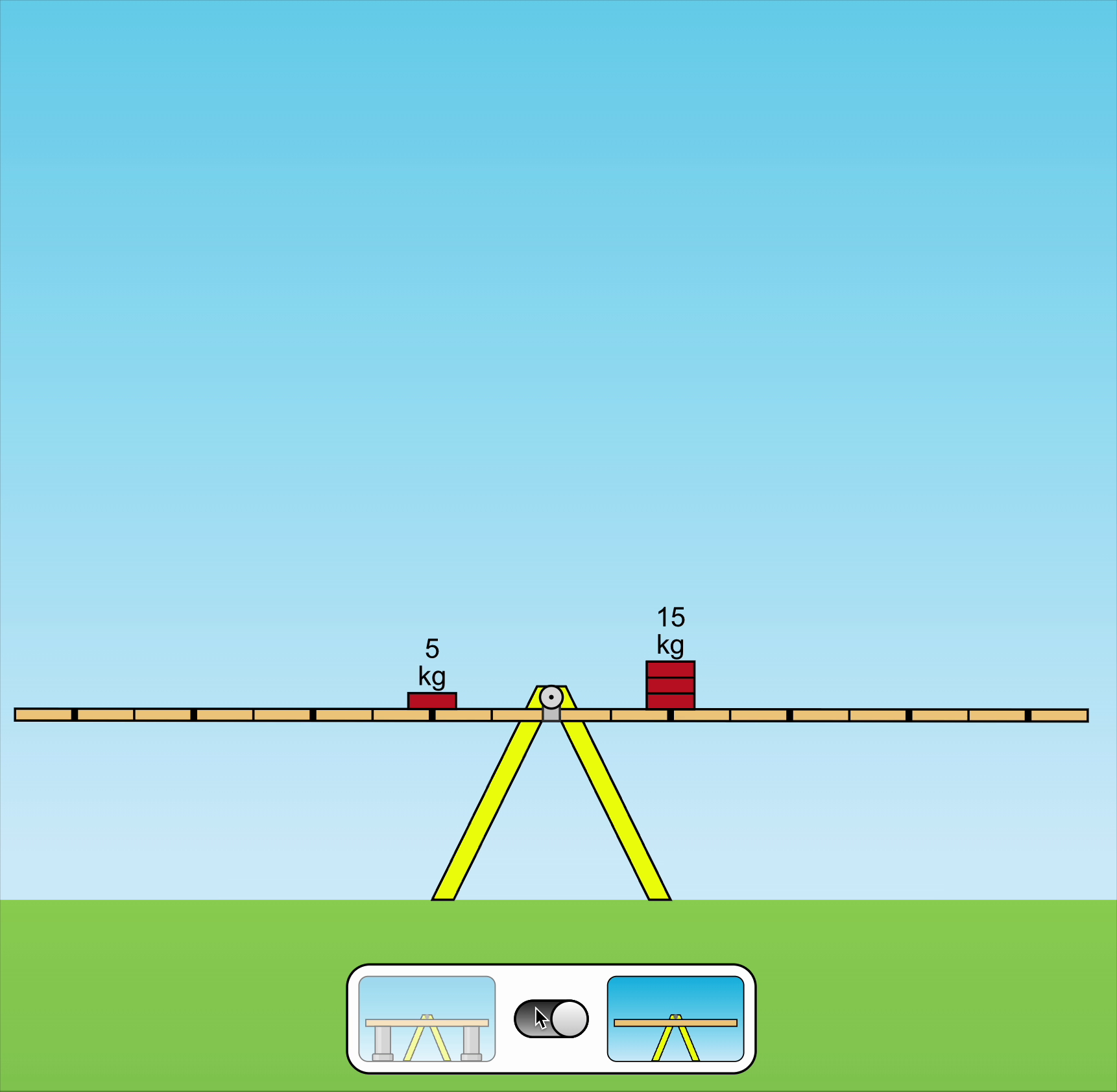}\\
    \small \textit{Initial frame provided to the I2V model for Newton’s First Law video generation.}
\end{center}

\textit{
“In a lever balance condition experiment, against a blue background, a lever is horizontally mounted on a fulcrum with support pillars underneath keeping it balanced. One object is suspended on each side of the lever at equal distances from the fulcrum, with the left side object being lighter.”
}
\end{textcolorbox}

\begin{textcolorbox}[Video Generation Prompt Example of Newton’s First Law (NI)]
\begin{center}
    \includegraphics[width=0.25\linewidth,keepaspectratio]{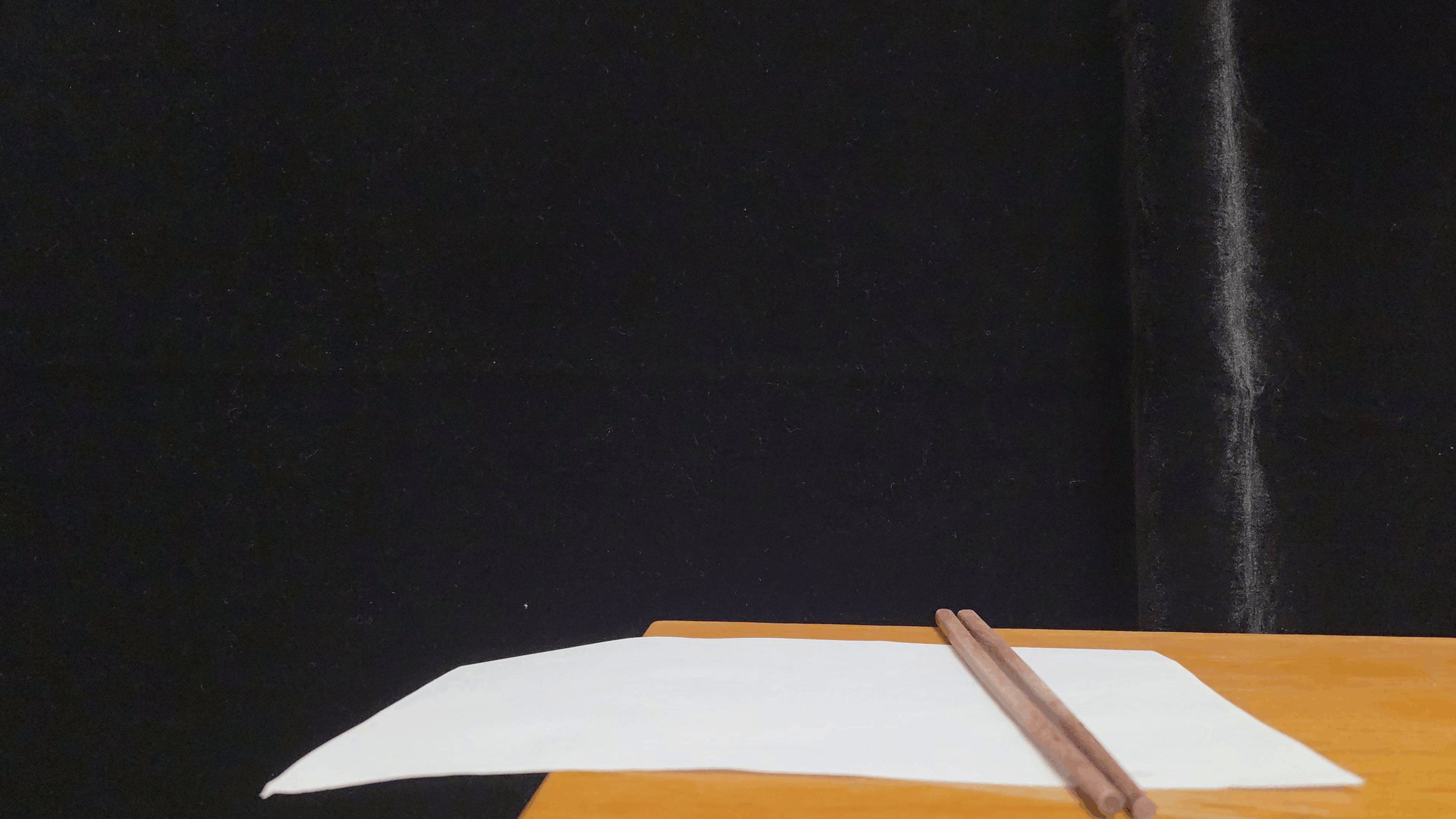}\\
    \small \textit{Initial frame provided to the I2V model for Newton’s First Law video generation.}
\end{center}

\textit{
“In an experiment to verify Newton's First Law, against a black background, a white paper is laid flat on the table with a lightweight chopstick placed on it. The experimenter quickly pulls the paper out horizontally.”
}
\end{textcolorbox}

\newpage
\section{Experiments}

\subsection{Experimental Setup and Evaluation Protocol}

This section describes how we evaluate models on the three canonical physics domains of \textsc{PhysicsMind}: Center of Mass, Lever Equilibrium, and Newton's First Law. Vision–language foundation models are assessed on VQA, while video generation models are assessed on physical rollouts, using standardized inputs, controlled settings, and shared metrics to ensure fair and reproducible comparison.




    
    


\paragraph{VQA Evaluation Protocol.}
In the VQA setting, each vision–language model takes an image and a physics question as input and outputs one of four multiple-choice answers (A, B, C, or D). Evaluation is performed by comparing the predicted answer directly against expert-verified ground truth.
To reflect the different types of physical reasoning in \textsc{PhysicsMind}, we evaluate models on three task domains:

\begin{itemize}
\item \textbf{Center-of-Mass Task:}
models reason about geometric and spatial relationships in suspension and balance scenes.
\item \textbf{Lever Equilibrium Task:}
models determine the correct lever outcome by applying torque-based reasoning.
\item \textbf{Newton's First Law Task:}
models infer inertial behavior when external support is removed, predicting displacement or stability.
\end{itemize}


\paragraph{Video Generation Evaluation Protocol.}
For video generation, each model receives an initial frame and a text prompt describing the physical setup, and is required to generate a short video that depicts the resulting motion. We then compare the generated video against the ground-truth rollout to measure physical consistency in object geometry, causal interaction, and kinematic behavior.

We use the same three task domains as in VQA:

\begin{itemize}
\item \textbf{Center-of-Mass Task:}
evaluates whether the generated motion preserves the object shape and the location of the effective Center of Mass.
\item \textbf{Lever Equilibrium Task:}
checks whether the lever reaches the correct terminal state (tilt left, tilt right, or balanced) under the applied torques.
\item \textbf{Newton's First Law Task:}
assesses trajectory, speed, and acceleration stability under zero external force.
\end{itemize}


\subsection{Specification of VLMs}
We evaluate 24 vision language models. Table~\ref{tab:all_vlms} summarizes their main characteristics, including availability, functional category (reasoning oriented vs.\ chat style), and the architecture used for visual encoding.

Most models in our study are accessed through proprietary APIs, such as those from OpenAI, Google Gemini, Anthropic Claude, and xAI Grok. Six models provide publicly released weights: the Qwen VL series (8B, 32B, 72B), DeepSeek-VL2, GLM-4.5V, and the Llama-3.2 vision models (11B and 90B). Models designed with explicit multistep inference capabilities are grouped as \emph{reasoning} models (e.g., GPT-5, DeepSeek-R1, Claude Sonnet), whereas assistants designed for general purpose multimodal tasks are categorized as \emph{chat} models.

All evaluated systems adopt a ViT or CLIP style backbone for visual encoding, although the integration mechanisms vary across model families. Common designs include fusion via cross attention (OpenAI, Llama-3.2), Q-Former modules (DeepSeek), projection layers based on resamplers (Qwen VL), and proprietary multimodal encoders (Gemini). These components project visual features into the language model's embedding space, enabling unified multimodal processing.

\begin{table}[h]
\caption{Specifications of the 24 evaluated vision--language models. The table reports availability (open vs.\ closed), functional category (reasoning vs.\ chat), and the architecture of each model's multimodal vision encoder.}
\label{tab:all_vlms}
\centering
\small
\begin{tabular}{@{}lccl@{}}
\toprule
\textbf{Model} & \textbf{Open/Closed} & \textbf{Type} & \textbf{Vision Backbone} \\
\midrule
Claude-Sonnet-4.5 & Closed & Reasoning & ViT-based encoder \\
Claude-3.5-Sonnet & Closed & Reasoning & ViT-based encoder \\
Claude-3.7-Sonnet & Closed & Reasoning & ViT-based encoder \\
DeepSeek-R1 & Closed & Reasoning & Q-Former + ViT \\
DeepSeek-VL2 & Open & Chat & Hybrid ViT + Q-Former \\
Gemini-2.5-Pro & Closed & Reasoning & Gemini multimodal encoder \\
Gemini-2.5-Flash & Closed & Chat & Gemini encoder \\
Gemini-2.5-Flash-Image & Closed & Chat & Gemini encoder \\
Gemini-Pro-Vision & Closed & Chat & Gemini encoder \\
GLM-4.5V & Open & Chat & ViT-based encoder \\
Grok-4 & Closed & Reasoning & ViT + cross-attn \\
GPT-4-Turbo & Closed & Reasoning & ViT + cross-attn \\
GPT-4o & Closed & Chat & ViT + cross-attn \\
GPT-4o-mini & Closed & Chat & ViT + cross-attn \\
GPT-5 & Closed & Reasoning & ViT + cross-attn \\
Llama-3.2-11B-Vision & Open & Chat & ViT + cross-attn \\
Llama-3.2-90B-Vision & Open & Chat & ViT + cross-attn \\
o4-mini & Closed & Reasoning & ViT + cross-attn \\
Qwen-VL-Max & Closed & Chat & ViT + resampler \\
Qwen2.5-VL-32B-Instruct & Open & Chat & ViT + resampler \\
Qwen2.5-VL-72B-Instruct & Open & Chat & ViT + resampler \\
Qwen3-VL-8B-Instruct & Open & Chat & ViT + resampler \\

\bottomrule
\end{tabular}
\end{table}

\subsection{Specifications of Video Models}
We evaluate a diverse set of contemporary video generation models with different frame rates and output resolutions (Table~\ref{tab:video-models}). For all models, we standardize the prompt format and generated clip length to make the physics evaluation as comparable as possible.


\begin{table}[htbp]
\caption{Video generation models used in the PhysicsMind evaluation.}
\label{tab:video-models}
\centering
\begin{tabular}{lcc}
\toprule
\textbf{Model} & \textbf{FPS} & \textbf{Resolution} \\
\midrule
Veo-3.1              & 24 & 1280$\times$720 \\
Sora-2               & 24 & 1280$\times$704 \\
LTX-Video            & 30 & 768$\times$512 \\
CogVideoX1.5-5B-I2V  & 16 & 768$\times$1360 \\
Pyramid Flow         & 24 & 1280$\times$768 \\
Wan-2.2 14B          & 16 & 1280$\times$720 \\
Cosmos-predict2 2B   & 16 & 960$\times$704 \\
\bottomrule
\end{tabular}
\end{table}


\section{Evaluate Metrics}

In this section, we introduce the physics-aware metrics used to evaluate video generation on \textsc{PhysicsMind}. Rather than relying on generic perceptual scores, we compare generated videos against ground truth along three canonical mechanics scenarios: (i) \emph{Center-of-Mass}, where we measure whether the model preserves object shape and the location of the effective mass distribution; (ii) \emph{Lever Equilibrium}, where we check if the final lever state matches the torque implied by masses and lever arms; and (iii) \emph{Newton's First Law (Inertia)}, where we examine full-rollout trajectories, speeds, accelerations, and overall motion direction. Together, these metrics provide a complementary view of how well a video model respects basic physical laws, beyond producing visually plausible footage.

\subsection{Center-of-Mass}

The Center of Mass is the latent variable that governs how a rigid body hangs, rotates, and re-balances under gravity. 
If a video model does not place objects with a physically consistent Center of Mass, subsequent lever and inertia behaviors will also be unreliable. 
Our center-of-mass metrics therefore focus on whether the generator can recover the \emph{shape and location} of the mass distribution, rather than only producing visually plausible silhouettes.

\subsubsection{Segmentation Mask IoU}

\noindent \textbf{Definition:} This metric measures the geometric overlap fidelity between the generated object shape and the ground truth. We utilize the Segment Anything Model (SAM) to derive binary masks from tracked points to assess shape preservation.

\noindent \textbf{Formulation:} Let \(M_{\text{gt}}\) and \(M_{\text{pred}}\) correspond to the binary masks of the ground-truth and predicted objects, respectively. The Intersection over Union (IoU) is defined as:
\begin{equation}
    \text{IoU} = \frac{\sum_{i,j} M_{\text{gt}}(i,j) \cdot M_{\text{pred}}(i,j)}{\sum_{i,j} \max(M_{\text{gt}}(i,j), M_{\text{pred}}(i,j))}
\end{equation}
The IoU ranges from \([0, 1]\), where higher values indicate better segmentation accuracy.

\subsubsection{Segmentation Mask Center Distance}

\noindent \textbf{Definition:} This metric quantifies the spatial implementation error of the mass distribution. It calculates the pixel-level Euclidean distance between the centroid of the predicted object mask and the ground-truth mask.

\noindent \textbf{Formulation:} The centroids of the ground-truth mask (\(\mathbf{c}_{\text{gt}}\)) and predicted mask (\(\mathbf{c}_{\text{pred}}\)) are calculated as the average coordinates of their non-zero pixels. The Center Distance is:
\begin{equation}
    \text{Center Distance} = \left\| \mathbf{c}_{\text{gt}} - \mathbf{c}_{\text{pred}} \right\|_2
\end{equation}
This metric is measured in pixels, with smaller values indicating higher spatial accuracy.


\subsection{Lever Equilibrium}

Lever Equilibrium exposes whether a model can respect torque balance, instead of just copying common “left-down/right-down” visual patterns. 
In our tabletop setups the lever dynamics are essentially one-dimensional and the outcome is categorical (left, right, balanced), 
so the most physically meaningful signal is whether the \emph{final state} matches the torque implied by the masses and lever arms, rather than pixel-wise similarity of intermediate frames.

\subsubsection{Final State Accuracy}

\noindent \textbf{Definition:} The proportion of samples for which the model correctly predicts the final state of the lever (e.g., left-down, right-down, or balanced), consistent with the VLM-based comparison in Fig.~X.

\noindent \textbf{Formulation:}
\begin{equation}
    \text{Accuracy} = \frac{1}{S}\sum_{i=1}^{S} \mathbb{I}\bigl(y_{\text{pred}}^{(i)} = y_{\text{true}}^{(i)}\bigr)
\end{equation}
where \(S\) is the total number of samples and \(\mathbb{I}(\cdot)\) is the indicator function. The accuracy ranges from \([0, 1]\).


\subsection{Newton's First Law (Inertia)}

Newton's first law controls whether objects keep moving at constant velocity when no external forces act on them. 
For this scenario, it is not enough to check a single end frame: a model might coincidentally land at the right location while producing highly non-inertial motion in between. 
We therefore evaluate the entire rollout using trajectory-, speed-, and acceleration-level metrics, together with a global direction check.

\noindent Let \(T\) be the number of frames and \(N\) the number of tracked points (objects). The position of point \(n\) at frame \(t\) is denoted as \(\mathbf{p}_{\text{gt}}^{(t,n)}\) for ground truth and \(\mathbf{p}_{\text{pred}}^{(t,n)}\) for prediction.

\subsubsection{Trajectory RMSE}

\noindent \textbf{Definition:} The Root Mean Squared Error (RMSE) measures the global deviation between the predicted motion path and the ground-truth trajectory. It is computed in normalized coordinates to ensure scale invariance across different video resolutions.

\noindent \textbf{Formulation:} We first normalize pixel coordinates \((x,y)\) to \([0, 1]\) by dividing by the frame width \(W\) and height \(H\). The normalized trajectory RMSE is then:
\begin{equation}
    \text{RMSE}_{\text{traj}} =
    \sqrt{\frac{1}{T N} \sum_{t=1}^{T} \sum_{n=1}^{N}
    \left\| \mathbf{p}_{\text{gt}}^{(t,n)} - \mathbf{p}_{\text{pred}}^{(t,n)} \right\|_2^2}
\end{equation}
Smaller values indicate higher trajectory accuracy.

\subsubsection{Final Position Error (FPE)}

\noindent \textbf{Definition:} This metric assesses the model's long-term forecasting ability by measuring the Euclidean distance between the predicted and ground-truth final positions, normalized by the total length of the actual motion path.

\noindent \textbf{Formulation:}
\begin{equation}
    \text{FPE} =
    \frac{\left\| \mathbf{p}_{\text{gt}}^{(T)} - \mathbf{p}_{\text{pred}}^{(T)} \right\|_2}{L_{\text{gt}}}
\end{equation}
where
\begin{equation}
    L_{\text{gt}} = \sum_{t=1}^{T-1}
    \left\| \mathbf{p}_{\text{gt}}^{(t+1)} - \mathbf{p}_{\text{gt}}^{(t)} \right\|_2
\end{equation}
is the total length of the ground-truth trajectory path. Smaller values are better.

\subsubsection{Speed Similarity}

\noindent \textbf{Definition:} The cosine similarity between predicted and ground-truth velocity vectors, averaged over the trajectory, as in the ``Speed Similarity'' block in Fig.~X. 
This metric captures whether the model keeps the \emph{magnitude and local direction} of motion consistent with an inertia-respecting rollout.

\noindent \textbf{Formulation:} For each point \(n\) and frame \(t \ge 2\), we compute velocity vectors
\begin{equation}
    \mathbf{v}_{\text{gt}}^{(t,n)}   = \mathbf{p}_{\text{gt}}^{(t,n)}   - \mathbf{p}_{\text{gt}}^{(t-1,n)}, \quad
    \mathbf{v}_{\text{pred}}^{(t,n)} = \mathbf{p}_{\text{pred}}^{(t,n)} - \mathbf{p}_{\text{pred}}^{(t-1,n)}.
\end{equation}
The speed similarity is defined as the average cosine similarity:
\begin{equation}
    S_{\text{vel}} =
    \frac{1}{(T-1)N}
    \sum_{n=1}^{N} \sum_{t=2}^{T}
    \frac{\mathbf{v}_{\text{gt}}^{(t,n)} \cdot \mathbf{v}_{\text{pred}}^{(t,n)}}
    {\left\|\mathbf{v}_{\text{gt}}^{(t,n)}\right\|_2 \,
     \left\|\mathbf{v}_{\text{pred}}^{(t,n)}\right\|_2}.
\end{equation}
This metric lies in \([-1,1]\); in our setting it typically falls in \([0,1]\). Larger values indicate that the instantaneous motion directions and magnitudes are more consistent with the ground truth.

\subsubsection{Acceleration Similarity}

\noindent \textbf{Definition:} The cosine similarity between predicted and ground-truth acceleration vectors, averaged over the trajectory, corresponding to the ``Acceleration Similarity'' block in Fig.~X. 
This term is particularly sensitive to non-physical jitter or unrealistic speed-up/slow-down patterns that would violate constant-velocity motion.

\noindent \textbf{Formulation:} For each point \(n\) and frame \(t \ge 3\), accelerations are obtained by differencing velocities:
\begin{equation}
    \mathbf{a}_{\text{gt}}^{(t,n)}   = \mathbf{v}_{\text{gt}}^{(t,n)}   - \mathbf{v}_{\text{gt}}^{(t-1,n)}, \quad
    \mathbf{a}_{\text{pred}}^{(t,n)} = \mathbf{v}_{\text{pred}}^{(t,n)} - \mathbf{v}_{\text{pred}}^{(t-1,n)}.
\end{equation}
The acceleration similarity is then
\begin{equation}
    S_{\text{acc}} =
    \frac{1}{(T-2)N}
    \sum_{n=1}^{N} \sum_{t=3}^{T}
    \frac{\mathbf{a}_{\text{gt}}^{(t,n)} \cdot \mathbf{a}_{\text{pred}}^{(t,n)}}
    {\left\|\mathbf{a}_{\text{gt}}^{(t,n)}\right\|_2 \,
     \left\|\mathbf{a}_{\text{pred}}^{(t,n)}\right\|_2}.
\end{equation}
Higher values indicate that the acceleration (i.e., speed-up and slow-down patterns) is better matched.

\subsubsection{Directional Consistency}

\noindent \textbf{Definition:} A global measure of whether the predicted trajectory moves in the same overall direction as the ground-truth trajectory. 
Compared to the local similarity metrics above, this captures whether the model preserves the \emph{coarse, scene-level} motion tendency implied by the setup (e.g., sliding forward vs.\ backward).

\noindent \textbf{Formulation:} Let \(\mathbf{p}_{\text{gt}}^{(1)}\) and \(\mathbf{p}_{\text{gt}}^{(T)}\) be the ground-truth start and end positions, and \(\mathbf{p}_{\text{pred}}^{(1)}\), \(\mathbf{p}_{\text{pred}}^{(T)}\) be the predicted ones. We define displacement vectors
\begin{equation}
    \mathbf{d}_{\text{gt}}   = \mathbf{p}_{\text{gt}}^{(T)}   - \mathbf{p}_{\text{gt}}^{(1)}, \quad
    \mathbf{d}_{\text{pred}} = \mathbf{p}_{\text{pred}}^{(T)} - \mathbf{p}_{\text{pred}}^{(1)}.
\end{equation}
The angle \(\theta\) between them is
\begin{equation}
    \theta = \arccos\left(
    \frac{\mathbf{d}_{\text{gt}} \cdot \mathbf{d}_{\text{pred}}}
         {\left\|\mathbf{d}_{\text{gt}}\right\|_2 \,
          \left\|\mathbf{d}_{\text{pred}}\right\|_2}
    \right),
\end{equation}
where \(\theta\) is measured in degrees. The directional consistency score is
\begin{equation}
    S_{\text{dir}} = \frac{180 - \theta}{180}.
\end{equation}
Thus \(S_{\text{dir}} = 1\) when the two trajectories share exactly the same overall direction (\(\theta = 0^\circ\)), and \(S_{\text{dir}} = 0\) when they move in opposite directions (\(\theta = 180^\circ\)).
\section{Additional Statistical analyses for PhysicsMind}
\label{sec:appendix_stats}

This appendix complements the main results with a more systematic analysis of model behavior on \textsc{PhysicsMind} for both VQA and video generation.
We organise the discussion around three research questions:

\medskip
\noindent\textbf{RQ1 (Task difficulty).}
Do the three physics tasks exhibit a consistent and statistically supported difficulty ordering?

\noindent\textbf{RQ2 (Model families).}
Are there significant performance differences across model ecosystems and between ``chat'' and ``reasoning'' model variants?

\noindent\textbf{RQ3 (Cross-modality consistency).}
Do VQA and video-based evaluation expose aligned physical failure modes, or do they stress different aspects of physical competence?

\noindent Unless otherwise noted, all analyses use the same data splits, metrics, and experimental protocol as in the main paper.

\subsection{VQA performance across tasks and model families (RQ1, RQ2)}
\label{sec:appendix_vqa}

We analyse the performance of 22 models on the three PhysicsMind VQA tasks using descriptive statistics, one-way and two-way ANOVA, Pearson correlations, and two-sample t-tests. The focus is on task difficulty, category effects, and the extent to which the tasks capture complementary abilities.

\subsubsection{Task-level statistics and difficulty ordering}

Aggregating accuracies of all models on the three VQA tasks (Newton's First Law, Lever Equilibrium, Center of Mass) yields the following patterns:

\begin{itemize}
    \item Newton's First Law has the highest mean accuracy at roughly $60.8\%$.
    \item Lever Equilibrium has a mean accuracy around $48.0\%$.
    \item Center of Mass is the hardest, with a mean accuracy around $39.8\%$.
    \item All three tasks have standard deviations between $12$ and $13$ percentage points.
\end{itemize}

\noindent Thus, the main differences lie in the \emph{mean level} rather than in dispersion across models: the spread of accuracies is similar, but the average difficulty follows
\[
\text{Newton's First Law} > \text{Lever Equilibrium} > \text{Center of Mass}.
\]
This matches the intended complexity ladder: from relatively direct inertia judgements, to lever balance with explicit geometric reasoning, to Center of Mass problems that are more sensitive to geometric configuration and mass distribution.

\subsubsection{Model category effects: one-way ANOVA}

\begin{figure}[H]
\centering
\includegraphics[width=0.7\textwidth]{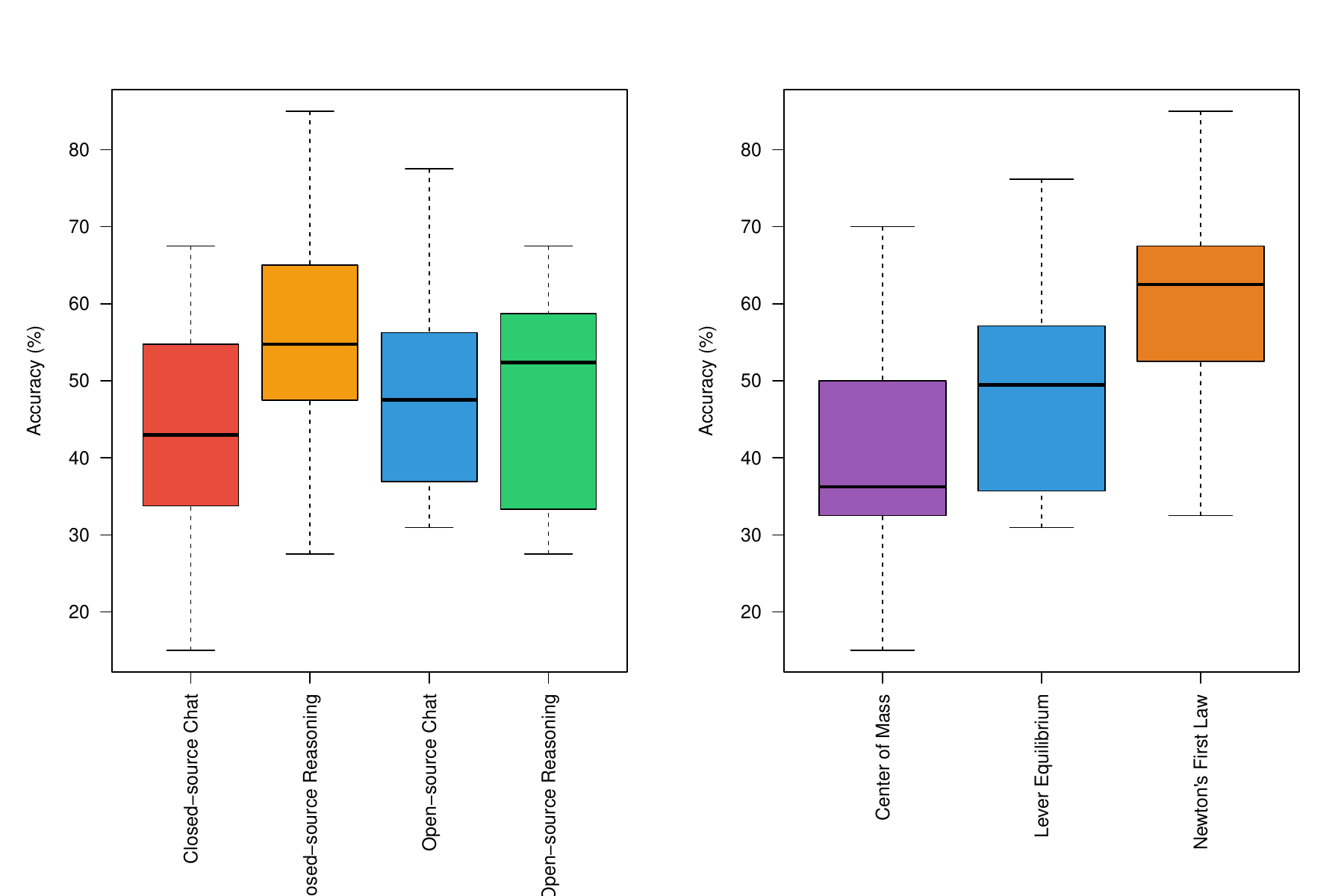}
\caption{Overall PhysicsMind VQA accuracy by model category.}
\label{fig:boxplot}
\end{figure}

Figure~\ref{fig:boxplot} shows boxplots of overall VQA accuracy for four model categories (closed-source / open-source $\times$ chat / reasoning):

\begin{itemize}
    \item Closed-source reasoning models have the highest median accuracy.
    \item Closed-source chat models have the lowest median.
    \item The height of the boxes and whiskers is comparable across categories, indicating similar within-category variability.
\end{itemize}

We run a one-way ANOVA with \texttt{Category} as the independent variable and overall VQA accuracy as the dependent variable:

\begin{table}[H]
\centering
\caption{One-way ANOVA of overall VQA accuracy by model category.}
\begin{tabular}{lccccl}
\toprule
\textbf{Source} & \textbf{df} & \textbf{SS} & \textbf{MS} & \textbf{F} & \textbf{p} \\
\midrule
Category  & 3  & 1279  & 426.4 & 1.963 & 0.1288 \\
Residuals & 62 & 13468 & 217.2 &  --   & --     \\
\bottomrule
\end{tabular}
\end{table}

The ANOVA yields
\[
F(3,62) = 1.963,\quad p = 0.1288 > 0.05,
\]
so we cannot reject the null hypothesis that the four categories have the same mean accuracy. In Tukey HSD post-hoc tests, the largest contrast (closed-source reasoning vs.\ closed-source chat) has a mean difference of about $11.9$ percentage points, but the multiple-comparisons-adjusted p-value is $\approx 0.09$ and not statistically significant.
Under the current sample size and variance, there is no robust evidence that any single category enjoys a stable advantage on PhysicsMind VQA. This is consistent with the main paper's observation that basic physics remains challenging across ecosystems.

\subsubsection{Task main effects and Category--Task interaction: two-way ANOVA}

We next fit a two-way ANOVA with \texttt{Category} and \texttt{Task} as factors (no repeated-measures structure):

\begin{table}[H]
\caption{Two-way ANOVA of VQA accuracy with Category and Task.}
\centering
\begin{tabular}{lccccl}
\toprule
\textbf{Source} & \textbf{df} & \textbf{SS} & \textbf{MS} & \textbf{F} & \textbf{p} \\
\midrule
Category       & 3  & 1279 & 426.4  &  3.095 & 0.034          \\
Task           & 2  & 4922 & 2460.9 & 17.867 & $<0.001^{***}$ \\[2pt]
Category:Task  & 6  & 1108 & 184.7  &  1.341 & 0.255          \\
Residuals      & 54 & 7438 & 137.7  &   --   & --             \\
\bottomrule
\end{tabular}
\end{table}

The main effect of \texttt{Task} is highly significant ($p<0.001$), formally confirming the difficulty ordering discussed above. The main effect of \texttt{Category} reaches $p=0.034$ in this specification, but the effect size is small and somewhat unstable when compared with the one-way ANOVA and Tukey tests. Crucially, the \texttt{Category:Task} interaction is not significant ($p=0.255>0.05$).

\begin{figure}[H]
\centering
\includegraphics[width=0.8\textwidth]{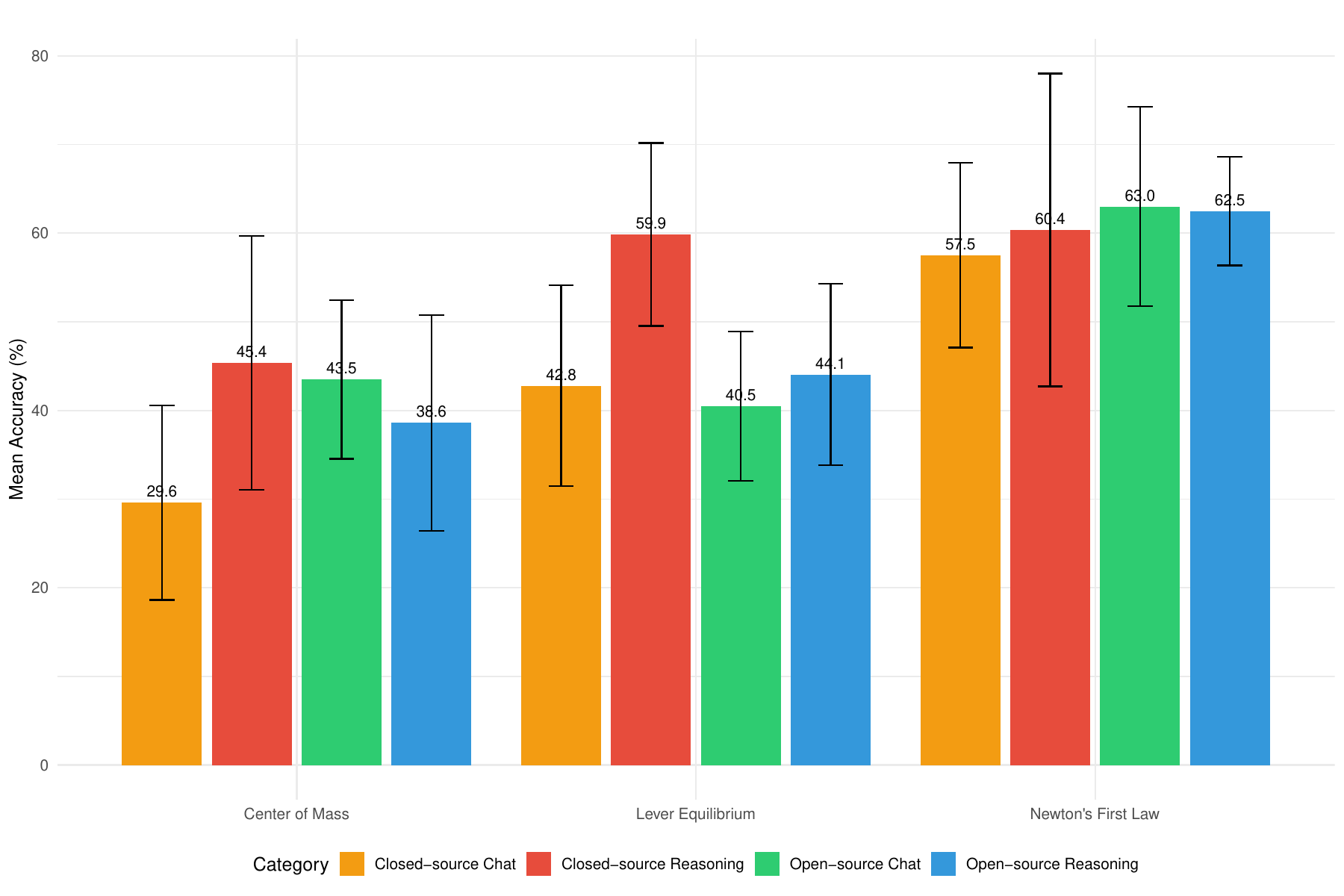}
\caption{Mean VQA accuracy of four model categories on the three tasks.}
\label{fig:interaction}
\end{figure}

The interaction is visualised in Figure~\ref{fig:interaction}: the performance curves of the four categories over the three tasks are roughly parallel. Every category performs best on Newton's First Law and worst on Center of Mass, which supports the view that difficulty is driven primarily by the underlying physical structure of each task rather than by ecosystem-specific preferences or training recipes.

\subsubsection{Task correlations: complementary rather than redundant}

\begin{figure}[H]
\centering
\includegraphics[width=0.8\textwidth]{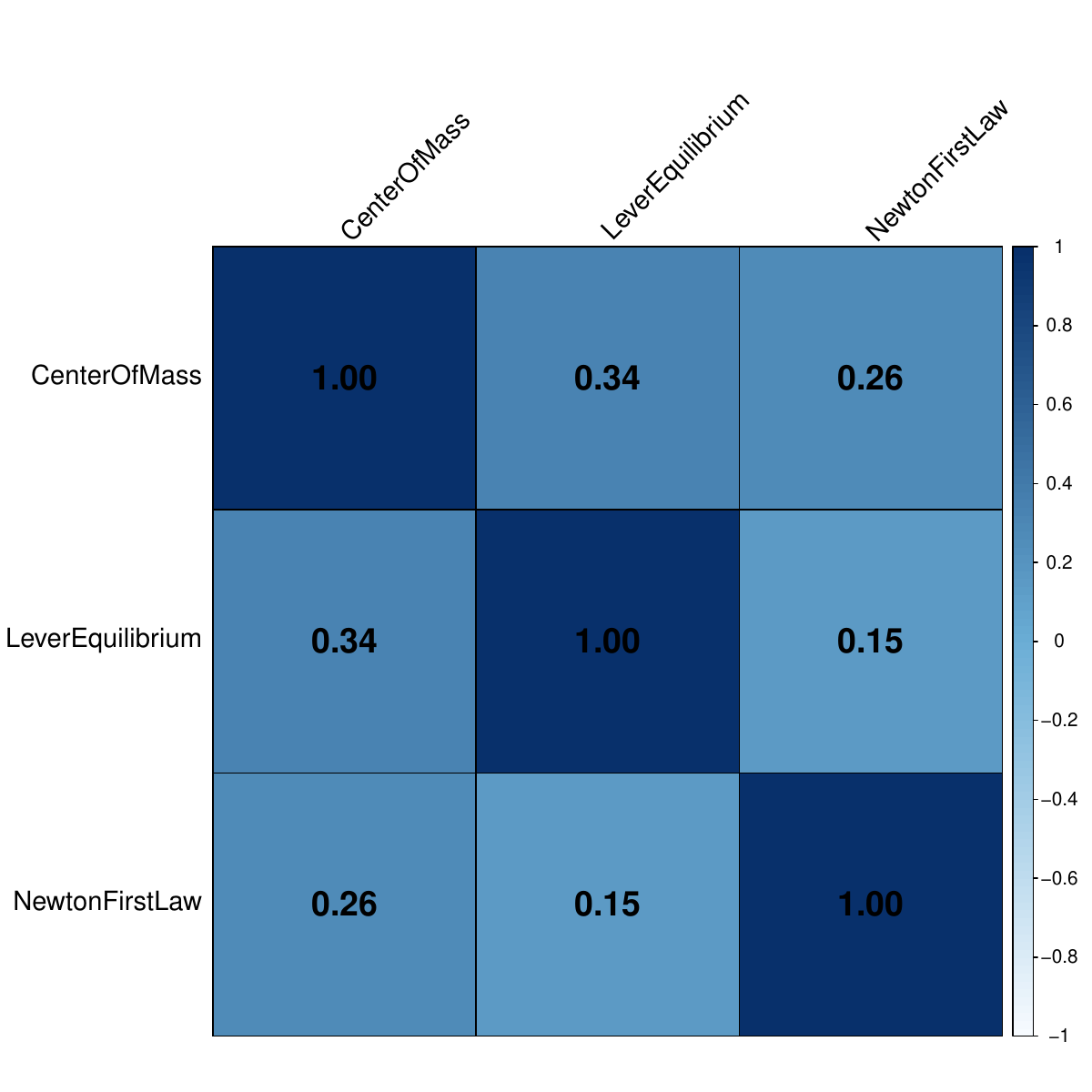}
\caption{Pearson correlation matrix among the three PhysicsMind VQA tasks.}
\label{fig:correlation}
\end{figure}

Figure~\ref{fig:correlation} and Table~\ref{tab:correlation} show the Pearson correlation matrix of accuracies across the three VQA tasks.

\begin{table}[H]
\caption{Pearson correlation coefficients among VQA tasks.}
\centering
\label{tab:correlation}
\begin{tabular}{lccc}
\toprule
 & \textbf{Center of Mass} & \textbf{Lever Equilibrium} & \textbf{Newton's First Law} \\
\midrule
Center of Mass      & 1.000 & 0.337 & 0.265 \\
Lever Equilibrium   & 0.337 & 1.000 & 0.151 \\
Newton's First Law  & 0.265 & 0.151 & 1.000 \\
\bottomrule
\end{tabular}
\end{table}

All off-diagonal correlations lie between $0.15$ and $0.34$, i.e., weak to moderate positive correlations. At the $0.05$ significance level, none of them are statistically significant: the corresponding p-values are approximately $0.12$, $0.23$, and $0.50$.

In practice, high accuracy on one task does not strongly predict performance on the others. From the task design:

\begin{itemize}
    \item Center of Mass emphasizes spatial/geometric reasoning and mass distribution.
    \item Lever Equilibrium focuses on torque balance and lever-arm comparison.
    \item Newton's First Law targets understanding of motion states, inertia, and the presence or absence of external forces.
\end{itemize}

The three tasks therefore behave as complementary probes rather than redundant measurements of a single latent ``physics score''. This empirically supports the main paper's aim of capturing multi-dimensional physical competence.

\subsubsection{Ecosystem and paradigm comparisons: two-sample t-tests}

We finally report two standard splits via independent two-sample t-tests, used in the main paper to support the claim that physics is a common failure mode across ecosystems.

\paragraph{Closed-source vs.\ open-source models.}

\begin{table}[H]
\caption{Two-sample t-test of overall VQA accuracy: closed-source vs.\ open-source models.}
\centering
\begin{tabular}{ll}
\toprule
Test statistic      & $t(64) = 0.420$ \\
p-value             & 0.676           \\
Mean difference     & 1.55\% (closed-source higher) \\
95\% CI             & [-5.81\%, 8.91\%] \\
Closed-source mean  & 50.24\% \\
Open-source mean    & 48.69\% \\
\bottomrule
\end{tabular}
\end{table}

The p-value is far above $0.05$ and the confidence interval crosses zero, indicating no statistically significant difference in mean performance between closed- and open-source models on PhysicsMind VQA. The ``closed vs.\ open'' label alone does not predict physics reasoning ability.

\paragraph{Reasoning vs.\ chat models.}

\begin{table}[H]
\caption{Two-sample t-test of overall VQA accuracy: reasoning vs.\ chat models.}
\centering
\begin{tabular}{ll}
\toprule
Test statistic      & $t(62.38) = -1.695$ \\
p-value             & 0.095 \\
Mean difference     & -6.21\% (reasoning higher) \\
95\% CI             & [-13.53\%, 1.11\%] \\
Reasoning mean      & 52.36\% \\
Chat mean           & 46.15\% \\
\bottomrule
\end{tabular}
\end{table}

Reasoning models have higher mean accuracy, and the p-value ($\approx 0.095$) is close to conventional significance thresholds. However, the effect is not statistically significant under a strict $0.05$ level. This aligns with the main paper: generic chain-of-thought or instruction tuning helps on physics questions, but does not remove systematic physical errors.

\paragraph{Relation to the research questions.}
Taken together, the VQA analyses above (i) establish a clear and statistically supported difficulty ordering across the three physics tasks, directly answering \textbf{RQ1}, and (ii) show that ecosystem- and paradigm-level differences are modest and often not statistically robust, informing \textbf{RQ2} by indicating that physics remains a shared challenge across model families.

\subsection{Video-based physical evaluation and cross-modality consistency (RQ1, RQ3)}
\label{sec:appendix_video}

We now discuss the video evaluation results of the seven video generation models on the three PhysicsMind scenarios, based on the metrics reported in the main paper. Unlike VQA, video evaluation involves multiple real-valued physics metrics, some ``higher is better'', some ``lower is better'', and some potentially negative. We therefore focus on qualitative patterns rather than constructing a single aggregate score.

\subsubsection{Task-level patterns}

\paragraph{Center of Mass (video).}

In this scenario, we measure whether generated videos obey center-of-mass constraints using the mask IoU and the error of the object's center-of-mass position. Overall:

\begin{itemize}
    \item All models achieve IoU values very close to zero, far below typical ``usable'' segmentation levels.
    \item Center-of-mass position errors are large in pixel space (mean around $\sim 180$ pixels). Although there are differences across models, all lie in a regime of substantial deviation.
\end{itemize}

This suggests that when the generator needs to balance appearance with physical consistency, current models have almost no reliable control over the Center of Mass. They tend to place objects in a visually plausible way rather than aligning them to a physically consistent mass distribution.

\paragraph{Lever Equilibrium (video).}

For the lever scenario, we only consider whether the final balanced state is correct. According to the main paper results, even the strongest model achieves only about $45$--$50\%$ final-state accuracy. The overall mean is around $35\%$, noticeably below the corresponding VQA mean of roughly $48\%$.

In other words, even when a model can answer ``which side is heavier'' in a static question, driving the full generation process towards the correct equilibrium state is considerably harder. There is a disconnect between linguistic reasoning and physical trajectory generation.

\paragraph{Newton's First Law (video).}

In the inertia scenario, we use trajectory RMSE, final position error, velocity and acceleration similarity, and directional consistency to assess compliance with inertial motion:

\begin{itemize}
    \item Trajectory RMSE lies in a relatively narrow range across models ($\sim 0.35$--$0.41$), indicating comparable error magnitude in the global shape of the trajectory.
    \item Final position error and directional consistency vary more: some models with reasonable RMSE still exhibit large terminal offsets or obvious ``turn-back'' motion.
    \item Velocity and acceleration similarity span near-zero and strongly negative values, reflecting cases where the motion looks smooth but the direction of change in velocity is physically inverted.
\end{itemize}

Among the three video scenarios, Newton's First Law is relatively better handled, yet there remains a noticeable gap between the generated motion and physically plausible trajectories.

\subsubsection{Model differences and metric heterogeneity}

Unlike VQA, where accuracy has a single direction, video evaluation metrics differ in both scale and direction: smaller is better for position error and RMSE, larger is better for IoU and directional consistency, and similarity scores can be negative. Rather than forcing a global score, several stable patterns emerge:

\begin{itemize}
    \item Most models excel on only a few specific metrics; for instance, some achieve higher final-state accuracy in the lever scenario but do not reduce center-of-mass error.
    \item No model is best on all physical metrics simultaneously: a model that is strong on global trajectory shape (RMSE) may still be weak on terminal position or directional consistency.
    \item Each model shows strong task dependence: a model that is relatively stable on inertia can still perform poorly on center-of-mass scenarios.
\end{itemize}

Taken together, current video generators look more like collections of local strengths than coherent physical world models. The heterogeneity across metrics reinforces the need for multi-metric, scenario-wise evaluation.

\subsubsection{Comparing VQA and video: a gap between understanding and generation}

Comparing Section~\ref{sec:appendix_vqa} with the video analysis above yields several observations that support the main paper:

\begin{itemize}
    \item \textbf{Video generation is consistently harder than VQA under the same physics setup.}
    For example, in the lever scenario, final-state accuracy in generation is lower than VQA accuracy, indicating a systematic gap between answering correctly and producing a physically correct process.
    \item \textbf{Center of Mass scenarios are the hardest on both sides.}
    Center of Mass questions have the lowest VQA accuracies, and the corresponding video metrics (IoU and position error) are also the worst. Spatial geometry and mass distribution emerge as joint bottlenecks for understanding and generation.
    \item \textbf{The difficulty ordering is consistent across understanding and generation.}
    In both settings, we observe
    \[
    \text{Center of Mass} > \text{Lever Equilibrium} > \text{Newton's First Law}
    \]
\noindent From hardest to easiest, reinforcing the interpretation of the three scenarios as distinct levels of physical complexity.
\end{itemize}

Overall, the VQA and video analyses are qualitatively aligned: the three PhysicsMind tasks form complementary axes for static understanding and reveal analogous failure modes in dynamic generation. Together, they provide a unified and fine-grained lens on the physical consistency of large models.

\paragraph{Relation to the research questions.}
The video-based analyses primarily address \textbf{RQ3}: VQA and video evaluation expose strongly aligned difficulty patterns and shared bottlenecks, while also revealing a persistent gap between ``knowing'' and ``generating''. At the same time, they refine \textbf{RQ1} by showing that the difficulty ordering observed for static VQA persists when models are required to produce full physical rollouts, highlighting the need for architectures that couple visible dynamics with coherent internal world models.

\clearpage
\section{Prompt Ablations for Physics-Aware VQA and Video Generation}

\subsection{Prompt Design and Ablation for VQA}

We investigate how different prompting strategies influence GPT-5's performance on our multiple-choice Physics VQA benchmarks. All experiments use identical images, questions, and evaluation settings; only the prompt template is varied. Decoding hyperparameters are fixed across conditions (temperature 0.3, maximum output length 8k tokens, and up to five retries on API failure).

\begin{table}[!ht]
  \caption{
    Prompt ablation on the Physics VQA datasets for two representative models (GPT-5 and GPT-4o).
    We report subtype-level and overall accuracies (\%) across the three prompting strategies:
    Direct Answering (DA), Chain-of-Physics (CoP), and Law-Conditioned (LC).
  }
  \centering

  \label{tab:vqa_prompt_ablation_full}
  \vspace{0.3em}
  \resizebox{\textwidth}{!}{
    \begin{tabular}{l|ccc|ccc|ccc}
      \toprule
      \multirow{3}{*}{Model \& Prompt} &
      \multicolumn{3}{c|}{Center of Mass (VQA)} &
      \multicolumn{3}{c|}{Lever Equilibrium (VQA)} &
      \multicolumn{3}{c}{Newton's First Law (VQA)} \\
      \cmidrule(lr){2-4} \cmidrule(lr){5-7} \cmidrule(lr){8-10}
       & Position & Rotation & Overall &
         Equilibrium & Balance Adj. & Overall &
         Obj. Pos. & Obj. Stability & Overall \\
       & Acc (\%) & Acc (\%) & Acc (\%) &
         Acc (\%) & Acc (\%) & Acc (\%) &
         Acc (\%) & Acc (\%) & Acc (\%) \\
      \midrule

      GPT-5 (DA) &
        60.0 & 80.0 & 70.0 &
        85.7 & 66.7 & 76.2 &
        60.0 & 95.0 & 77.5 \\

      GPT-5 (CoP) &
        69.2 & 64.0 & 65.8 &
        81.0 & 61.9 & 71.4 &
        70.0 & 95.0 & 82.5 \\

      GPT-5 (LC) &
        25.0 & 75.0 & 50.0 &
        81.0 & 66.7 & 73.8 &
        65.0 & 85.0 & 75.0 \\

      \midrule

      GPT-4o (DA) &
        50.0 & 70.0 & 60.0 &
        28.6 & 61.9 & 45.2 &
        45.0 & 75.0 & 60.0 \\

      GPT-4o (CoP) &
        38.5 & 85.7 & 67.6 &
        42.9 & 42.9 & 42.9 &
        40.0 & 89.5 & 64.1 \\

      GPT-4o (LC) &
        25.0 & 50.0 & 37.5 &
        42.9 & 61.9 & 52.4 &
        40.0 & 90.0 & 65.0 \\

      \bottomrule
    \end{tabular}
  }
\end{table}

\vspace{0.5em}

\subsection{VQA: Effect of Prompt Design on Physical Question Answering}
\label{sec:appendix_prompt}

Table~\ref{tab:vqa_prompt_ablation_full} summarizes the effects of the three prompting strategies across the three Physics VQA domains: Lever Equilibrium (LE), hanging Center of Mass questions (CoM), and the Newton's first law (NFL) paper pulling setup.

\noindent For GPT-5, the DA prompt already yields strong performance across all domains (overall: $76.2\%$ on LE, $70.0\%$ on CoM, and $77.5\%$ on NFL). The CoP prompt provides the largest gains on the NFL dataset, which requires intensive reasoning, increasing overall accuracy to $82.5\%$ primarily by improving the object position subtype (from $60.0\%$ to $70.0\%$) while maintaining high object stability accuracy ($95.0\%$). On Lever Equilibrium, CoP and LC remain close to DA, with LC giving a small improvement to $73.8\%$ overall, likely because the torque relationships are visually explicit. For CoM, DA remains the strongest overall. CoP improves position understanding but reduces rotation accuracy, while LC underperforms substantially on the position subtype, resulting in a lower overall score ($50.0\%$).

\noindent In contrast, GPT-4o exhibits lower absolute performance but still shows informative trends that depend on the prompt. Under DA, GPT-4o underperforms GPT-5 across all domains, especially on Lever Equilibrium (overall $45.2\%$) where questions about equilibrium state are particularly challenging. Prompts that incorporate physics awareness partially mitigate these weaknesses: on NFL, both CoP and LC raise overall accuracy (to $64.1\%$ and $65.0\%$), mainly via substantially improved object stability performance (approximately $90\%$). On CoM, CoP achieves GPT-4o's best overall accuracy ($67.6\%$), driven by a considerable improvement in rotation predictions. On LE, LC provides a moderate improvement over DA, although questions about equilibrium state remain difficult across all prompts.

\noindent Taken together, these results demonstrate two key findings. First, GPT-5 is consistently more robust than GPT-4o across all prompting strategies. Second, prompting that is guided by physics principles, particularly the CoP template, is most beneficial in settings that require genuine multistep physical reasoning (NFL and, to a lesser extent, CoM rotation), whereas in visually explicit tasks involving torque balancing, the simple DA prompt already captures most of the relevant information.

\clearpage
\subsection{Detailed Prompt Templates}

Below we present the complete prompt templates used in our ablation study. All three prompts share the same output format requirement but differ in the level of physical guidance provided to the model.

\paragraph{Direct Answering (DA).}

The DA baseline instructs the model to inspect the image and select the correct option without requiring any intermediate reasoning:

\begin{textcolorbox}[Prompt for DA, before skip=0.3em, after skip=0.5em,fontupper=\footnotesize]
Please answer the following multiple choice question based on the image.

Question: \{question\}

Instructions:
- Look carefully at the image
- Choose the correct answer from the options provided
- Only return the letter of your answer (A, B, C, or D)
- Format: Just output ``Answer: X'' where X is A, B, C, or D
\end{textcolorbox}

\paragraph{Law-Conditioned (LC).}

The LC prompt identifies the relevant physical law (e.g., torque balance, Newton's first law, center-of-mass rule) but does not require explicit reasoning steps. Instead, the model is instructed to \emph{use} the named law as guidance when selecting the answer.

\begin{textcolorbox}[Prompt for LC (Law-Cued), before skip=0.3em, after skip=0.5em,fontupper=\footnotesize]

\textbf{You are a physics expert.} This question is mainly about \textcolor{colorcardborder}{\textbf{\{law\}}}.

\noindent\textbf{Question:} \{question\}

\noindent\textbf{Instructions:}
\begin{itemize}[topsep=0pt, itemsep=0pt, parsep=0pt, leftmargin=1.2em]
    \item Carefully inspect the associated image.
    \item Use the named physical law(s) as the main guideline when choosing the answer.
    \item You do not need to write detailed step-by-step reasoning.
    \item At the end, on a new line, output only ``\texttt{Answer: X}'' where X is one of A, B, C, or D.
\end{itemize}

\hrule

\vspace{0.5em}
\noindent\textcolor{colorcardborder}{\textbf{Law Mapping by Question Category:}}

{\small
\setlength{\tabcolsep}{3pt}
\renewcommand{\arraystretch}{0.9}
\begin{tabular}{@{} l @{\hspace{0.4em}$\rightarrow$\hspace{0.4em}} p{0.80\linewidth} @{}}
equilibrium\_state & the torque balance law for levers \\[0.1em]
balance\_adjustment & the torque balance law for levers \\[0.1em]
object\_position & Newton's first law of motion (inertia of translational motion) \\[0.1em]
object\_stability & Newton's first law and rotational stability of rigid bodies \\[0.1em]
arrow\_direction & the rule that a suspended body rotates until its Center of Mass lies directly below the suspension point \\[0.1em]
trajectory & the same center-of-mass principle for suspended bodies \\
\end{tabular}
}

\noindent\textit{Note:} For each question, the placeholder \texttt{\{law\}} is automatically filled according to the question category using the mapping above.

\end{textcolorbox}

\paragraph{Chain-of-Physics (CoP).}

The CoP prompt explicitly requests short, structured physical reasoning before producing an answer. The instructions adapt to the question category: lever questions ask the model to enumerate masses and lever arms and compare torques; Newton's-first-law (NFL) questions compare inertia and friction under different pulling speeds; center-of-mass (CoM) questions remind the model that suspended objects rotate until the Center of Mass lies vertically beneath the suspension point.

\vspace{0.3em}

\begin{textcolorbox}[Prompt Template for CoP, before skip=0.3em, after skip=0.5em,fontupper=\footnotesize]

\textbf{You are a physics tutor.} The question below comes from a mechanics experiment.

\vspace{0.2em}
\noindent\textbf{Question:} \{question\}

\vspace{0.3em}
\noindent\textbf{Instructions:} Choose the appropriate reasoning approach based on the question type.

\vspace{0.3em}
\hrule
\vspace{0.3em}

\noindent\textbf{Output Format:}
\vspace{0.1em}
\begin{enumerate}[topsep=0pt, itemsep=1pt, parsep=0pt, leftmargin=1.5em]
    \item Begin with step-by-step reasoning
    \item On a new line, output only ``\texttt{Answer: X}'' where X is one of A, B, C, or D
\end{enumerate}

\end{textcolorbox}

\begin{textcolorbox}[Prompt for CoP for Lever Equilibrium Questions, before skip=0.3em, after skip=0.5em,fontupper=\footnotesize]

\vspace{0.1em}
\begin{itemize}[topsep=0pt, itemsep=1pt, parsep=0pt, leftmargin=1.2em]
    \item Think like a physics teacher about Lever Equilibrium.
    \item \textbf{Step 1:} List all objects on the left and right of the fulcrum; note their masses and distances.
    \item \textbf{Step 2:} For each side, compute total torque = mass $\times$ distance.
    \item \textbf{Step 3:} Compare the total torque on the left and right.
    \item \textbf{Step 4:} Decide which side rotates downward or whether torques stay equal.
\end{itemize}

\vspace{0.1em}
\noindent\textit{Example:} ``On the left: 2\,kg at 1\,unit $\rightarrow$ torque = 2; on the right: 1\,kg at 3\,units $\rightarrow$ torque = 3. Right side has larger torque and will go down.''

\end{textcolorbox}

\begin{textcolorbox}[Prompt for CoP for Inertia Questions, before skip=0.3em, after skip=0.5em,fontupper=\footnotesize]


\vspace{0.1em}
\begin{itemize}[topsep=0pt, itemsep=1pt, parsep=0pt, leftmargin=1.2em]
    \item Think using Newton's first law (inertia).
    \item \textbf{Step 1:} Identify whether the pull is fast or slow, and whether the object is light or heavy.
    \item \textbf{Step 2:} Recall that a fast pull means the paper accelerates quickly while the object tends to stay at rest; a slow pull lets friction drag the object along.
    \item \textbf{Step 3:} Decide whether the object stays in place, moves with the paper, or shifts slightly; assess tipping risk based on shape and support.
\end{itemize}

\vspace{0.1em}
\noindent\textit{Example:} ``Because the paper is pulled very quickly and friction acts for a short time, the cylinder almost stays where it is due to inertia, so it remains at its original position.''

\end{textcolorbox}

\begin{textcolorbox}[Prompt for CoP for center-of-Mass Questions, before skip=0.3em, after skip=0.5em,fontupper=\footnotesize]

\vspace{0.1em}
\begin{itemize}[topsep=0pt, itemsep=1pt, parsep=0pt, leftmargin=1.2em]
    \item Use the idea that a suspended object rotates until its Center of Mass lies directly below the suspension point.
    \item \textbf{Step 1:} From the image, decide where the Center of Mass is relative to the suspension point (left / right / above / below).
    \item \textbf{Step 2:} Recall that gravity produces a torque that moves the Center of Mass to lie vertically under the suspension point.
    \item \textbf{Step 3:} Infer the direction of rotation (clockwise or anticlockwise) or whether there is no obvious net torque.
\end{itemize}

\vspace{0.1em}
\noindent\textit{Example:} ``The heavy part is to the left of the suspension point, so gravity makes the object rotate anticlockwise until that side hangs lower.''

\end{textcolorbox}

\clearpage
\subsection{Video Generation: Effect of Prompt Design on Physical Rollout}

For video generation, we select two representative models and evaluate their performance 
under the same three prompt templates. 
For each scenario, the initial frame is kept fixed while the textual description is modified 
according to the prompt type. 
We evaluate the generated videos on one representative physics-aware metric per task:
\begin{table*}[h] 
  \small
  \caption{ 
    \textbf{Prompt ablation for PhysicsMind video generation.} 
    Two representative video models (Sora2 and LTX-Video) evaluated on the three physics scenarios
    under three prompt templates. 
    Columns and metric definitions follow Table~\ref{tab:video_generation_results} 
    in the main paper. Values are reported as mean over $n$ runs. 
  } 
  \centering

  \label{tab:video_prompt_ablation_full} 
  \vspace{0.3em} 
  \resizebox{\textwidth}{!}{ 
    \begin{tabular}{llcccccccc} 
      \toprule
      \multirow{2}{*}{Model} & 
      \multirow{2}{*}{Prompt} &
      \multicolumn{2}{c}{Center of Mass (Video)} & 
      \multicolumn{1}{c}{\shortstack{Lever Equilibrium\\ (Video)}} & 
      \multicolumn{5}{c}{Newton's First Law (Video)} \\ 
      \cmidrule(lr){3-4} \cmidrule(lr){5-5} \cmidrule(lr){6-10} 
       & & 
       Seg. Mask IoU & Seg. Mask Center & 
       \shortstack{Final State\\Acc. (\%)} & 
       \shortstack{Trajectory\\RMSE} & 
       \shortstack{Final\\Position Error} & 
       \shortstack{Speed\\Similarity} & 
       \shortstack{Acceleration\\Similarity} & 
       \shortstack{Directional\\Consistency} \\[-3pt] 
      \midrule

      \multirow{3}{*}{Sora2} & Scene-only & 
        0.1002 & 459.74 & 30 & 
        0.384 & 0.087 & -0.018 & 0.004 & 0.4765 \\ 

       & Scene + Action & 
        0.167 & 121.42 & 40 & 
        0.380 & 0.199 & -0.042 & 0.017 & 0.5494 \\ 

       & Scene + Action + Law & 
        0.1341 & 239.77 & 40 & 
        0.402 & 0.243 & 0.052 & 0.021 & 0.5934 \\ 

      \midrule

      \multirow{3}{*}{LTX-Video} & Scene-only & 
        0.0982 & 92.72 & 8 & 
        0.301 & 0.203 & 0.009 & -0.023 & 0.2115 \\ 

       & Scene + Action & 
        0.005 & 76.37 & 4.76 & 
        0.406 & 0.213 & -0.011 & -0.056 & 0.5594 \\ 

       & Scene + Action + Law & 
        0.006 & 520.97 & 12 & 
        0.421 & 0.192 & 0.021 & 0.034 & 0.5479 \\ 

      \bottomrule
    \end{tabular} 
  } 
\end{table*}

\noindent Empirically, the effects of prompt design on video generation are more heterogeneous than on VQA.
For Sora2, moving from a Scene-only description to the canonical Scene+Action prompt consistently
improves center-of-mass tracking (IoU increases from $0.1002$ to $0.167$ and the mask-center error
drops from $459.7$ to $121.4$ pixels), final lever accuracy (from $30\%$ to $40\%$), and directional
consistency in the Newton's-first-law videos (from $0.4765$ to $0.5494$).
Adding an explicit law hint further sharpens local kinematic behavior:
both speed and acceleration similarity become positive (up to $0.052$ and $0.021$), and directional
consistency reaches $0.5934$, but at the cost of slightly worse global trajectory RMSE and
final-position error.
This suggests that Sora2 can partially exploit law-conditioned prompts to align short-horizon
motion trends with the intended physics, while still struggling to globally match the ground-truth path.

\noindent For LTX-Video, the trade-offs are even more pronounced.
Under Scene-only prompts, the model already attains the lowest trajectory RMSE on Newton's-first-law
videos ($0.301$), but with very poor directional consistency ($0.2115$) and low lever accuracy ($8\%$).
The Scene+Action prompt dramatically increases directional consistency (to $0.5594$) at the expense of
higher trajectory error and only marginal lever improvements.
Law-conditioned prompts further improve lever accuracy (to $12\%$) and reduce final-position error,
but also produce highly unstable center-of-mass behavior (mask-center error $>500$ pixels).
Overall, these results indicate that current video models can be steered by richer physics-aware prompts
toward more coherent local motion and lever outcomes, but this steering does not reliably translate
into globally accurate, physically faithful trajectories, and may even destabilize other aspects of the scene.

\noindent This analysis complements the main results by disentangling two factors: 
(i) how much current video generators rely on textual action descriptions versus visual setup; and 
(ii) whether explicit mention of the underlying physical law helps to enforce physically consistent motion, 
beyond purely semantic or aesthetic guidance.

\clearpage
\section{Case Studies}

The following case studies highlight how different classes of multimodal models handle physically grounded reasoning tasks. By examining both static VQA scenarios and dynamic video‑based experiments, we identify characteristic failure modes such as misestimation of geometric cues, over‑reliance on physics priors, and weakness in maintaining temporal or structural consistency. These examples illustrate not only the performance gaps across model families but also the broader challenge of integrating visual perception with physically plausible inference.

\subsection{VQA}

Our VQA case studies focus on situations where the underlying mechanics are
simple but the visual evidence and textual priors can easily come into
conflict. For each scenario, we show the input image, the multiple–choice
question, the ground-truth label, and the free-form rationales produced by
several representative VLMs. The lever-equilibrium example illustrates how different models apply torque reasoning: some attempt to
estimate distances from the fulcrum and derive a numerical comparison, others
assume symmetry or rely mainly on the mass labels, and they often disagree
both with each other and with the benchmark annotation. This exposes a
systematic difficulty in jointly using pixel-level geometry and symbolic
physics when answering even very short mechanical questions. The
center-of-mass example highlights a complementary
failure mode: several models verbally recall the correct rule that a suspended
object rotates until its Center of Mass lies vertically below the pivot, yet
still misidentify the direction of rotation because they mis-locate the mass
distribution in the image. 

\noindent In both cases, the detailed per-model thoughts and answers are printed inside
the figure panels for readability, while the surrounding text distils the main
qualitative patterns: current VLMs can often recite relevant physical laws but
struggle to ground them in precise visual measurements, leading to inconsistent
or self-contradictory predictions.

\clearpage
\begin{textcolorbox}[Lever-Equilibrium VQA Case Study]
\subsection*{Image}

\begin{center}
\begin{minipage}{0.75\linewidth}
    \centering
    \includegraphics[height=2in,width=0.95\linewidth,keepaspectratio]{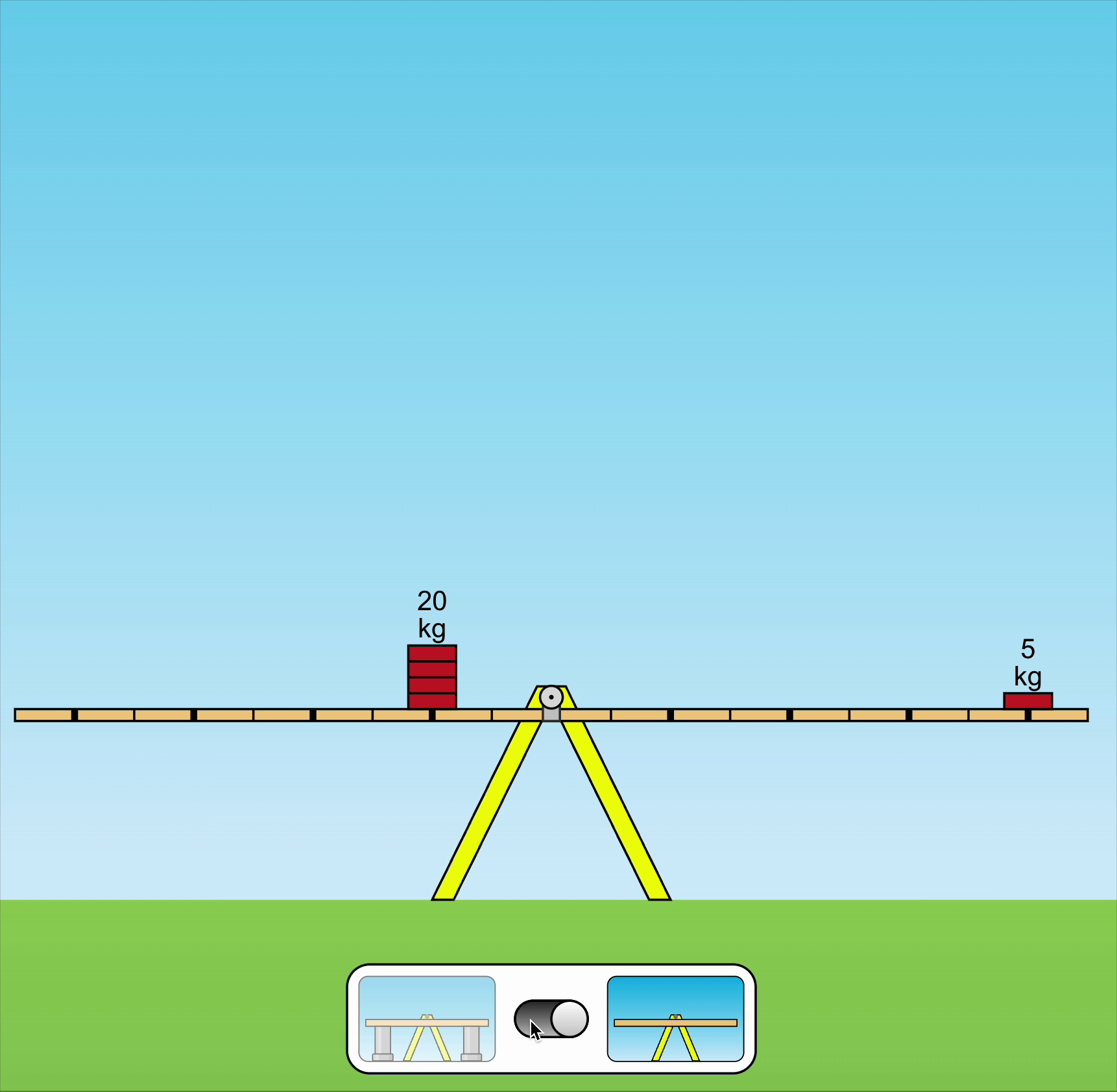}
    \vspace{6pt}

    {\small \textit{Input image for the lever-equilibrium VQA task.}}
\end{minipage}
\end{center}

\vspace{6pt}

\subsection*{Question}

In the image, a lever has a fulcrum at the center with objects suspended on both sides. Observe the mass labels on each object and their distances from the fulcrum. After releasing the lever, what state will the lever be in?

\begin{itemize}
    \item A) Remain horizontally balanced
    \item B) Left end sinks, right end rises
    \item C) Right end sinks, left end rises
    \item D) Cannot be determined
\end{itemize}

\textbf{Instructions:}
Look carefully at the image. Choose the correct answer from the options provided.  
Only return the letter of your answer (A, B, C, or D).  
Format: \textit{Your Thoughts: \textless thoughts\textgreater{} Answer: X}

\vspace{6pt}

\textbf{Ground-truth answer:} A

\end{textcolorbox}

\begin{textcolorbox}[Lever-Equilibrium VQA Case Study - Model Responses]
\subsubsection*{Model Responses}

\paragraph{o4 mini}
\textit{Your Thoughts:}  
The lever is balanced by \textbf{torques}. The 20 kg mass is about 1 unit from the fulcrum (torque \( \approx 20 \)), the 5 kg mass is about 8 units away (torque \( \approx 40 \)), so the \textbf{right-hand torque} wins and the \textbf{right end sinks}.  
\textit{Answer: \textbf{C}}

\vspace{4pt}

\paragraph{grok-4}
\textit{Your Thoughts:}  
The image depicts a seesaw (lever) with a \textbf{fulcrum at the center}. There is a 20 kg mass on the left side and a 5 kg mass on the right. Both masses appear to be placed at \textbf{equal distances} from the fulcrum.  
Left \textbf{torque} \(= 20 \times d\).  
Right \textbf{torque} \(= 5 \times d\).  
Since \(20d > 5d\), the \textbf{left side sinks}.  
\textit{Answer: \textbf{B}}

\vspace{4pt}

\paragraph{Gemini 2.5 Pro}
\textit{Your Thoughts:}  
1. Left mass = 20 kg at approx. 2 units.  
2. Right mass = 5 kg at approx. 4 units.  
Torque\(_{L}\) \(= 40\), Torque\(_{R}\) \(= 20\).  
\textbf{Left torque is greater} \( \rightarrow \) \textbf{left end sinks}.  
\textit{Answer: \textbf{B}}
\end{textcolorbox}

\textbf{Analysis.}
Although the physics implies that the lever should not remain balanced, the benchmark's ground-truth label is A. This case highlights a key challenge in physical VQA: some models (e.g., o4 mini) apply torque reasoning that contradicts the annotation, while others mis-estimate distances or assume symmetry. These discrepancies reveal differences in whether models rely on visual measurement, symbolic physics, or textual priors when interpreting mechanical scenes.

\clearpage
\begin{textcolorbox}[Center-of-Mass VQA Case Study]

\subsection*{Image}

\begin{center}
\begin{minipage}{1\linewidth}
    \centering
    \includegraphics[height=2in,width=0.95\linewidth,keepaspectratio]{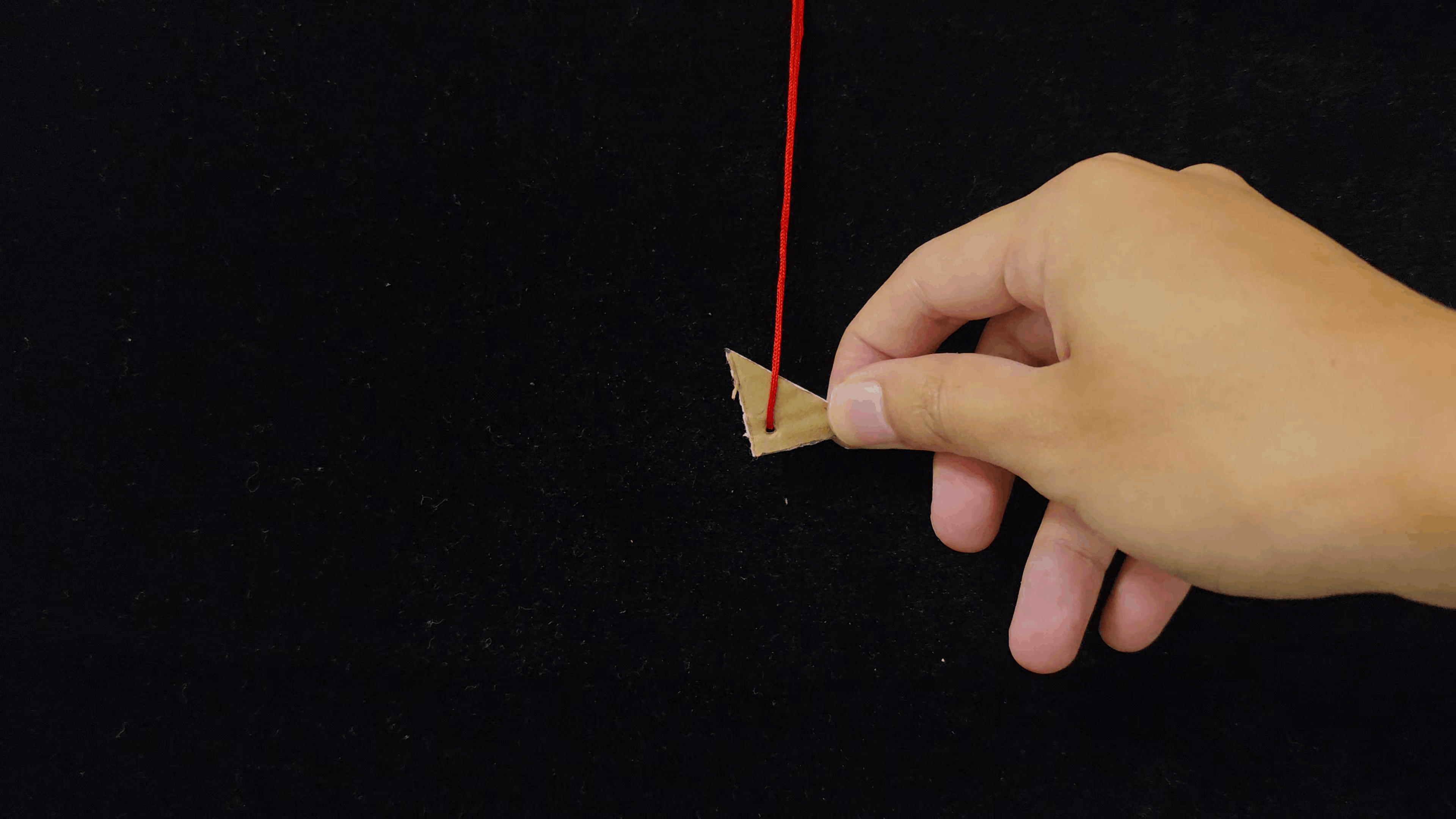}
    \vspace{6pt}

    {\small \textit{Input image for the center-of-mass VQA task.}}
\end{minipage}
\end{center}

\vspace{6pt}

\subsection*{Question}

In the image, where is the hand-held suspension point located relative to the object's center of gravity?

\begin{itemize}
    \item A) To the left of the center
    \item B) To the right of the center
    \item C) Above the center
    \item D) At the center / No clear offset
\end{itemize}

\textbf{Instructions:}
Look carefully at the image. Choose the correct answer from the options provided.  
Only return the letter of your answer (A, B, C, or D).  
Format: \textit{Your Thoughts: \textless thoughts\textgreater{} Answer: X}

\vspace{6pt}

\textbf{Ground-truth answer:} A

\vspace{6pt}

\end{textcolorbox}

\begin{textcolorbox}[Center-of-Mass VQA Case Study - Model Responses ]
\subsection*{Model Responses}

\paragraph{o4 mini}
\textit{Your Thoughts:}  
The red string is threaded through a hole at one corner of the triangular piece, which is clearly \textbf{offset toward the left side} relative to the object's centroid.  
\textit{Answer: \textbf{A}}

\vspace{4pt}

\paragraph{grok-4}
\textit{Your Thoughts:}  
The object hangs freely under gravity, so its \textbf{Center of Mass must lie vertically below the suspension point}. Because the string is vertical, the suspension point is \textbf{above the center of gravity}.  
\textit{Answer: \textbf{C}}

\vspace{4pt}

\paragraph{Gemini 2.5 Pro}
\textit{Your Thoughts:}  
A freely hanging object aligns such that the center of gravity lies directly below the suspension point. Therefore the \textbf{suspension point is above the center}.  
\textit{Answer: \textbf{C}}

\vspace{8pt}
\end{textcolorbox}
\textbf{Analysis.}
This case exposes a divergence between \textbf{geometric observation} and \textbf{physical inference}.  The benchmark label (A) is based on the visible fact that the string passes through a hole located on the \textbf{left side} of the cardboard piece. However, two models (grok-4 and Gemini 2.5 Pro) apply the physical rule that a hanging object’s Center of Mass must lie \textbf{directly below} the suspension point, leading them to answer C. These responses reveal that some models prioritize \textbf{implicit physics priors} over geometric cues, making this an effective diagnostic example for distinguishing visual versus physical reasoning modes in foundation models.

\clearpage
\subsection{Video Generation}

To further assess the strengths and limitations of current video generation models, we conduct detailed qualitative case studies under physically grounded scenarios. These experiments are designed to probe aspects of motion realism, short‑ and long‑term temporal consistency, and object‑to‑object interaction fidelity, which go beyond conventional aesthetic or smoothness metrics.

\noindent\textbf{Case 1: Hanging‑Method (Center of Mass Estimation).}

\begin{figure*}[htbp]
    \centering
    \includegraphics[width=\textwidth]{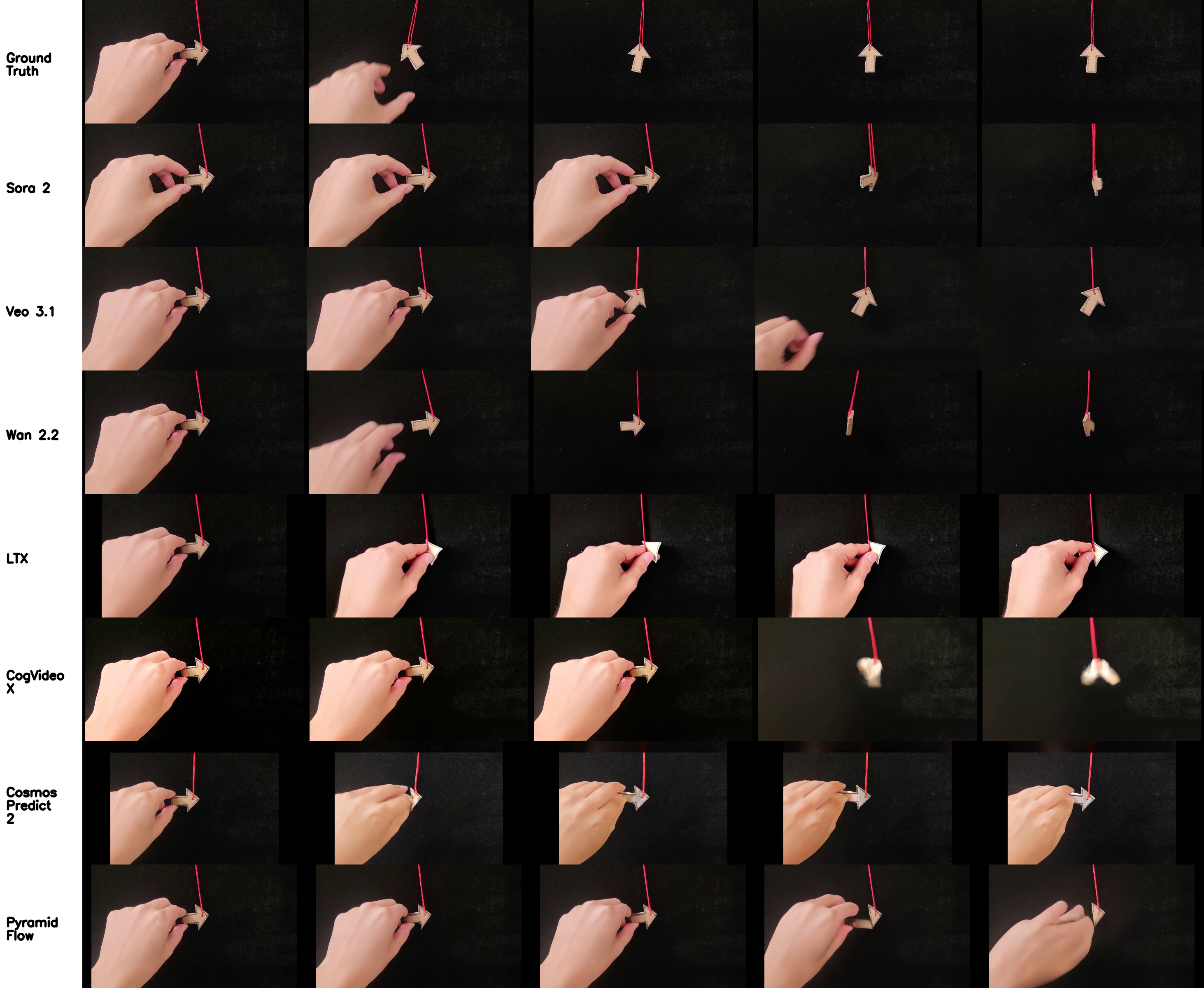}
    \caption{Qualitative comparison on the hanging‑method sequence, where an arrow is suspended by a string and allowed to hang freely. The task assesses temporal stability and physical realism in model‑generated videos.}
    \label{fig:vg_case1}
\end{figure*}

Figure~\ref{fig:vg_case1} visualizes generation results for the hanging‑method sequence, where a small wooden arrow is attached to a thread and released to swing freely until reaching equilibrium. The setup captures low‑frequency, gravity‑driven oscillations and offers a clear cue for assessing spatial stability and coupled motion between the thread and the suspended object.

We observe marked variation across methods. Image‑to‑video transformers such as Sora~2 and Veo~3.1 generally preserve the rigidity and relative geometry of the object–string system while sustaining smooth damping over time. Diffusion‑based models (Wan~2.2, LTX) exhibit locally plausible motion but sometimes fail to maintain proper tension at the joint, resulting in spatial jitter or partial disappearance of the string. Smaller autoregressive models (CogVideo~X, Cosmos~Predict~2, Pyramid~Flow) show pronounced frame‑to‑frame inconsistencies and degraded shape coherence, where the hanging arrow deforms or blends with the background. These artifacts indicate that while recent video models learn coarse dynamic priors, they still struggle with enforcing near‑rigid constraints and globally consistent motion patterns over extended temporal ranges. In particular, the inability to explicitly reason about physical tethering leads to motion drift or visual "snapping" when the string's anchor interacts with the moving body.

\noindent\textbf{Case 2: Rapid Paper‑Pull Experiment.}

\begin{figure*}[htbp]
    \centering
    \includegraphics[width=\textwidth]{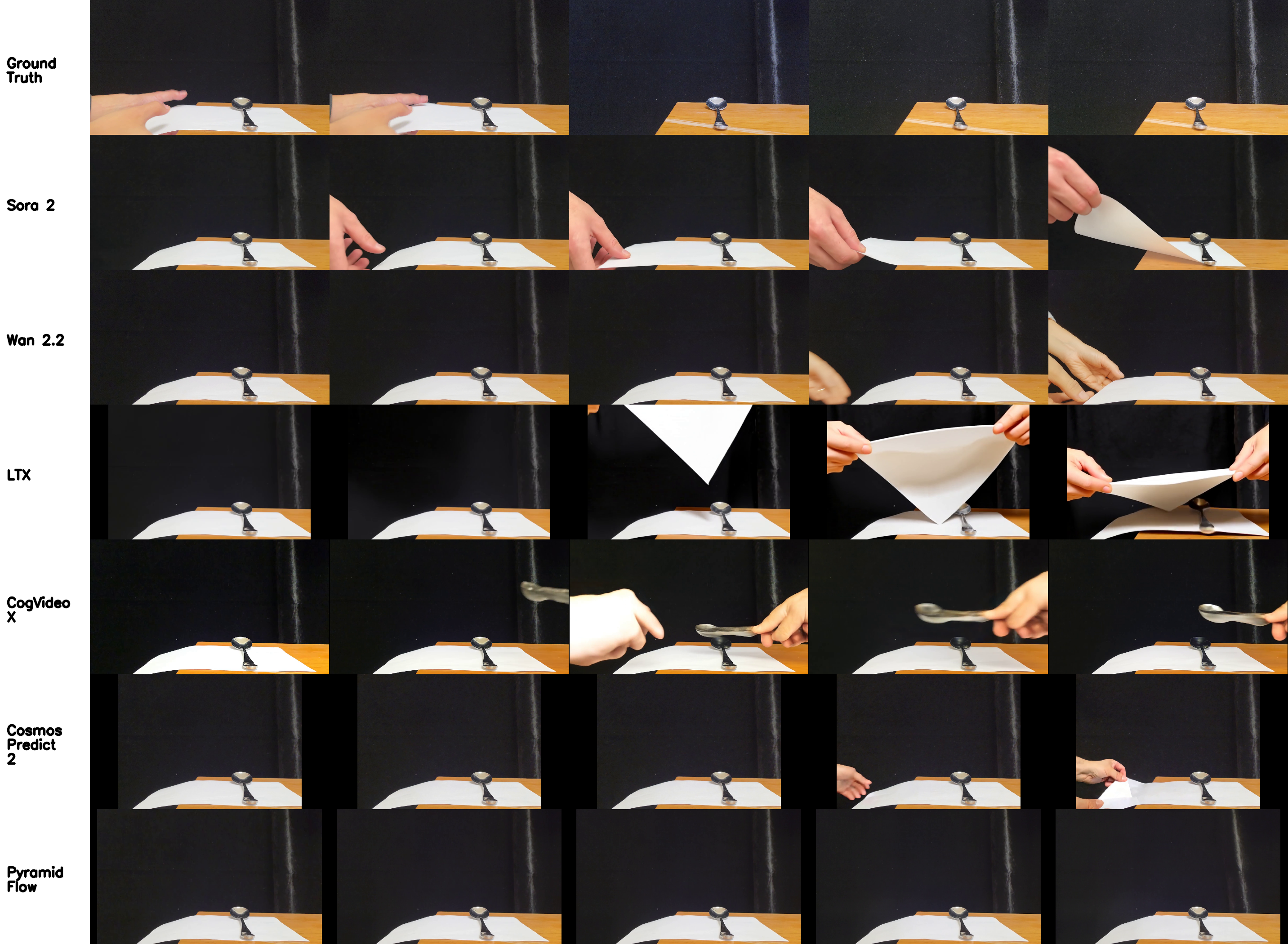}
    \caption{Qualitative comparison on the rapid paper‑pull experiment. A white sheet is quickly pulled from beneath a spoon to evaluate temporal consistency, object‑surface contact, and motion realism.}
    \label{fig:vg_case2}
\end{figure*}

Figure~\ref{fig:vg_case2} presents a contrasting fast‑motion scenario in which a sheet of white paper is swiftly pulled from beneath a metallic spoon. Unlike the previous low‑frequency oscillation task, this sequence emphasizes high‑acceleration, high‑contact dynamics with abrupt state transitions. The ground‑truth motion reveals fine‑grained frictional effects which the spoon remains nearly at rest as the sheet slips away.

Across generative models, we find that temporal desynchronization is the dominant failure mode: the paper's motion either lags behind or advances ahead of the contact event, producing physically implausible interactions. For several diffusion‑based variants, the spoon is occasionally displaced or dragged due to blurred motion conditioning, suggesting that temporal diffusion sampling may propagate appearance priors faster than scene dynamics. Some transformer‑based models show improved temporal sharpness but still exhibit under‑damped temporal blending (ghosting) or inconsistent texture tracking on the paper surface. The best outputs, although qualitatively convincing at a glance, lack the precise sub‑frame contact behavior observed in the reference recording. These cases expose the absence of explicit physical constraints within the generative objective—such as conservation of momentum or mass‑center stability—which are crucial for accurate modeling of sudden impulsive interactions.

\noindent\textbf{Insight and Discussion.}
Together, these two representative cases illustrate complementary dimensions of the video generation challenge. The hanging‑method experiment emphasizes the need for long‑range temporal stability and physically consistent trajectories under slow, continuous motion; the rapid paper‑pull experiment stresses high‑frequency responsiveness and contact‑aware motion reasoning. While large foundation‑scale models have made significant progress in reproducing globally coherent motion and rendering detail, they continue to exhibit a mismatch between visual realism and physical realism. These observations point toward integrating explicit physics priors or differentiable simulation modules into future generative architectures to bridge this gap. Incorporating such constraints could enhance both the temporal consistency and the causal fidelity of next‑generation video generators.

\clearpage
\section{Future work}
While \textsc{PhysicsMind} offers a first step toward systematically probing physical reasoning in modern vision--language and video generation models, it also exposes several limitations that suggest natural directions for future work.

First, our benchmark currently focuses on three canonical laws in relatively controlled tabletop scenes. Extending the suite to cover a broader range of phenomena, such as friction, elastic collisions, conservation of energy, or multi-body interactions, would allow a more complete assessment of physical competence. Increasing scene complexity (e.g., clutter, occlusions, and 3D camera motion) and lengthening the temporal horizon are also important for stress-testing whether models can maintain physically consistent behavior under more realistic conditions.

Second, our evaluation metrics are deliberately designed around a small set of interpretable quantities (final state, Center of Mass, trajectory, and kinematics). There is room to develop richer metrics that better align with human judgements of physical plausibility, or that decompose errors into complementary categories (e.g., geometric vs.\ dynamical mistakes). Another direction is to more tightly couple video metrics with downstream decision-making tasks, such as predicting the outcome of interventions or selecting actions that satisfy physical constraints.

Third, we have used \textsc{PhysicsMind} purely as a diagnostic benchmark. An open question is how to use such structured physics data during training: for example, as a curriculum for targeted finetuning, as an auxiliary self-supervision signal for world models, or as a regularizer that encourages consistency between verbal answers and visual rollouts. Studying whether improvements on \textsc{PhysicsMind} transfer to broader embodied or robotics benchmarks would also help clarify the practical value of stronger physical priors.

Finally, our analysis is limited to a snapshot of current model families. As multimodal foundation models and video generators continue to evolve, it will be important to update the benchmark with new architectures and training paradigms, and to refine the tasks based on community feedback. We hope that \textsc{PhysicsMind} can serve as a compact, extensible testbed that supports this iterative process rather than a definitive verdict on physical reasoning in large models.

\end{document}